\def\draft{1}  

\documentclass[11pt,letterpaper]{article}
\usepackage[margin=1in]{geometry}

\ifnum\draft=1
\fi 

\usepackage[utf8]{inputenc}
\usepackage{amsthm}
\usepackage{complexity}
\usepackage{mathtools,amsmath,amssymb}
\usepackage{graphicx}
\usepackage[table,xcdraw]{xcolor}
\usepackage{multirow}
\usepackage{tabularx}
\usepackage{makecell}
\usepackage[colorinlistoftodos]{todonotes}
\usepackage[colorlinks=true, allcolors=blue]{hyperref}
\usepackage{enumitem}
\usepackage{thm-restate}
\usepackage{natbib}
\usepackage{pifont}
\usepackage{subcaption}

\usepackage[colorlinks=true, allcolors=blue]{hyperref}
\usepackage[nameinlink,capitalise]{cleveref}
\hypersetup{
    citecolor={violet}
}

\ifnum\draft=1
\newcommand{\Cnotes}[1]{{\color{red} [Chi-Ning: #1}]}

\else  \ifnum\draft=0
\newcommand{\Snotes}[1]{}
\newcommand{\Cnotes}[1]{}
\fi \fi 

\usepackage{thmtools}
\usepackage{thm-restate}
\RequirePackage[colorlinks=true]{hyperref}
\hypersetup{
  linkcolor=[rgb]{0,0,0.5},
  citecolor=[rgb]{0, 0.5, 0},
  urlcolor=[rgb]{0.5, 0, 0}
}

\usepackage[breakable]{tcolorbox}
\newtcolorbox{reduction}[2][]
{
  colframe = gray!50,
  colback  = gray!10,
  coltitle = gray!10!black,
  before skip = 10pt,
  after skip = 10pt,
  title    = \textbf{#2},
  #1,
  breakable,
}

\newtcolorbox{game}[2][]
{
  colframe = blue!50,
  colback  = blue!10,
  coltitle = blue!10!black,
  before skip = 10pt,
  after skip = 10pt,
  title    = \textbf{#2},
  #1,
  breakable,
}

\newtcolorbox{examplebox}[2][]
{
  breakable,
  colframe = gray!50,
  colback  = gray!10,
  coltitle = gray!10!black,
  before skip = 10pt,
  after skip = 10pt,
  title    = \textbf{#2},
  #1,
}

\usepackage{algorithmicx}
\usepackage{algorithm}
\usepackage[noend]{algpseudocode}

\algnewcommand\algorithmicinput{\textbf{Input:}}
\algnewcommand\Input{\item[\algorithmicinput]}

\algnewcommand\algorithmicoutput{\textbf{Output:}}
\algnewcommand\Output{\item[\algorithmicoutput]}

\usepackage{amsthm}
\usepackage{thmtools,thm-restate}

\numberwithin{equation}{section}
\declaretheoremstyle[bodyfont=\it,qed=\qedsymbol]{noproofstyle}

\declaretheorem[name=Observation,numbered=no]{observation*}

\declaretheorem[numberlike=equation]{theorem}

\declaretheorem[name=Theorem,numbered=no]{theorem*}

\declaretheorem[name=Lemma,numbered=no]{lemma*}

\declaretheorem[name=Corollary,numbered=no]{corollary*}

\declaretheorem[numberlike=equation]{proposition}
\declaretheorem[name=Proposition,numbered=no]{proposition*}

\declaretheorem[numberlike=equation]{parameter}
\declaretheorem[name=Proposition,numbered=no]{parameter*}

\declaretheorem[name=Claim,numbered=no]{claim*}

\declaretheorem[name=Conjecture,numbered=no]{conjecture*}

\declaretheorem[name=Limitation,numbered=no]{limitation*}

\declaretheorem[name=Question,numbered=no]{question*}

\declaretheoremstyle[bodyfont=\it]{defstyle} 

\declaretheorem[numberlike=equation,style=defstyle]{definition}
\declaretheorem[unnumbered,name=Definition,style=defstyle]{definition*}

\declaretheorem[unnumbered,name=Example,style=defstyle]{example*}

\declaretheorem[unnumbered,name=Notation=defstyle]{notation*}

\declaretheorem[unnumbered,name=Construction,style=defstyle]{construction*}

\declaretheorem[numberlike=equation,style=defstyle]{setting}
\declaretheorem[unnumbered,name=Setting,style=defstyle]{setting*}

\declaretheorem[numberlike=equation,style=defstyle]{assumption}
\declaretheorem[unnumbered,name=Assumption,style=defstyle]{assumption*}

\declaretheoremstyle[]{rmkstyle}

\newcommand{\cmark}{\ding{52}}%
\newcommand{\xmark}{\ding{56}}%
\newcommand{\NTK}{\text{NTK}}
\newcommand{\CKA}{\text{CKA}}

\newcommand{\simcap}{\textsf{sim}}

\usepackage{amsmath,amsfonts,bm}

\def\eqref#1{equation~\ref{#1}}

\def\1{\bm{1}}

\DeclareMathAlphabet{\mathsfit}{\encodingdefault}{\sfdefault}{m}{sl}
\SetMathAlphabet{\mathsfit}{bold}{\encodingdefault}{\sfdefault}{bx}{n}

\DeclareMathOperator{\Tr}{Tr}

\newcommand{\conv}{\textsf{conv}}
\newcommand{\Exp}{\mathop{\mathbb{E}}}
\newcommand{\Acc}{\textsf{Acc}}

\newcommand{\Ptrain}{P_{\text{train}}}

\newcommand{\Real}{\mathbb{R}}
\newcommand{\N}{\mathbb{N}}

\newcommand{\cD}{\mathcal{D}}
\newcommand{\cM}{\mathcal{M}}
\newcommand{\cN}{\mathcal{N}}
\newcommand{\cX}{\mathcal{X}}

\newcommand{\ba}{\mathbf{a}}
\newcommand{\bc}{\mathbf{c}}
\newcommand{\be}{\mathbf{e}}
\newcommand{\bu}{\mathbf{u}}
\newcommand{\bs}{\mathbf{s}}
\newcommand{\bx}{\mathbf{x}}
\newcommand{\by}{\mathbf{y}}

\newcommand{\bv}{\mathbf{v}}

\newcommand{\bg}{\mathbf{g}}
\newcommand{\bh}{\mathbf{h}}

\newcommand{\relu}{\operatorname{ReLU}}

\newcommand{\bt}{\mathbf{t}}

\newcommand{\bV}{\mathbf{V}}

\newcommand{\sgn}{\textsf{sgn}}

\newcommand{\mf}{M}

\newcommand{\proj}{\textsf{proj}}

\usepackage{tikz}

\title{Feature Learning beyond the Lazy-Rich Dichotomy:\\Insights from Representational Geometry\footnote{This work was published in ICML 2025 and was selected for a spotlight presentation.}}

\usepackage{authblk}
\author[1]{Chi-Ning Chou\thanks{These authors contributed equally as first authors. Contact: \texttt{\{cchou, hle, schung\}@flatironinstitute.org}}}
\author[1]{Hang Le$^\dagger$}
\author[1,2]{Yichen Wang}
\author[1,3]{SueYeon Chung}
\affil[1]{Center for Computational Neuroscience, Flatiron Institute, New York, NY, USA.}
\affil[2]{University of California, UCLA, Los Angeles, CA, USA}
\affil[3]{Center for Neural Science, New York University, New York, NY, USA.}
\begin{document}
\date{}
\maketitle

\begin{abstract}
Integrating task-relevant information into neural representations is a fundamental ability of both biological and artificial intelligence systems. Recent theories have categorized learning into two regimes: the rich regime, where neural networks actively learn task-relevant features, and the lazy regime, where networks behave like random feature models. Yet this simple lazy–rich dichotomy overlooks a diverse underlying taxonomy of feature learning, shaped by differences in learning algorithms, network architectures, and data properties. To address this gap, we introduce an analysis framework to study feature learning via the geometry of neural representations. Rather than inspecting individual learned features, we characterize how task-relevant representational manifolds evolve throughout the learning process. We show, in both theoretical and empirical settings, that as networks learn features, task-relevant manifolds untangle, with changes in manifold geometry revealing distinct learning stages and strategies beyond the lazy–rich dichotomy. This framework provides novel insights into feature learning across neuroscience and machine learning, shedding light on structural inductive biases in neural circuits and the mechanisms underlying out-of-distribution generalization.

\end{abstract}


\renewcommand{\thefootnote}{\fnsymbol{footnote}}
\footnotetext[3]{Code is available at \href{https://github.com/chung-neuroai-lab/feature-learning-geometry}{https://github.com/chung-neuroai-lab/feature-learning-geometry}}
\renewcommand{\thefootnote}{\arabic{footnote}}

\section{Introduction}
Learning induces changes in brain activity, whether it involves navigating a new city, adapting novel motor skills, or solving new cognitive tasks. These changes are reflected in the incorporation of task-relevant features into neural representations~\citep{olshausen1996emergence,poort2015learning,niv2019learning,reinert2021mouse,gurnani2023signatures}. Similarly, the remarkable success of deep learning is often attributed to the ability of neural networks to learn problem-specific features\footnote{Different papers and communities might treat the definition of \textit{features} differently. In this paper, we adopt a more general way of thinking about features as measurable properties or characteristics of patterns in data/input represented by neural activity patterns.}. For example, in deep neural networks (DNNs)~\citep{lecun1998gradient,krizhevsky2012imagenet}, the ability to learn rich feature hierarchies enables superior image classification performance~\citep{girshick2014rich}. Meanwhile, the seminal work of Chizat et al.~\citep{chizat2019lazy} demonstrated that neural networks can perform well even when there are negligible changes in the weights of the networks. 
These observations raise important questions: Do neural networks always need to learn task-relevant features? How can we evaluate the quality of the features they learn?

\begin{figure}[ht!]
    \centering
    \includegraphics[width=0.9\textwidth]{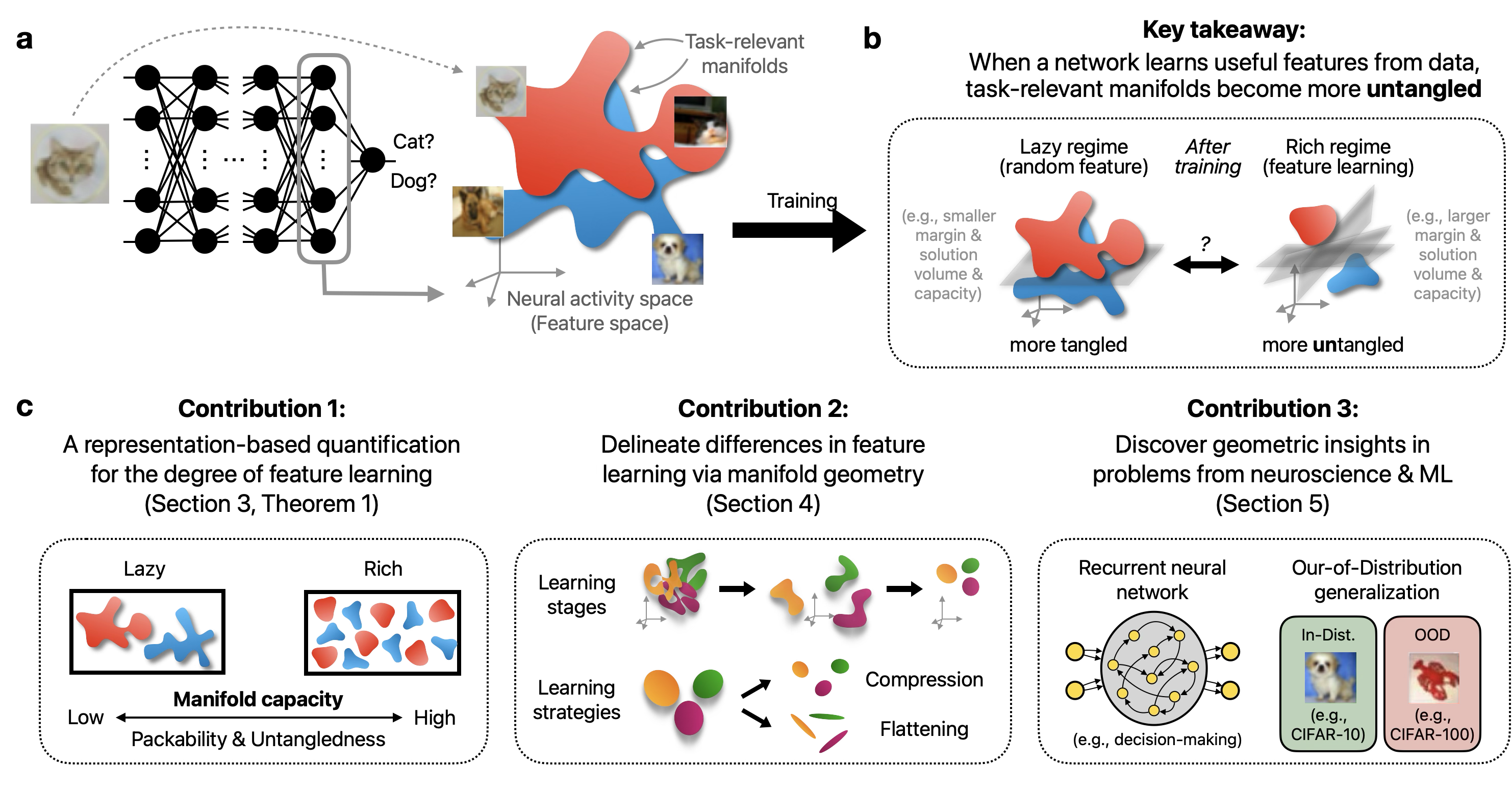}
    \caption{Schematic illustration. 
    \textbf{a}, We propose to investigate feature learning via the geometry of task-relevant manifolds. Here task-relevant manifolds refer to the collection of activity patterns (of a certain layer) to the same task condition (e.g., the same label class). 
    \textbf{b}, Specifically, we found that the degree of manifold untangling (quantified by manifold capacity, \autoref{def:capacity simulated} and~\autoref{def:capacity mean-field}) tracks the degree of richness in feature learning.
    \textbf{c}, Three main contributions of this paper. More details in the corresponding section.}
    \label{fig:schematic}
\end{figure}

To answer these questions, researchers in representation learning have developed several methods to determine whether a neural network operates in  the \textit{lazy} regime (learning without changing internal features) or the \textit{rich} regime (learning task-relevant features)\footnote{These two regimes are also known as \textit{kernel regime} and \textit{feature learning regime} .}. These methods include measuring changes in the weights of the network, tracking activated neurons, and assessing differences in the linearized model (also known as the neural tangent kernel, NTK~\citep{jacot2018neural}). 
Factors such as initial weight norm, learning rate, and readout weight have been found to play a role in whether a network is lazy or rich~\citep{chizat2019lazy}.
Moreover, recent theoretical evidence has suggested that networks could perform better in the rich regime compared to the lazy regime~\citep{yang2021tensor,shi2022theoretical,karp2021local,damian2022neural,ba2022high}.

However, feature learning is much \textit{richer} than the lazy versus rich dichotomy. For example, changes in representations are not always beneficial as they can lead to issues such as catastrophic forgetting~\citep{kirkpatrick2017overcoming}. Moreover, different network architectures, training procedures, and objective functions, initializations, can result in different inductive biases for feature learning~\citep{chizat2019lazy,bordelon2022influence,ba2022high,damian2022neural}, yet all of these scenarios could fall under the broad category of rich learning. Lastly, current limitations in neuroscience technology for precisely tracking synaptic weight changes in neural circuits necessitate a framework based on neural representations rather than network weights or neural tangent kernel.


\subsection{Contributions}
In this work, we go beyond the lazy versus rich dichotomy and address the above-mentioned gaps by investigating feature learning though the geometric properties of task-relevant manifolds.
Here, task-relevant manifolds refer to the point clouds of neural activity patterns that are related to the tasks. For example, in a classification task, a manifold could be the point cloud of neural activations corresponding to stimuli in a given category (e.g., the cat and dog manifolds in~\autoref{fig:schematic}a, left). In other domains, a manifold could correspond to a context (e.g., environmental cues) in a neuroscience experiment or to a concept (e.g., semantic categories) in a language model.

In a network that does not learn task-relevant features (e.g., lazy learning, random features,~\autoref{fig:schematic}b, left), the manifolds are poorly organized, making them harder to distinguish (e.g., smaller margin, smaller solution volume).
In contrast, when a network learns task-relevant features (e.g., rich learning, neural collapse~\autoref{fig:schematic}b, right), the manifolds become well-organized and easier to separate (e.g., larger margin, larger solution volume).
From this perspective, feature learning can be viewed as a process of \textit{untangling} task-relevant manifolds—structuring the neural representational space to improve separation among manifolds.

To make this intuition concrete and quantitative, 
we propose the usage of \textit{manifold capacity}~\citep{chung2018classification,chou2025glue} to quantify the degree of richness in feature learning (\autoref{fig:schematic}c, left).
Specifically, manifold capacity (\autoref{def:capacity simulated} and~\autoref{def:capacity mean-field}) quantifies the degree of manifold untangling via an average-case notion of how separable the manifolds are: manifold packability\footnote{We remark that the margin in support vector machine (SVM) theory quantifies the degree of separability in the worst-case setting. Here the manifold capacity theory is average-case in the sense of the random projection in~\autoref{def:capacity simulated} and the random up-lifting in~\autoref{def:capacity mean-field}. This average-case nature of manifold capacity enables its connection to geometrical properties of the manifolds.}.
Additionally, manifold capacity is analytically connected to a collection of geometric measures, also known as Geometry Linked to Untangling Efficiency (GLUE)~\citep{chou2025glue}, which provide a set of mechanistic descriptors to explain how these manifolds untangle. 

To demonstrate our proposed method, we examine problems in neuroscience and machine learning and find insights that have not been reported. Our contributions are summarized below.

\begin{itemize}
\item (\autoref{sec:capacity quantification}) We use manifold capacity as a representation-based method to quantify the degree of feature learning and demonstrate that it is better than conventional measures across a wide range of settings.
\item (\autoref{sec:effective geometry}) Manifold geometry reveals previously unreported subtypes of feature learning. We find that the training of neural networks undergoes various \textit{learning stages} as shown by the dynamics of manifold geometry, and there are diverse emergent \textit{learning strategies} from networks having different degree of richness in learning.
\item (\autoref{sec:applications}) We find new geometric insights that have not been reported in problems from neuroscience (e.g., structural inductive biases in neural circuits) and machine learning (e.g., out-of-distribution generalization).
\end{itemize}

\subsection{Related work}
Feature learning has been a fundamental research problem in various domains, including neuroscience and machine learning. In neuroscience, understanding the relationship between neural representations and task performance is a central focus~\citep{gao2015simplicity}. Representational geometry~\citep{chung2021neural} has emerged as a promising approach to investigate how different organizations of features can lead to better task performance~\citep{bernardi2020geometry,flesch2022orthogonal,gurnani2023signatures}. There were also works that attempted to infer the underlying learning rules of a neural network using representational geometry~\citep{cao2020characterizing,sorscher2022neural} and low-order statistics~\citep{nayebi2020identifying}.
In machine learning, \textit{visualization techniques}~\citep{zeiler2014visualizing} have been widely used to gain intuitive insights into learned representations, often supplemented with specialized measures to quantify specific properties. 
On the theoretical front, the \textit{kernel method}~\citep{jacot2018neural,lee2019wide} has been a leading approach to analytically characterize the behavior of neural networks, particularly in terms of their deviation from the corresponding kernel. This line of research includes studies on the distinction between lazy and rich regimes~\citep{chizat2019lazy,geiger2020disentangling} and identifying problem settings where neural networks with feature learning outperform kernel methods~\citep{ba2022high,dandi2023two,yang2021tensor}.
For a more comprehensive overview of related work, see~\autoref{app:related work}.

\paragraph{A remark on terminologies.}
Here we clarify the terminology of \textit{manifold untangling}, \textit{manifold packability}, and \textit{manifold capacity}. Manifold untangling originates from neuroscience~\citep{dicarlo2007untangling} and refers to the intuition that task-relevant manifolds become increasingly separable in a high-dimensional state space. Several methods have been proposed to quantify the degree of untangling, such as decoding accuracy~\citep{yamins2016using, hong2016explicit}. Manifold packability, inspired by the sphere-packing intuitions from physics and mathematics, represents an average-case notion of manifold separability. It naturally provides a more concrete framework for defining manifold untangling within a system. Finally, manifold capacity analytically quantifies manifold packability in terms of linear classification, making it a useful metric for assessing the degree of untangling. Further details on manifold capacity theory will be discussed in~\autoref{sec:method MCT} and~\autoref{app:MCT}.

\section{Method and Setup}\label{sec:method}

\subsection{Rich and lazy learning in neural networks}
We studied rich versus lazy learning in two standard settings: 2-layer non-linear neural networks on synthetic data and feedforward deep neural networks on real image classification datasets~\citep{chizat2019lazy}. All analyses were performed on the test data representations in the last layer.

\paragraph{A scale factor for interpolating between rich and lazy regime.}
In all experiments, we use the \textit{inverse scale factor} $\bar{\eta}$ as a tunable ground truth for the degree of feature learning. In particular, $\bar{\eta}$ controls the magnitude of the output of the network as in~\citep{chizat2019lazy}. Intuitively, a larger $\bar{\eta}$ indicates that the learning rate of intermediate layers is faster compared to that of the readout weights, resulting in a richer learning process.
See~\autoref{app:2-layer} and~\ref{app:dnn} for more details. 

\paragraph{2-Layer non-linear neural networks.}
We considered standard 2-layer neural networks with non-linear activation functions and trained with gradient descent. We also considered a data model to generate random point clouds as input manifolds. This setting serves as a well-curated testbed for testing the proposed methodology and showcasing intuitions. See~\autoref{app:2-layer} for more details.

\paragraph{Deep neural networks.}
The goal of this work is to develop a framework to understand neural representations rather than pushing the benchmark. Therefore, we focused on models and settings that are large enough to see interesting phenomena, while the computational cost is still reasonable. 
Specifically, we considered feedforward DNN architectures such as VGG-11~\citep{simonyan2015very} and ResNet-18~\citep{he2016deep} and datasets CIfAR-10~\citep{cifardataset}, CIfAR-100~\citep{cifardataset}, CIfAR-10C~\citep{hendrycks2018benchmarking}. This setting illustrates the applicability of our methodology to DNNs. See~\autoref{app:dnn} for more details.

\paragraph{Task-relevant manifolds.}
Let $P$ be the number of classes and $N$ be the number of neurons (in the layer of interest\footnote{In this paper, we primarily consider the last layer in most experiments unless specified otherwise.}). The $i$-th class manifold is modeled as the convex set\footnote{In the context of linear classification, it is mathematically equivalent to study the convex hull of a manifold.} $\cM_i=\conv(\{\Phi(x):x\in \cX_i\})$ where $\cX_i$ is the collection of inputs in the $i$-th class, $\Phi(x)$ is the representation for $x$, and $\conv(\cdot)$ denotes the convex hull of a set.

\subsection{Manifold capacity theory}\label{sec:method MCT}
Manifold capacity theory~\citep{chung2018classification,chou2025glue} was originally developed to study the \textit{untangling hypothesis}\footnote{In computational neuroscience, the ``untangling hypothesis'' posits that the brain transforms complex, entangled sensory inputs into more linearly separable representations, facilitating efficient object recognition.} of invariant object recognition in vision neuroscience ~\citep{dicarlo2007untangling}.
The theory extends the classic notion of storage capacity of points~\citep{cover1965geometrical,gardner1988optimal,gardner1988space} to object manifolds, i.e., the collection of neural representations that are invariant to the same input category (\autoref{fig:schematic}a, left). 
A \textit{simulated} version of manifold capacity is defined as follows.

\begin{figure}[h]
    \centering
    \begin{minipage}{0.6\textwidth} 
        \begin{restatable}[Simulated manifold capacity~\citep{cohen2020separability,chou2025glue}]{definition}{simcapdef}\label{def:capacity simulated}
        Let $P,N\in\mathbb{N}$ and $\cM_i\subseteq\Real^N$ be convex sets for each $i\in[P]=\{1,\dots,P\}$. For each $n\in[N]$, define
        \[
        p_n := \Pr_{\by,\Pi_n}[\exists \theta\in\Real^n\, :\, y_i\langle\theta,\bs\rangle\geq0,\, \forall i\in[P],\, \bs\in\cM_i)]
        \]
        where $\by$ is a random dichotomy sampled from $\{\pm1\}^P$ and $\Pi_n$ is a random projection operator from $\Real^N$ to $\Real^n$.
        Suppose $p_N=1$, the simulated capacity of $\{\cM_i\}_{i\in[P]}$ is defined as
        \[
        \alpha_{\simcap} := \frac{P}{\sum_{n\in[N]}(1-p_n)} \, .
        \]
        \end{restatable}
    \end{minipage}%
    \hfill
    \begin{minipage}{0.3\textwidth} 
        \centering
        \includegraphics[width=\linewidth]{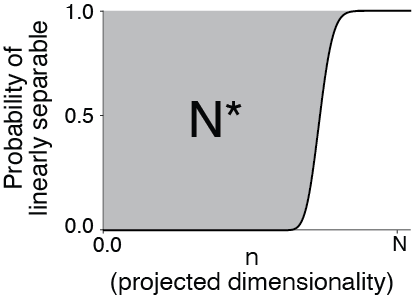} 
        \caption{Simulated capacity.}
        \label{fig:example}
    \end{minipage}
\end{figure}


Intuitively, the simulated manifold capacity measures the \textit{packability}~\citep{chung2018classification} of manifolds by determining the smallest dimensional subspace needed to ensure that the manifolds can be separated. Namely, manifolds that are more packable (i.e., separable when projected to smaller dimensional subspaces) exhibit higher manifold capacity.
While~\autoref{def:capacity simulated} provides a quantitative description for \textit{packability}, it is computationally expensive to estimate and is not analytically trackable.
In~\citep{chou2025glue}, the authors resolved these issues by considering a mean-field version of the manifold capacity (formal definition deferred to~\autoref{def:capacity mean-field}), denoted as $\alpha_{\mf}$, which is analytically trackable and has the property that $|\alpha_{\simcap}-\alpha_{\mf}|=O(1/N)$. In particular,~\citep{chou2025glue} derived that
\begin{align}
\alpha^{-1}_{\mf} &= \frac{1}{P}\Exp_{\substack{\by\sim\{\pm1\}^P\\\bt\sim\cN(0,I_{N})}}\left[\max_{\bs_i\in\cM_i}\left\{\|\proj_{\textsf{cone}(\{y_i\bs_i\})}\bt\|_2^2\right\}\right] \label{eq:capacity formula main text}
\end{align}
where $\cN(\mu,\Sigma)$ denotes the multivariate Gaussian distribution with mean $\mu$ and covariance $\Sigma$ and $\textsf{cone}(\cdot)$ is the convex cone spanned by the vectors, i.e., $\textsf{cone}(\{y_i\bs_i\})=\{\sum_i\lambda_iy_i\bs_i\, :\, \lambda_i\geq0\}$.

\subsection{Geometry linked to untangling efficiency (GLUE) via manifold capacity}
\begin{figure}[ht]
    \centering
    \includegraphics[width=\linewidth]{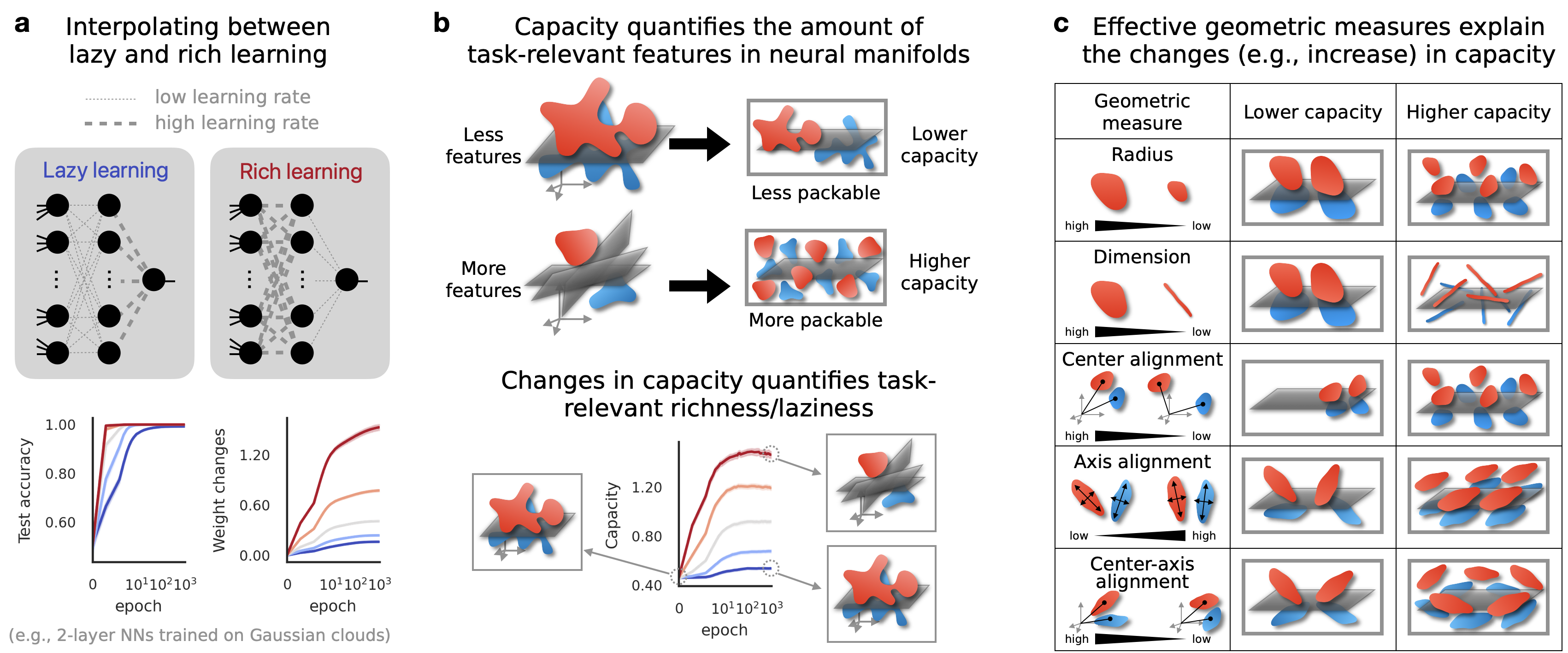}
    \caption{
    Our methods. 
    \textbf{a}, Top: We adopt the method from~\citep{chizat2019lazy} which interpolates lazy and rich learning via adjusting a scale factor of learning rate. Bottom: Test accuracy increases during both lazy and rich training, however, the network's weights would not change much during lazy training.
    \textbf{b}, Top: Higher capacity means that the neural representational space can pack more manifolds (\autoref{def:capacity simulated}). Bottom: We propose to use changes of capacity across training to study task-relevant richness/laziness in feature learning (\autoref{sec:capacity quantification}). Top: we consider the setting in~\citep{chizat2019lazy} where VGG-11 was trained on CIfAR-10. An inverse scale factor was introduced to interpolate between lazy and rich training, where smaller value (blue) corresponds to lazier learning and larger value corresponds to richer learning (red). Bottom: we show that the changes in capacity faithfully tracks the degree of richness in feature learning.
    \textbf{c}, Effective geometric measures drive the capacity value, providing mechanistic descriptors to study representational changes in feature learning. Center-axis alignment has a more complex relationship with capacity, discussed~\autoref{app:geometry}.
    }
    \label{fig:method}
\end{figure}

The advantages of mean-field manifold capacity are: (i) $\alpha_{\mf}$ can be estimated via solving a quadratic program (\autoref{alg:capacity}) and (ii)~\autoref{eq:capacity formula main text} connects manifold capacity to the structure of the manifolds $\{\cM_i\}$. Specifically, for each $\by,\bt$, define$\{\bs_i(\by,\bt)\} = y_i\cdot\arg\max_{\{\bs_i\}}\|\proj_{\textsf{cone}(\{y_i\bs_i\})}\bt\|_2^2$ as the \textit{anchor points} with respect to $\by$ and $\bt$.
Intuitively, these anchor points are the support vectors with respect to some random projection and dichotomy as in~\autoref{def:capacity simulated}.
Namely, the randomness\footnote{Also known as \textit{disorder} in spin glass theory.} induces a distribution of anchor points supported on the manifolds $\{\cM_i\}$.
Specifically, these anchor points are analytically linked to manifold capacity via~\autoref{eq:capacity formula main text} through the capacity formula (\autoref{eq:capacity formula main text}).
This connection inspired the previous work~\citep{chung2018classification,chou2025glue} to define the following effective manifold geometric measures that capture the structure of manifolds while being analytically connected to capacity (see~\autoref{fig:method}c and~\autoref{app:MCT}).

\begin{definition}[Effective manifold geometric measures~\citep{chou2025glue}, simplified version]
For each $i\in[P]$, define $\bs_i^0:=\Exp_{\by,\bt}[\bs_i(\by,\bt)]$ as the \textbf{center} of the $i$-th manifold and define $\bs_i^1(\by,\bt):=\bs_i(\by,\bt)-\bs_i^0$ to be the \textbf{axis} part of $\bs_i(\by,\bt)$ for each pair of $(\by,\bt)$.
\begin{itemize}
\item \textbf{Manifold dimension} captures the degree of freedom of the noises/variations within the manifolds. It is approximately $D_\mf\approx\Exp_{\by,\bt}\left[\frac{1}{P}\sum_i\left(\frac{\langle\bs^1_i(\by,\bt),\bt\rangle}{\|\bs^1_i(\by,\bt)\|_2}\right)^2\right]$, which is analogous to the Gaussian width of the manifolds~\cite[Chapter 7]{vershynin2018high}. See~\autoref{def:geometry full} for the formal definition.
\item \textbf{Manifold radius} captures the noise-to-signal ratio of the manifolds. It is approximately $R_\mf\approx\Exp_{\by,\bt}\left[\frac{1}{P}\sum_i\frac{\|\bs^1_i(\by,\bt)\|_2^2}{\|\bs^0_i\|_2^2}\right]$. See~\autoref{def:geometry full} for the formal definition.
\item \textbf{Center alignment} captures the correlation between the center of different manifolds. Formally, it is defined as $\rho^c_\mf:=\frac{1}{P(P-1)}\sum_{i\neq j}|\langle\bs_i^0,\bs_j^0\rangle|$.
\item \textbf{Axis alignment} captures the correlation between the axis of different manifolds. Formally, it is defined as $\rho^a_\mf:=\frac{1}{P(P-1)}\sum_{i\neq j}\Exp_{\by,\bt}[|\langle\bs_i^1(\by,\bt),\bs_j^1(\by,\bt)\rangle|]$.
\item \textbf{Center-axis alignment} captures the correlation between the center and axis of different manifolds. Formally, it is defined as $\psi_\mf:=\frac{1}{P(P-1)}\sum_{i\neq j}\Exp_{\by,\bt}[|\langle\bs_i^0,\bs_j^1(\by,\bt)\rangle|]$.
\end{itemize}
\end{definition}

Three important remarks on effective manifold geometric measures to be made: First, the changes in manifold capacity can be explained by the changes of these geometric measures. For example, the decrease of manifold radius and dimension makes the capacity higher (see~\cref{fig:method}c,~\cref{app:geometry}).
Second, these effective geometric measures faithfully track the corresponding underlying geometric properties in well-studied mathematical settings (see~\cref{app:connection between capacity and geometry}). Moreover, there is a simple formula connecting manifold capacity with effective geometric measure: $\alpha_\mf \approx (1+R^{-2}_\mf)/D_\mf$ (see~\cref{app:MCT} for details).
Finally, combining the above two points, these effective geometric measures serve as intermediate-level descriptors to investigate how different structural properties of neural manifolds contribute to the changes of task-level performance.

\section{Manifold capacity quantifies the degree of feature learning }\label{sec:capacity quantification}

In this section, we provide both empirical and theoretical justifications for using the increase in capacity during training as a measure to quantify the degree of richness (or the amount of task-relevant features) in feature learning. Furthermore, we compare our method with conventional approaches in the study of lazy versus rich learning, highlighting the new insights uncovered by our approach.

\subsection{Justifications of capacity for quantifying the lazy versus rich dichotomy}

\begin{figure}[ht]
    \centering
    \includegraphics[width=0.95\textwidth]{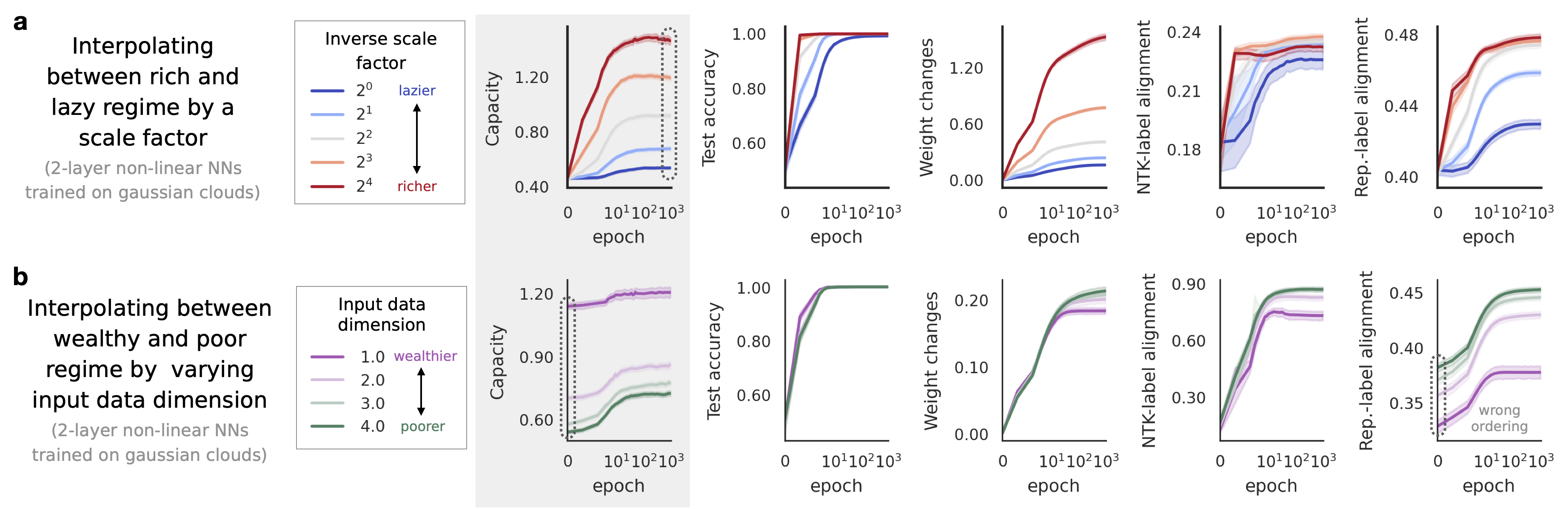}
    \caption{Capacity as a measure for the degree of feature learning. See~\autoref{app:2l_experiment_setup} for the experimental setup.
    \textbf{a}, We interpolated between lazy and rich regime in 2-layer NNs trained to classify Gaussian clouds. We found that capacity could tell the difference between the underlying scale parameter better than the other conventional methods.
    \textbf{b}, We fixed a scale parameter and initialized the input Gaussian clouds with different dimensions (the higher the poorer the initial representations are for each class). We found that capacity could tell the difference in the amount of tasks-relevant features at initialization than other conventional methods. Specifically, the representation-label alignment would characterize the wrong ordering of wealthiness in initial features. 
    }
    \label{fig:capacity}
\end{figure}

\paragraph{Empirical justification in standard settings.}
We start with empirically justifying the use of capacity to quantify the degree of feature learning. 
A classic result in the literature of lazy versus rich training is to train a lazy network where the test accuracy improves, but the weight matrices (or kernels) do not change much before and after training. We consider two settings in~\citep{chizat2019lazy}, one is feedforward DNNs (VGG-11 and ResNet-18) trained on CIfAR-10 (\autoref{fig:method}b), and the other is 2-layer non-linear NNs trained on random point clouds (\autoref{fig:capacity}a).
In both cases, we observe that the manifolds are more untangled when training is richer and capacity correctly tracks the degree of feature learning (the ground truth being the scale parameter $\bar{\eta}$). This provides empirical justification for the use of capacity as well as evidence for manifold untangling in the rich learning regime.

\paragraph{Theoretical justification on 2-layer non-linear neural networks.} 
To strengthen the connection between capacity and feature learning, we next consider a well-studied theoretical model~\citep{ba2022high,montanari2019generalization} and analytically characterize the relationship between capacity, prediction error, and the effective degree of richness.
Concretely, we consider the training of a fully-connected 2-layer network of the form $f(\bx)=\frac{1}{\sqrt{N}}\ba^\top\sigma(W^\top\bx)$, where $\bx\in\Real^d$ is an input, $W\in\Real^{N\times d}$ is the hidden layer matrix, $\ba\in\Real^N$ is the readout weight, and $\sigma:\Real\to\Real$ is the (non-linear) activation function. 
To study feature learning in this setting, it is common to consider $W$ to be randomly initialized (i.e., random feature model~\citep{rahimi2007random}) and update via gradient descent with squared loss.
Meanwhile, the readout weight $\ba$ is randomly initialized and fixed to avoid lazy learning (where the network minimally adjusts the hidden layer and focuses on learning a good readout weight) as well as enable mathematical analysis~\citep{ba2022high}.
Input data and label $(\bx_1,y_1),\dots,(\bx_{P_{\text{train}}},y_{P_{\text{train}}})$ were randomly generated by a teacher-student setting, where there is a hidden signal direction $\beta^*$ that correlates with the label (see~\autoref{setting:2-layer} for the full setting). 
As previously proved in~\citep{ba2022high} (see~\autoref{prop:ba}), in the proportional asymptotic limit (i.e., $P_{\text{train}},d,N\to\infty$ at the same rate), the first-step gradient update can be approximated by a rank-1 matrix that contains label information, resulting in the updated weight to be more aligned with the hidden signal $\beta^*$. Hence, in this setting, the learning rate $\eta$ can be used as the ground-truth to measure the amount of task-relevant information (i.e., richness in learning) in the model representation after gradient updates.

We extend the previous results in~\citep{ba2022high} from a regression setting to a classification setting. Specifically, We prove that capacity correctly tracks the effective degree of richness after one gradient step\footnote{Here we follow the convention in~\citep{ba2022high} and study only the first gradient step as the key Gaussian equivalence step might not hold for more steps as remarked in footnote 2 of~\citep{ba2022high}.}. Moreover, we derive a monotone connection between capacity and prediction accuracy. 
Here, we provide an informal statement of our results and leave the formal version and proof in~\autoref{app:2-layer theory}.

\begin{theorem}\label{thm:1}
Given~\autoref{assumption:2-layer} and~\autoref{setting:2-layer}. Let $0<\eta<\infty$ be the learning rate of a one-step gradient descent with squared loss and $\psi_1=\frac{N}{d},\psi_2=\frac{P_{\text{train}}}{d}$ where $P_{\text{train}}$ is the number of training points, $d$ is the input dimension, and $N$ is the number of hidden neurons. Let $\alpha_{P_{\text{train}},d,N}(\eta)$ be the capacity and let $\Acc_{P_{\text{train}},d,N}(\eta)$ be the prediction accuracy after a gradient step with learning rate $\eta$.
We have
\begin{enumerate}[leftmargin=15pt,topsep=0pt, itemsep=0pt]
\item (Capacity tracks the degree of richness) 
$
\alpha_{P_{\text{train}},d,N}(\eta) \xrightarrow{P_{\text{train}},d,N\to\infty} \alpha(\eta,\psi_1,\psi_2)
$
where $\alpha(\cdot,\cdot,\cdot)$ is defined in~\autoref{thm:storage capacity formal}. Specifically, $\alpha(\eta,\psi_1,\psi_2)<\alpha(\eta',\psi_1,\psi_2)$ for every $0<\eta<\eta'$.
\item (Capacity links to prediction accuracy)
$
\Acc_{P_{\text{train}},d,N}(\eta) \xrightarrow{P_{\text{train}},d,N\to\infty} \Acc(\eta,\psi_1,\psi_2)
$
where $\Acc(\eta,\psi_1,\psi_2)$ is formally defined in~\autoref{thm:storage capacity formal}. In particular, there exists an increasing and invertible function $h_{\psi_1,\psi_2}:\Real_+\to[0,1]$ such that $\Acc(\eta,\psi_1,\psi_2)=h_{\psi_1,\psi_2}(\alpha(\eta,\psi_1,\psi_2))$.
\end{enumerate}
\end{theorem}
The above theorem justifies the usage of capacity as a measure for the degree of richness in feature learning within a well-studied theoretical setting.
We remark that our proof requires substantial technical improvements from~\citep{ba2022high} due to the difference between regression and classification  (e.g., analyzing the margin of the Gaussian equivalent model after one-step gradient using tools from~\citep{montanari2019generalization}, ~\autoref{prop:MRSY}).

\subsection{Comparison with conventional feature learning measures.}\label{para:comparing_feature_metrics}
Here we compare the capacity with several common measures for feature learning: accuracy curves, weight changes, and alignment methods. Concretely, weight changes at the $t$-th epoch is defined as $\|W_t-W_0\|_F/\|W_0\|_F$ where $W_t$ is the weight matrix at the $t$-th epoch. NTK-label alignment and representation-label alignment at the $t$-th epoch are defined as $\CKA(K^{\NTK}_t,\by\by^\top)$ and $\CKA(X_tX_t^\top,\by\by^\top)$ respectively, where $\by$ is the label vector, $\CKA(\cdot,\cdot)$ is the center kernel alignment measure~\citep{kornblith2019similarity}, $K^{\NTK}_t$ is the neural tangent kernel and $X_t$ is the representational matrix at the $t$-th epoch. 
In order to test these measures in a wide variety of settings, we consider 2-layer NNs with synthetic data where we can vary a wide range of parameters. See~\autoref{app:related work} for a detailed introduction to these methods and~\autoref{app:2-layer} for more experimental details.

\paragraph{Capacity can detect the presence of task-relevant features in data.}
In~\autoref{fig:capacity}a, we consider 2-layer NNs trained on random Gaussian clouds with gradient descent. We vary the scale parameter of the network to interpolate between lazy and rich regimes as done in~\citep{chizat2019lazy}.
We find that capacity is better at telling the difference of effective richness (i.e., the scale parameter) of the training than other conventional measures (\autoref{fig:capacity}a). In particular, when the training is richer, we expect the representations to exhibit more complex structures. 
Manifold capacity excels at extracting task-relevant structures in representations because it is data-driven and free from additional statistical assumptions on the data~\citep{chung2018classification,chou2025glue}.

\paragraph{Capacity can quantify the differences in task-relevant features at initialization.}
When comparing two networks with different initializations, focusing solely on network changes can overlook differences in features present at initialization. 
Here, we use the capacity value at initialization to determine whether a network is in a wealthy regime (i.e., possessing more task-relevant features) or a poor regime (i.e., possessing less task-relevant features), as shown in (\autoref{fig:capacity}b). 
The wealthy versus poor distinction provides insight into the network’s initial state, allowing for a more comprehensive comparison of different settings (see`\autoref{subsec:rnn} for an example).

\begin{table}[ht]
    \centering
    \fontsize{8}{8}\selectfont
    \renewcommand{\arraystretch}{2}
    \begin{tabular}{l>{\columncolor[HTML]{EFEFEF}}ccccc}
        \Xhline{2\arrayrulewidth}
        & \makecell{Our approach\\(manifold geometry)} & \makecell{Accuracy} & \makecell{Weight\\changes} & \makecell{NTK-label\\alignment} & \makecell{Representation-label\\ alignment} \\ \Xhline{2\arrayrulewidth}
        \makecell{Detect the changes\\in features} & \cmark & \xmark & \cmark & \cmark & \cmark \\ \hline
        \makecell{Quantify the amount of\\task-relevant features} & \cmark & \xmark & \xmark & \xmark\footnotemark & \xmark\footnotemark[\value{footnote}]\\ \hline
        \makecell{Representation-based} & \cmark & \xmark & \xmark & \xmark & \cmark \\ \hline
        \makecell{Delineate subtypes of\\feature learning} & \cmark & \xmark & \xmark & \xmark & \xmark \\ \Xhline{2\arrayrulewidth}
    \end{tabular}
    \caption{Comparison with conventional measures used in lazy versus rich learning.}
    \label{tab:feature_changes}
\end{table}
\footnotetext{See~\autoref{fig:capacity} for examples of how NTK-label alignment and representation-label alignment could fail at quantifying the amount task-relevant features.}

\section{Manifold Geometry Reveals Subtypes of Feature Learning}\label{sec:effective geometry}
In this section, we demonstrate that feature learning is much richer than the lazy versus rich dichotomy.
In particular, we use manifold geometric measures (\autoref{fig:method}c, and~\autoref{app:MCT} for details)
to delineate the differences in the learned features (learning strategies) of neural networks and representational changes throughout training (learning stages).
The key takeaway from this section is the ability of our method to reveal task-relevant changes in neural representations. 


\subsection{Geometric differences in learned features: Learning strategies}\label{subsec:learning_strategies}
To increase capacity, a network can shrink the radius or compress the dimension of neural manifolds (\autoref{fig:method}c). 
We demonstrate in 2-layer NNs the emergence of distinct learning strategies driven by different factors. In~\autoref{fig:learning geometry}a, we consider the setting in~\autoref{fig:capacity}a where we interpolate the degree of richness in feature learning via an inverse scale factor. As training moves from the lazy to a richer regime (blue to gray), the network compresses both the radius and dimension to increase capacity. Interestingly, in an even richer regime (gray to red), the network sacrifices radius to further reduce dimension.
In~\autoref{fig:learning geometry}b, we consider the setting in~\autoref{fig:capacity}b where we interpolate the wealth of initialization by varying input data dimension. For the wealthiest initialization (purple), the network primarily compresses radius. For poorer initialization (green), both radius and dimension are compressed in lazier training, while in the richer regime (e.g., inverse scale factor $2^4$), the network sacrifices radius for further dimension compression. 
In summary, varying degrees of richness in feature learning can exhibit different learning mechanisms, as captured by manifold geometry.

\subsection{Manifold geometry changes through the course of training: Learning stages}\label{subsec:learning_stages}
Neural networks learn in a highly non-monotonic manner throughout the training period. Examples include double descent~\citep{belkin2019reconciling,nakkiran2021deep,mei2022generalization} and grokking~\citep{power2022grokking,liu2022towards,nanda2023progress,kumargrokking}. Previous works have analytically or empirically described the different stages/phases such as comprehension, grokking, memorization, and confusion~\citep{liu2022towards} through the trajectory of accuracy curves. 

\begin{figure}[ht]
    \centering
    \includegraphics[width=\textwidth]{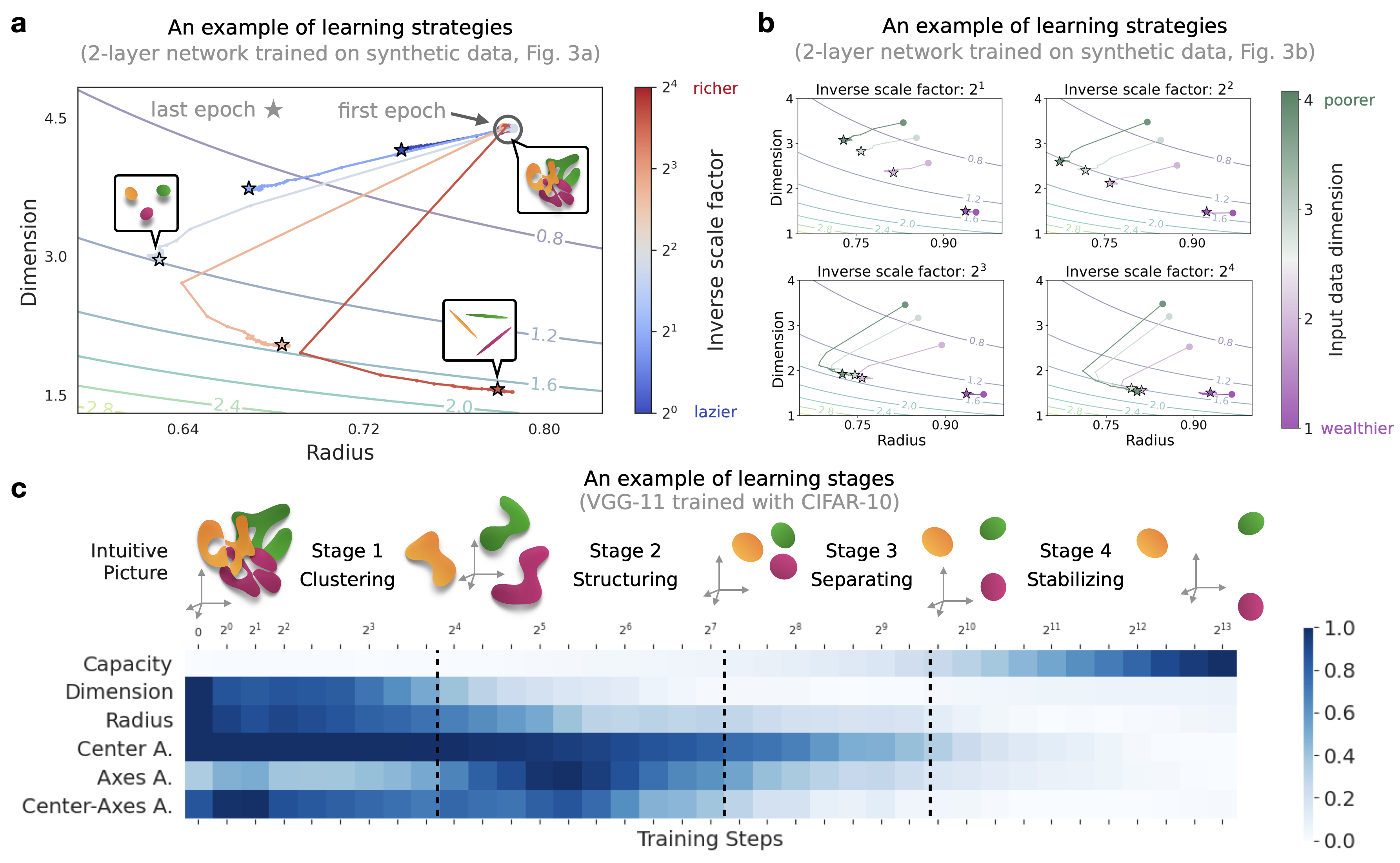}
    \caption{Manifold geometry characterizes learning strategies and learning stages.
    \textbf{a}, Capacity contour plot of the example from~\autoref{fig:capacity}a. The x-axis is the average manifold radius $R_\mf$, the y-axis is the average manifold dimension $D_\mf$, and the contour is the geometric approximation of capacity, i.e., $\alpha_\mf\approx(1+R_\mf^{-2})/D_\mf$ (see~\autoref{app:MCT} for details).
    \textbf{b}, Capacity contour plot of the example from~\autoref{fig:capacity}b.
    \textbf{c}, Normalized manifold geometry dynamics plot of VGG-11 trained with CIfAR-10. The values in each row are rescaled so that the max value is 1 and the min value is 0.
    }
    \label{fig:learning geometry}
\end{figure}

From~\autoref{fig:learning geometry}a,b we observe distinct stages of manifold geometry evolution during training in 2-layer networks. In the very rich regime, the network initially compresses both radius and dimension, then increases radius to further reduce dimension.
In~\autoref{fig:learning geometry}c, we examine a standard setting where VGG-11 is trained on CIfAR-10. Despite the rapid saturation of training and test accuracy, at least four stages of geometric changes are evident (see~\autoref{fig:method}c for analytical connections between geometric measures and capacity): a \textit{clustering stage} (initial manifold compression), followed by a \textit{structuring stage} (increasing alignment), a \textit{separating stage} (decreasing alignment to push manifolds apart), and a final \textit{stabilizing stage} (further reducing center alignment).

\section{Applications to Neuroscience and Machine Learning Problems}\label{sec:applications}
In previous sections, we used capacity to quantify the degree of feature learning and delineate the learning stages and strategies through effective geometry. In this section, we apply our framework to find geometric insights in problems from neuroscience and machine learning.

\subsection{Structural inductive biases in neural circuits}\label{subsec:rnn}

\begin{figure}[ht]
    \centering
    \includegraphics[width=\linewidth]{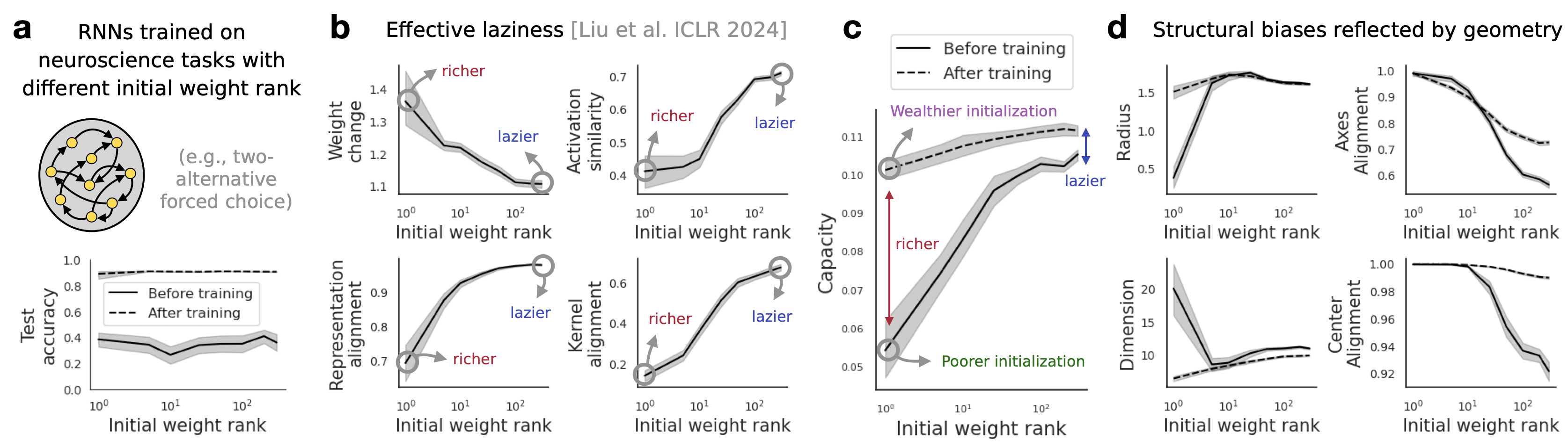}
    \caption{Structural inductive biases in neural circuits.
    \textbf{a}, We consider RNNs trained on standard neuroscience tasks.
    \textbf{b}, Previous work~\citep{liu2024connectivity} found that the initial weight rank of the recurrent connectivity matrix leads to an inductive bias toward effectively richer or lazier training.
    \textbf{c}, We find that RNNs trained with different initial weight rank reach the same capacity value at final epoch. It is the difference in capacity at initialization that makes RNNs with small initial weight rank richer in training.
    \textbf{d}, Despite having the same capacity at final epoch, RNNs with different initial weight rank exhibit different manifold geometry.
    }
    \label{fig:RNN}
\end{figure}

We study recurrent neural networks (RNNs) that are trained on standard neuroscience tasks such as perceptual decision making ~\citep{britten1992} (\autoref{fig:RNN}a). We adopt the setting from previous work~\citep{liu2024connectivity} on investigating how differences in connectivity initialization affect the learning process. In particular, previous work used the weight changes of RNNs before and after training as a measure to quantify if a network is in rich or lazy training regimes (\autoref{fig:RNN}b). Here, we use our methods of capacity and its effective geometry to study such structural biases of neural circuits in a data-driven way (i.e., from neural activity patterns instead of weight matrix).

\paragraph{Experimental setup.}
We use the \texttt{neurogym} package~\citep{neurogym} to simulate common cognitive tasks, including perceptual decision making, delayed matching, etc. To study how connectivity structure impacts learning strategies, we initialize recurrent neural networks (RNN) weights with varying ranks (low-rank weight has lower connectivity and higher initial bias and vice versa) via Singular Value Decomposition (similar setup used in ~\citep{liu2024connectivity}). The RNN have 300 hidden units, 1 layer, with ReLU activations, and are trained for 10000 iterations using \texttt{SGD} optimizer. 
(more details can be found in the Appendix section ~\ref{app:rnn}). 
Manifold capacity and effective geometric measures are computed using representations from the hidden states.

\paragraph{Our findings.}
First, we study the training dynamics of capacity value in RNNs with various initial weight rank (\autoref{fig:RNN}c). In agreement with the previous finding in~\citep{liu2024connectivity} using weight changes, we find that the capacity changes of the small initial weight rank RNNs are higher than those of the large initial weight rank RNNs. Interestingly, the capacity values at the final epoch are about the same for RNNs with different initial weight rank. It is the difference in capacity value at initialization that distinguishes the learning dynamics of RNNs with different initial weight rank. Namely, small initial weight rank RNNs are in the poorer-richer feature learning regime, while large initial weight rank RNNs are in the wealthier-lazier feature learning (\autoref{fig:RNN}c).

Next, although the capacity values of RNNs at the final epoch are about the same for different initial weight ranks, we find that their geometric organizations are quite different (\autoref{fig:RNN}d). For example, poorer-richer learning (i.e., small initial weight rank) ends up with a larger radius but smaller dimension, while it is the opposite for wealthier-lazier learning (i.e., large initial weight rank). This finding suggests that there are structural biases in RNNs at the manifold geometry level. 

\paragraph{Takeaways.}
Conventional methods for studying rich versus lazy learning may only quantify the relative improvement of task-relevant features and overlook the potential difference due to the absolute encoding capacity of different initialization configurations (e.g, low-rank vs high-rank initialization). Our method of using manifold capacity and GLUE is able to overcome this limitation and provide a wide range of geometric signatures to investigate the structural inductive biases of learning in neural circuits.

\subsection{Out-of-distribution generalization}\label{subsec:ood_few_shot}
Out-of-distribution (OOD) generalization refers to the scenario when the training distribution $(\bx,y)\sim\cD_{\text{train}}$ is different from the test distribution $(\bx,y)\sim\cD_{\text{test}}$. Here we focus on the case where the label set in $\cD_{\text{test}}$ is different from that in $\cD_{\text{train}}$. 

\begin{figure}[ht]
    \centering
    \includegraphics[width=\textwidth]{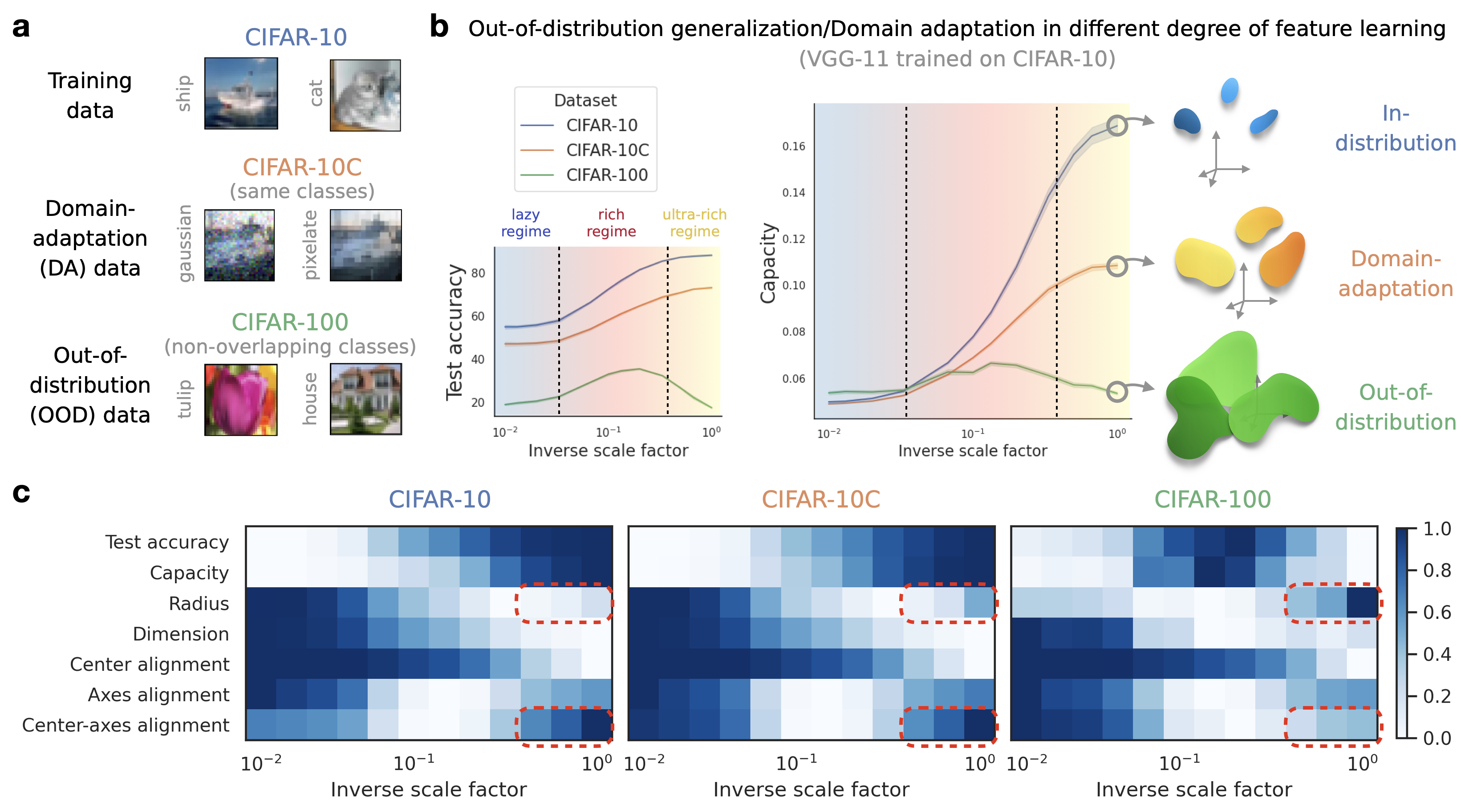}
    \caption{Out of distribution generalization.
    \textbf{a}, CIfAR-10c as a domain adaptation (DA) dataset and CIfAR-100 as an OOD dataset.
    \textbf{b}, Test accuracy improves for CIfAR-10 and CIfAR-10C as the training becomes richer and richer while the linear probe accuracy for CIfAR-100 would drastically drop in the ultra-rich training regime.
    \textbf{c}, Effective manifold geometry of CIfAR-100 reveals that the expansion of manifold radius and the increase of center-axis alignment explain the failure of OOD generalization in the ultra-rich regime. The color is normalized for each row respectively.
    }
    \label{fig:OOD}
\end{figure}

\paragraph{Experimental setup.}
For each model pre-trained on CIfAR-10, we train a linear classifier (i.e., linear probe~\citep{alain2016understanding}) on top of the last-layer representation with CIfAR-100 train set, and then evaluate the linear probe's performance on CIfAR-100 test set (see more details in Appendix~\ref{app:OOD}). We also consider a corrupted version of CIfAR-10, the CIfAR-10C dataset~\citep{hendrycks2018benchmarking} as an example of domain adaptation (DA) task. Finally, we compute the manifold capacity and effective geometric measures on these last-layer representations.

\paragraph{Our findings.}
We see that the test accuracy of the OOD dataset increases when the network enters the rich learning regime ($\bar{\eta}$ around $0.1$) but decreases drastically when the degree of feature learning is too rich ($\bar{\eta}$ around $1.0$). The failure in such \textit{ultra-rich} feature learning regime is different from the test accuracy of both CIfAR-10 and CIfAR-10C (~\autoref{fig:OOD}b). Looking at the capacity and effective geometry (~\autoref{fig:OOD}c), we first see strong correlations between the capacity and test accuracy, which warrants the use of effective geometry. Next, we find that the expansion of manifold radius and the increase of center-axis alignment in the ultra-rich regime explain the drop of capacity. Interestingly, we also see an architectural difference where it is the increment in dimension in the ultra-rich regime explaining the drop of capacity in ResNet-18 (\autoref{fig:app_ood_resnet}).

\paragraph{Takeaways.}
Our method of using manifold capacity and GLUE is able to identify correlations between the geometric signatures of the in-distribution manifolds and that of the OOD manifolds. This may lead to potential applications in mitigating the failure of OOD generalization by merely inspecting the training data.
We leave it as an interesting future direction to extend our study, applying these geometric insights to improve OOD generalization performance.

\section{Conclusion and Discussion}\label{sec:discussion}
Feature learning serves as a crucial \textit{feature} in the study of neural networks in both computational neuroscience and machine learning, and it is much \textit{richer} than the lazy versus rich dichotomy. Understanding the connection between feature learning and performance further promises the future design of network architectures and learning algorithms with enhanced reliability and the requisite model transparency for practical applications.

The primary contribution of this work is to demonstrate how the perspective of task-relevant manifold untangling (quantified by manifold capacity and delineated by manifold geometric measures) can enhance our understanding of feature learning at an intermediate level.
We propose several promising future directions, including extending the theoretical analysis to more realistic settings, exploring applications in other types of DNN (e.g., recurrent networks, transformers) and addressing relevant scientific inquiries in neuroscience, such as inferring plasticity mechanisms from observed learning dynamics in neural data, and predicting learning-induced changes across brain regions. 
We believe that investigations in these intermediate-level understandings can be leveraged to design more robust, generalizable, and safer deep neural networks, as well as more accurate models for neuroscience applications.


\subsubsection*{Acknowledgments}
We thank the members of Chung lab for the discussion regarding many preliminary results and early versions of the manuscript.
This work was supported by the Center for Computational Neuroscience at the Flatiron Institute, Simons Foundation. S.C. was partially supported by a Sloan Research Fellowship, a Klingenstein-Simons Award, and the Samsung Advanced Institute of Technology project, ``Next Generation Deep Learning: From Pattern Recognition to AI.'' All experiments were performed using the Flatiron Institute's high-performance computing cluster.

\bibliography{main}

\begin{thebibliography}{}

\bibitem[Alain and Bengio, 2016]{alain2016understanding}
Alain, G. and Bengio, Y. (2016).
\newblock Understanding intermediate layers using linear classifier probes.
\newblock {\em arXiv preprint arXiv:1610.01644}.

\bibitem[Allen-Zhu et~al., 2019]{allen2019learning}
Allen-Zhu, Z., Li, Y., and Liang, Y. (2019).
\newblock Learning and generalization in overparameterized neural networks, going beyond two layers.
\newblock {\em Advances in neural information processing systems}, 32.

\bibitem[Amit et~al., 1987]{amit1987statistical}
Amit, D.~J., Gutfreund, H., and Sompolinsky, H. (1987).
\newblock Statistical mechanics of neural networks near saturation.
\newblock {\em Annals of physics}, 173(1):30--67.

\bibitem[Ansuini et~al., 2019]{ansuini2019intrinsic}
Ansuini, A., Laio, A., Macke, J.~H., and Zoccolan, D. (2019).
\newblock Intrinsic dimension of data representations in deep neural networks.
\newblock {\em Advances in Neural Information Processing Systems}, 32.

\bibitem[Arora et~al., 2019]{arora2019fine}
Arora, S., Du, S., Hu, W., Li, Z., and Wang, R. (2019).
\newblock Fine-grained analysis of optimization and generalization for overparameterized two-layer neural networks.
\newblock In {\em International Conference on Machine Learning}, pages 322--332. PMLR.

\bibitem[Ba et~al., 2022]{ba2022high}
Ba, J., Erdogdu, M.~A., Suzuki, T., Wang, Z., Wu, D., and Yang, G. (2022).
\newblock High-dimensional asymptotics of feature learning: How one gradient step improves the representation.
\newblock {\em Advances in Neural Information Processing Systems}, 35:37932--37946.

\bibitem[Bahri et~al., 2021]{bahri2021explaining}
Bahri, Y., Dyer, E., Kaplan, J., Lee, J., and Sharma, U. (2021).
\newblock Explaining neural scaling laws.
\newblock {\em arXiv preprint arXiv:2102.06701}.

\bibitem[Belinkov et~al., 2017]{belinkov2017}
Belinkov, Y., Durrani, N., Dalvi, F., Sajjad, H., and Glass, J. (2017).
\newblock What do neural machine translation models learn about morphology?
\newblock In Barzilay, R. and Kan, M.-Y., editors, {\em Proceedings of the 55th Annual Meeting of the Association for Computational Linguistics (Volume 1: Long Papers)}, pages 861--872, Vancouver, Canada. Association for Computational Linguistics.

\bibitem[Belkin et~al., 2019]{belkin2019reconciling}
Belkin, M., Hsu, D., Ma, S., and Mandal, S. (2019).
\newblock Reconciling modern machine-learning practice and the classical bias--variance trade-off.
\newblock {\em Proceedings of the National Academy of Sciences}, 116(32):15849--15854.

\bibitem[Bernardi et~al., 2020]{bernardi2020geometry}
Bernardi, S., Benna, M.~K., Rigotti, M., Munuera, J., Fusi, S., and Salzman, C.~D. (2020).
\newblock The geometry of abstraction in the hippocampus and prefrontal cortex.
\newblock {\em Cell}, 183(4):954--967.

\bibitem[Bordelon et~al., 2020]{bordelon2020spectrum}
Bordelon, B., Canatar, A., and Pehlevan, C. (2020).
\newblock Spectrum dependent learning curves in kernel regression and wide neural networks.
\newblock In {\em International Conference on Machine Learning}, pages 1024--1034. PMLR.

\bibitem[Bordelon and Pehlevan, 2022]{bordelon2022influence}
Bordelon, B. and Pehlevan, C. (2022).
\newblock The influence of learning rule on representation dynamics in wide neural networks.
\newblock In {\em The Eleventh International Conference on Learning Representations}.

\bibitem[Britten et~al., 1992]{britten1992}
Britten, K., Shadlen, M., Newsome, W., and Movshon, J. (1992).
\newblock The analysis of visual motion: a comparison of neuronal and psychophysical performance.
\newblock {\em Journal of Neuroscience}.

\bibitem[Canatar et~al., 2021]{canatar2021spectral}
Canatar, A., Bordelon, B., and Pehlevan, C. (2021).
\newblock Spectral bias and task-model alignment explain generalization in kernel regression and infinitely wide neural networks.
\newblock {\em Nature communications}, 12(1):2914.

\bibitem[Cao et~al., 2020]{cao2020characterizing}
Cao, Y., Summerfield, C., and Saxe, A. (2020).
\newblock Characterizing emergent representations in a space of candidate learning rules for deep networks.
\newblock {\em Advances in Neural Information Processing Systems}, 33:8660--8670.

\bibitem[Chizat et~al., 2019]{chizat2019lazy}
Chizat, L., Oyallon, E., and Bach, F. (2019).
\newblock On lazy training in differentiable programming.
\newblock {\em Advances in neural information processing systems}, 32.

\bibitem[Chou et~al., 2025]{chou2025glue}
Chou, C.-N., Kim, R., Arend, L., Yang, Y.-Y., Mensh, B., Shim, W.~M., Perich, M., and Chung, S. (2025).
\newblock Geometry linked to untangling efficiency reveals structure and computation in neural populations.
\newblock {\em bioRxiv}.

\bibitem[Chung and Abbott, 2021]{chung2021neural}
Chung, S. and Abbott, L.~F. (2021).
\newblock Neural population geometry: An approach for understanding biological and artificial neural networks.
\newblock {\em Current opinion in neurobiology}, 70:137--144.

\bibitem[Chung et~al., 2018]{chung2018classification}
Chung, S., Lee, D.~D., and Sompolinsky, H. (2018).
\newblock Classification and geometry of general perceptual manifolds.
\newblock {\em Physical Review X}.

\bibitem[Cohen et~al., 2020]{cohen2020separability}
Cohen, U., Chung, S., Lee, D.~D., and Sompolinsky, H. (2020).
\newblock Separability and geometry of object manifolds in deep neural networks.
\newblock {\em Nature communications}.

\bibitem[Cover, 1965]{cover1965geometrical}
Cover, T.~M. (1965).
\newblock Geometrical and statistical properties of systems of linear inequalities with applications in pattern recognition.
\newblock {\em IEEE transactions on electronic computers}, pages 326--334.

\bibitem[Damian et~al., 2022]{damian2022neural}
Damian, A., Lee, J., and Soltanolkotabi, M. (2022).
\newblock Neural networks can learn representations with gradient descent.
\newblock In {\em Conference on Learning Theory}, pages 5413--5452. PMLR.

\bibitem[Dandi et~al., 2023]{dandi2023two}
Dandi, Y., Krzakala, F., Loureiro, B., Pesce, L., and Stephan, L. (2023).
\newblock How two-layer neural networks learn, one (giant) step at a time.
\newblock In {\em NeurIPS 2023 Workshop on Mathematics of Modern Machine Learning}.

\bibitem[DiCarlo and Cox, 2007]{dicarlo2007untangling}
DiCarlo, J.~J. and Cox, D.~D. (2007).
\newblock Untangling invariant object recognition.
\newblock {\em Trends in cognitive sciences}, 11(8):333--341.

\bibitem[Du et~al., 2019]{du2019gradient}
Du, S., Lee, J., Li, H., Wang, L., and Zhai, X. (2019).
\newblock Gradient descent finds global minima of deep neural networks.
\newblock In {\em International conference on machine learning}, pages 1675--1685. PMLR.

\bibitem[Du et~al., 2018]{du2018gradient}
Du, S.~S., Zhai, X., Poczos, B., and Singh, A. (2018).
\newblock Gradient descent provably optimizes over-parameterized neural networks.
\newblock {\em arXiv preprint arXiv:1810.02054}.

\bibitem[Ehrlich et~al., 2021]{ehrlich2021psychrnn}
Ehrlich, D.~B., Stone, J.~T., Brandfonbrener, D., Atanasov, A., and Murray, J.~D. (2021).
\newblock Psychrnn: An accessible and flexible python package for training recurrent neural network models on cognitive tasks.
\newblock {\em eneuro}, 8(1).

\bibitem[Engstrom et~al., 2019]{engstrom2019adversarial}
Engstrom, L., Ilyas, A., Santurkar, S., Tsipras, D., Tran, B., and Madry, A. (2019).
\newblock Adversarial robustness as a prior for learned representations.
\newblock {\em arXiv preprint arXiv:1906.00945}.

\bibitem[Erhan et~al., 2009]{pascal2009visualize}
Erhan, D., Bengio, Y., Courville, A., and Vincent, P. (2009).
\newblock Visualizing higher-layer features of a deep network.
\newblock {\em Technical Report, Univeristé de Montréal}.

\bibitem[Flesch et~al., 2022]{flesch2022orthogonal}
Flesch, T., Juechems, K., Dumbalska, T., Saxe, A., and Summerfield, C. (2022).
\newblock Orthogonal representations for robust context-dependent task performance in brains and neural networks.
\newblock {\em Neuron}, 110(7):1258--1270.

\bibitem[Gao and Ganguli, 2015]{gao2015simplicity}
Gao, P. and Ganguli, S. (2015).
\newblock On simplicity and complexity in the brave new world of large-scale neuroscience.
\newblock {\em Current opinion in neurobiology}, 32:148--155.

\bibitem[Gardner, 1988]{gardner1988space}
Gardner, E. (1988).
\newblock The space of interactions in neural network models.
\newblock {\em Journal of physics A: Mathematical and general}, 21(1):257.

\bibitem[Gardner and Derrida, 1988]{gardner1988optimal}
Gardner, E. and Derrida, B. (1988).
\newblock Optimal storage properties of neural network models.
\newblock {\em Journal of Physics A: Mathematical and general}, 21(1):271.

\bibitem[Geiger et~al., 2020]{geiger2020disentangling}
Geiger, M., Spigler, S., Jacot, A., and Wyart, M. (2020).
\newblock Disentangling feature and lazy training in deep neural networks.
\newblock {\em Journal of Statistical Mechanics: Theory and Experiment}, 2020(11):113301.

\bibitem[Ghosh et~al., 2022]{ghosh2022investigating}
Ghosh, A., Mondal, A.~K., Agrawal, K.~K., and Richards, B. (2022).
\newblock Investigating power laws in deep representation learning.
\newblock {\em arXiv preprint arXiv:2202.05808}.

\bibitem[Girshick et~al., 2014]{girshick2014rich}
Girshick, R., Donahue, J., Darrell, T., and Malik, J. (2014).
\newblock Rich feature hierarchies for accurate object detection and semantic segmentation.
\newblock In {\em Proceedings of the IEEE conference on computer vision and pattern recognition}, pages 580--587.

\bibitem[Glorot and Bengio, 2010]{glorot2010understanding}
Glorot, X. and Bengio, Y. (2010).
\newblock Understanding the difficulty of training deep feedforward neural networks.
\newblock In {\em Proceedings of the thirteenth international conference on artificial intelligence and statistics}, pages 249--256. JMLR Workshop and Conference Proceedings.

\bibitem[Gurnani and Gajic, 2023]{gurnani2023signatures}
Gurnani, H. and Gajic, N. A.~C. (2023).
\newblock Signatures of task learning in neural representations.
\newblock {\em Current opinion in neurobiology}, 83:102759.

\bibitem[He et~al., 2016]{he2016deep}
He, K., Zhang, X., Ren, S., and Sun, J. (2016).
\newblock Deep residual learning for image recognition.
\newblock In {\em Proceedings of the IEEE conference on computer vision and pattern recognition}, pages 770--778.

\bibitem[Hendrycks and Dietterich, 2018]{hendrycks2018benchmarking}
Hendrycks, D. and Dietterich, T. (2018).
\newblock Benchmarking neural network robustness to common corruptions and perturbations.
\newblock In {\em International Conference on Learning Representations}.

\bibitem[Hofmann et~al., 2008]{hofmann2008kernel}
Hofmann, T., Sch{\"o}lkopf, B., and Smola, A.~J. (2008).
\newblock Kernel methods in machine learning.
\newblock {\em The Annals of Statistics}, 36(3):1171--1220.

\bibitem[Hong et~al., 2016]{hong2016explicit}
Hong, H., Yamins, D.~L., Majaj, N.~J., and DiCarlo, J.~J. (2016).
\newblock Explicit information for category-orthogonal object properties increases along the ventral stream.
\newblock {\em Nature neuroscience}, 19(4):613--622.

\bibitem[Hopfield, 1982]{hopfield1982neural}
Hopfield, J.~J. (1982).
\newblock Neural networks and physical systems with emergent collective computational abilities.
\newblock {\em Proceedings of the national academy of sciences}, 79(8):2554--2558.

\bibitem[Hu et~al., 2024]{hu2024representational}
Hu, B., Temiz, N.~Z., Chou, C.-N., Rupprecht, P., Meissner-Bernard, C., Titze, B., Chung, S., and Friedrich, R.~W. (2024).
\newblock Representational learning by optimization of neural manifolds in an olfactory memory network.
\newblock {\em bioRxiv}, pages 2024--11.

\bibitem[Jacot et~al., 2018]{jacot2018neural}
Jacot, A., Gabriel, F., and Hongler, C. (2018).
\newblock Neural tangent kernel: Convergence and generalization in neural networks.
\newblock {\em Advances in neural information processing systems}, 31.

\bibitem[Karp et~al., 2021]{karp2021local}
Karp, S., Winston, E., Li, Y., and Singh, A. (2021).
\newblock Local signal adaptivity: Provable feature learning in neural networks beyond kernels.
\newblock {\em Advances in Neural Information Processing Systems}, 34:24883--24897.

\bibitem[Kirkpatrick et~al., 2017]{kirkpatrick2017overcoming}
Kirkpatrick, J., Pascanu, R., Rabinowitz, N., Veness, J., Desjardins, G., Rusu, A.~A., Milan, K., Quan, J., Ramalho, T., Grabska-Barwinska, A., et~al. (2017).
\newblock Overcoming catastrophic forgetting in neural networks.
\newblock {\em Proceedings of the national academy of sciences}, 114(13):3521--3526.

\bibitem[Kirsanov et~al., 2025]{kirsanov2025}
Kirsanov, A., Chou, C.-N., Cho, K., and Chung, S. (2025).
\newblock The geometry of prompting: Unveiling distinct mechanisms of task adaptation in language models.
\newblock In {\em Findings of the Association for Computational Linguistics: NAACL 2025}, pages 1855--1888. Association for Computational Linguistics.

\bibitem[Kornblith et~al., 2019]{kornblith2019similarity}
Kornblith, S., Norouzi, M., Lee, H., and Hinton, G. (2019).
\newblock Similarity of neural network representations revisited.

\bibitem[Kriegeskorte and Kievit, 2013]{kriegeskorte2013representational}
Kriegeskorte, N. and Kievit, R.~A. (2013).
\newblock Representational geometry: integrating cognition, computation, and the brain.
\newblock {\em Trends in cognitive sciences}, 17(8):401--412.

\bibitem[Krizhevsky and Hinton, 2009]{cifardataset}
Krizhevsky, A. and Hinton, G. (2009).
\newblock Learning multiple layers of features from tiny images.
\newblock Technical report, University of Toronto, Toronto, Ontario.

\bibitem[Krizhevsky et~al., 2012]{krizhevsky2012imagenet}
Krizhevsky, A., Sutskever, I., and Hinton, G.~E. (2012).
\newblock Imagenet classification with deep convolutional neural networks.
\newblock {\em Advances in neural information processing systems}, 25.

\bibitem[Kumar et~al., 2024]{kumargrokking}
Kumar, T., Bordelon, B., Gershman, S.~J., and Pehlevan, C. (2024).
\newblock Grokking as the transition from lazy to rich training dynamics.
\newblock In {\em The Twelfth International Conference on Learning Representations}.

\bibitem[Kuoch et~al., 2024]{kuoch2024probing}
Kuoch, M., Chou, C.-N., Parthasarathy, N., Dapello, J., DiCarlo, J.~J., Sompolinsky, H., and Chung, S. (2024).
\newblock Probing biological and artificial neural networks with task-dependent neural manifolds.
\newblock In {\em Conference on Parsimony and Learning (Proceedings Track)}.

\bibitem[LeCun et~al., 1998]{lecun1998gradient}
LeCun, Y., Bottou, L., Bengio, Y., and Haffner, P. (1998).
\newblock Gradient-based learning applied to document recognition.
\newblock {\em Proceedings of the IEEE}, 86(11):2278--2324.

\bibitem[Lee et~al., 2019]{lee2019wide}
Lee, J., Xiao, L., Schoenholz, S., Bahri, Y., Novak, R., Sohl-Dickstein, J., and Pennington, J. (2019).
\newblock Wide neural networks of any depth evolve as linear models under gradient descent.
\newblock {\em Advances in neural information processing systems}, 32.

\bibitem[Liu et~al., 2024]{liu2024connectivity}
Liu, Y.~H., Baratin, A., Cornford, J., Mihalas, S., SheaBrown, E.~T., and Lajoie, G. (2024).
\newblock How connectivity structure shapes rich and lazy learning in neural circuits.
\newblock In {\em The Twelfth International Conference on Learning Representations}.

\bibitem[Liu et~al., 2022]{liu2022towards}
Liu, Z., Kitouni, O., Nolte, N.~S., Michaud, E., Tegmark, M., and Williams, M. (2022).
\newblock Towards understanding grokking: An effective theory of representation learning.
\newblock {\em Advances in Neural Information Processing Systems}, 35:34651--34663.

\bibitem[Mante et~al., 2013]{mante2013context}
Mante, V., Sussillo, D., Shenoy, K.~V., and Newsome, W.~T. (2013).
\newblock Context-dependent computation by recurrent dynamics in prefrontal cortex.
\newblock {\em Nature}, 503(7474):78--84.

\bibitem[Mei and Montanari, 2022]{mei2022generalization}
Mei, S. and Montanari, A. (2022).
\newblock The generalization error of random features regression: Precise asymptotics and the double descent curve.
\newblock {\em Communications on Pure and Applied Mathematics}, 75(4):667--766.

\bibitem[Miller et~al., 1996]{miller1996neural}
Miller, E.~K., Erickson, C.~A., and Desimone, R. (1996).
\newblock Neural mechanisms of visual working memory in prefrontal cortex of the macaque.
\newblock {\em Journal of neuroscience}, 16(16):5154--5167.

\bibitem[Molano-Mazon et~al., 2022]{neurogym}
Molano-Mazon, M., Barbosa, J., Pastor-Ciurana, J., Fradera, M., ZHANG, R.-Y., Forest, J., del Pozo~Lerida, J., Ji-An, L., Cueva, C.~J., de~la Rocha, J., and et~al. (2022).
\newblock Neurogym: An open resource for developing and sharing neuroscience tasks.

\bibitem[Montanari et~al., 2019]{montanari2019generalization}
Montanari, A., Ruan, F., Sohn, Y., and Yan, J. (2019).
\newblock The generalization error of max-margin linear classifiers: High-dimensional asymptotics in the overparametrized regime.
\newblock {\em arXiv e-prints}, pages arXiv--1911.

\bibitem[Nakkiran et~al., 2021]{nakkiran2021deep}
Nakkiran, P., Kaplun, G., Bansal, Y., Yang, T., Barak, B., and Sutskever, I. (2021).
\newblock Deep double descent: Where bigger models and more data hurt.
\newblock {\em Journal of Statistical Mechanics: Theory and Experiment}, 2021(12):124003.

\bibitem[Nanda et~al., 2023]{nanda2023progress}
Nanda, N., Chan, L., Lieberum, T., Smith, J., and Steinhardt, J. (2023).
\newblock Progress measures for grokking via mechanistic interpretability.
\newblock {\em arXiv preprint arXiv:2301.05217}.

\bibitem[Nayebi et~al., 2020]{nayebi2020identifying}
Nayebi, A., Srivastava, S., Ganguli, S., and Yamins, D.~L. (2020).
\newblock Identifying learning rules from neural network observables.
\newblock {\em Advances in Neural Information Processing Systems}, 33:2639--2650.

\bibitem[Niv, 2019]{niv2019learning}
Niv, Y. (2019).
\newblock Learning task-state representations.
\newblock {\em Nature neuroscience}, 22(10):1544--1553.

\bibitem[Olah et~al., 2017]{olah2017feature}
Olah, C., Mordvintsev, A., and Schubert, L. (2017).
\newblock Feature visualization.
\newblock {\em Distill}.
\newblock https://distill.pub/2017/feature-visualization.

\bibitem[Olshausen and Field, 1996]{olshausen1996emergence}
Olshausen, B.~A. and Field, D.~J. (1996).
\newblock Emergence of simple-cell receptive field properties by learning a sparse code for natural images.
\newblock {\em Nature}, 381(6583):607--609.

\bibitem[Paraouty et~al., 2023]{paraouty2023sensory}
Paraouty, N., Yao, J.~D., Varnet, L., Chou, C.-N., Chung, S., and Sanes, D.~H. (2023).
\newblock Sensory cortex plasticity supports auditory social learning.
\newblock {\em Nature communications}, 14(1):5828.

\bibitem[Poort et~al., 2015]{poort2015learning}
Poort, J., Khan, A.~G., Pachitariu, M., Nemri, A., Orsolic, I., Krupic, J., Bauza, M., Sahani, M., Keller, G.~B., Mrsic-Flogel, T.~D., et~al. (2015).
\newblock Learning enhances sensory and multiple non-sensory representations in primary visual cortex.
\newblock {\em Neuron}, 86(6):1478--1490.

\bibitem[Power et~al., 2022]{power2022grokking}
Power, A., Burda, Y., Edwards, H., Babuschkin, I., and Misra, V. (2022).
\newblock Grokking: Generalization beyond overfitting on small algorithmic datasets.
\newblock {\em arXiv preprint arXiv:2201.02177}.

\bibitem[Raghu et~al., 2021]{raghu2021vit}
Raghu, M., Unterthiner, T., Kornblith, S., Zhang, C., and Dosovitskiy, A. (2021).
\newblock Do vision transformers see like convolutional neural networks?
\newblock In Ranzato, M., Beygelzimer, A., Dauphin, Y., Liang, P., and Vaughan, J.~W., editors, {\em Advances in Neural Information Processing Systems}, volume~34, pages 12116--12128. Curran Associates, Inc.

\bibitem[Rahaman et~al., 2019]{rahaman2019spectral}
Rahaman, N., Baratin, A., Arpit, D., Draxler, F., Lin, M., Hamprecht, F., Bengio, Y., and Courville, A. (2019).
\newblock On the spectral bias of neural networks.
\newblock In {\em International conference on machine learning}, pages 5301--5310. PMLR.

\bibitem[Rahimi and Recht, 2007]{rahimi2007random}
Rahimi, A. and Recht, B. (2007).
\newblock Random features for large-scale kernel machines.
\newblock {\em Advances in neural information processing systems}, 20.

\bibitem[Reinert et~al., 2021]{reinert2021mouse}
Reinert, S., H{\"u}bener, M., Bonhoeffer, T., and Goltstein, P.~M. (2021).
\newblock Mouse prefrontal cortex represents learned rules for categorization.
\newblock {\em Nature}, 593(7859):411--417.

\bibitem[Selvaraju et~al., 2017]{selvaraju2017grad}
Selvaraju, R.~R., Cogswell, M., Das, A., Vedantam, R., Parikh, D., and Batra, D. (2017).
\newblock Grad-cam: Visual explanations from deep networks via gradient-based localization.
\newblock In {\em Proceedings of the IEEE international conference on computer vision}, pages 618--626.

\bibitem[Shi et~al., 2022]{shi2022theoretical}
Shi, Z., Wei, J., and Liang, Y. (2022).
\newblock A theoretical analysis on feature learning in neural networks: Emergence from inputs and advantage over fixed features.
\newblock {\em arXiv preprint arXiv:2206.01717}.

\bibitem[Simonyan et~al., 2013]{simonyan2013deep}
Simonyan, K., Vedaldi, A., and Zisserman, A. (2013).
\newblock Deep inside convolutional networks: Visualising image classification models and saliency maps.
\newblock {\em arXiv preprint arXiv:1312.6034}.

\bibitem[Simonyan and Zisserman, 2015]{simonyan2015very}
Simonyan, K. and Zisserman, A. (2015).
\newblock Very deep convolutional networks for large-scale image recognition.
\newblock In {\em 3rd International Conference on Learning Representations (ICLR 2015)}. Computational and Biological Learning Society.

\bibitem[Sorscher et~al., 2022]{sorscher2022neural}
Sorscher, B., Ganguli, S., and Sompolinsky, H. (2022).
\newblock Neural representational geometry underlies few-shot concept learning.
\newblock {\em Proceedings of the National Academy of Sciences}, 119(43):e2200800119.

\bibitem[Stephenson et~al., 2021]{stephenson2021geometry}
Stephenson, C., Padhy, S., Ganesh, A., Hui, Y., Tang, H., and Chung, S. (2021).
\newblock On the geometry of generalization and memorization in deep neural networks.
\newblock {\em arXiv preprint arXiv:2105.14602}.

\bibitem[Sundararajan et~al., 2017]{sundararajan2017axiomatic}
Sundararajan, M., Taly, A., and Yan, Q. (2017).
\newblock Axiomatic attribution for deep networks.
\newblock In {\em International conference on machine learning}, pages 3319--3328. PMLR.

\bibitem[Vershynin, 2018]{vershynin2018high}
Vershynin, R. (2018).
\newblock {\em High-dimensional probability: An introduction with applications in data science}, volume~47.
\newblock Cambridge university press.

\bibitem[Wakhloo et~al., 2023]{wakhloo2023linear}
Wakhloo, A.~J., Sussman, T.~J., and Chung, S. (2023).
\newblock Linear classification of neural manifolds with correlated variability.
\newblock {\em Physical Review Letters}.

\bibitem[Woodworth et~al., 2020]{woodworth2020kernel}
Woodworth, B., Gunasekar, S., Lee, J.~D., Moroshko, E., Savarese, P., Golan, I., Soudry, D., and Srebro, N. (2020).
\newblock Kernel and rich regimes in overparametrized models.
\newblock In {\em Conference on Learning Theory}, pages 3635--3673. PMLR.

\bibitem[Yamins and DiCarlo, 2016]{yamins2016using}
Yamins, D.~L. and DiCarlo, J.~J. (2016).
\newblock Using goal-driven deep learning models to understand sensory cortex.
\newblock {\em Nature neuroscience}, 19(3):356--365.

\bibitem[Yang and Hu, 2021]{yang2021tensor}
Yang, G. and Hu, E.~J. (2021).
\newblock Tensor programs iv: Feature learning in infinite-width neural networks.
\newblock In {\em International Conference on Machine Learning}, pages 11727--11737. PMLR.

\bibitem[Yerxa et~al., 2023]{yerxa2023learning}
Yerxa, T., Kuang, Y., Simoncelli, E., and Chung, S. (2023).
\newblock Learning efficient coding of natural images with maximum manifold capacity representations.
\newblock {\em Advances in Neural Information Processing Systems}, 36:24103--24128.

\bibitem[Zeiler and Fergus, 2014]{zeiler2014visualizing}
Zeiler, M.~D. and Fergus, R. (2014).
\newblock Visualizing and understanding convolutional networks.
\newblock In {\em Computer Vision--ECCV 2014: 13th European Conference, Zurich, Switzerland, September 6-12, 2014, Proceedings, Part I 13}, pages 818--833. Springer.

\end{thebibliography}
\bibliographystyle{apalike}

\newpage
\appendix

\section{More on Related Work}\label{app:related work}

\paragraph{Visualization.}
Due to the black-box and complex nature of deep neural networks, various visualization techniques have been developed to attempt to characterize the features that models learn during training (\textit{feature visualization}) and identify which input pixel and / or feature activation in the hidden layers contribute significantly to the final model outputs (\textit{feature attribution}). \textit{Feature visualization} techniques visualize features (e.g convolutional filter in the case of CNNs) by generating the input sample that maximizes the activation of that given feature via gradient descent~\citep{olah2017feature}~\citep{pascal2009visualize}~\citep{zeiler2014visualizing}. With its vivid visualization, \textit{feature visualization} provide good intuition about the qualitative characteristics of the features that DNNs learn across layers~\citep{zeiler2014visualizing} as well as different types of models (e.g, standard vs adversarially robust ~\citep{engstrom2019adversarial}). \textit{Feature attribution} techniques generally identify how much each input and/or hidden features contribute to the final model prediction by computing the gradient of that input/hidden features to the output (some example techniques include saliency map~\citep{simonyan2013deep}, Grad-cam~\citep{selvaraju2017grad}, integrated gradient~\citep{sundararajan2017axiomatic}). Although both \textit{feature visualization} and \textit{feature attribution} offer intuitive understanding about the model's feature characteristics, the qualitative nature of visualization makes it difficult to quantify the degree of relevance of the learned features to a given task.

\paragraph{Kernel dynamics.}
Kernel methods~\citep{hofmann2008kernel} have been classic machine learning techniques, where the primary goal is to design an effective embedding that maps inputs to a feature space, thus facilitating efficient algorithms to find good solutions (e.g., linear classifier). While neural networks are inherently complex, seminal works~\citep{jacot2018neural,lee2019wide} have shown that in the infinite width limit, a network can be linearized by its \textit{neural tangent kernel (NTK)}. Thus, studying the NTK of a network allows an analytical understanding of various properties of neural networks, such as convergence to global minima~\citep{du2018gradient,du2019gradient}, generalization performance~\citep{allen2019learning,arora2019fine}, implicit bias~\citep{bordelon2020spectrum,canatar2021spectral}, and neural scaling laws~\citep{bahri2021explaining}. 

When a network is properly initialized~\citep{chizat2019lazy}, gradient descent can converge to the NTK of the random initialization, a setting known as the \textit{kernel regime} (a.k.a., \textit{lazy training} or \textit{random feature regime}). On the other hand, a network can also enter what is known as the \textit{feature learning regime} (a.k.a., \textit{rich training} or \textit{mean-field limit}), where it deviates from the NTK of the initialization~\citep{geiger2020disentangling}. Extensive research has been conducted to characterize lazy versus rich regimes~\citep{geiger2020disentangling,woodworth2020kernel} and to demonstrate instances where feature learning outperforms lazy training~\citep{yang2021tensor,ba2022high,dandi2023two}. It is important to note that even when a network undergoes feature learning, the NTK can still be defined at each epoch. Previous works also analytically characterized the dynamics of kernel in simpler models~\citep{bordelon2020spectrum}. Studying such kernel dynamics also provides a lens for exploring questions related to feature learning, such as grokking~\citep{kumargrokking}.

\paragraph{Representational geometry.}
The visualization approaches mentioned above focus on studying the geometric properties of the feature map itself. Another fruitful direction is to examine the geometric properties of the neural representations of inputs (i.e., embedding vectors) and their connections to performance~\citep{chung2021neural,gurnani2023signatures}. Various dimensionality reduction methods (e.g., principal components analysis (PCA), Isomap, t-SNE, MDS, and UMAP) have been proposed to build intuitions about the organization of high-dimensional feature spaces. In addition, there are approaches that study lower-order statistics of embedding vectors, such as representational similarity~\citep{kriegeskorte2013representational} and spectral methods~\citep{rahaman2019spectral,bahri2021explaining,ghosh2022investigating}. Methods for extracting higher-level geometric properties (e.g., dimension) have also been proposed~\citep{chung2018classification,cohen2020separability,chou2025glue,ansuini2019intrinsic}, with wide applications in both machine learning (e.g., memorization~\citep{stephenson2021geometry}, grokking of modular arithmetic~\citep{liu2022towards,nanda2023progress}, in-context learning in LLM~\citep{kirsanov2025}, self-supervised learning~\citep{yerxa2023learning,kuoch2024probing}), and neuroscience (e.g., perceptual untangling in object categorization~\citep{chung2018classification}, olfactory memory~\citep{hu2024representational}, abstraction~\citep{bernardi2020geometry}, few-shot learning~\citep{sorscher2022neural}, social learning~\citep{paraouty2023sensory}).

\subsection{Previous work on storage capacity}
Storage capacity is defined as the information load for linear readouts and has been studied in several communities, including learning theory~\citep{cover1965geometrical} and statistical physics of neural networks~\citep{gardner1988optimal,gardner1988space}. To enable a mathematical treatment, we focus on the proportional limit (a.k.a. the high-dimensional limit, the thermodynamic limit), i.e., $N,P\to\infty$ and $\lim_{N,P\to\infty} N/P=O(1)$. For a given network and input data, we denote the representation of the $i$-th input $x_i$ as $\Phi(x_i)\in\Real^N$ where $\Phi$ is the (non-linear) feature map. The storage capacity of $\Phi$ is defined as.
\begin{equation}\label{def:storage capacity}
\alpha(\Phi) := \lim_{N\to\infty}\max_{P}\left\{ \frac{P}{N}\, :\, \Pr_{\by}\left[\exists\theta\in\Real^N,\ \forall i\in[P],\ y_i\langle\theta,\Phi(x_i)\rangle\geq0\right]\geq1-o_N(1) \right\}
\end{equation}
where $\by\in\{\pm1\}^P$ is uniformly random sampled, $\theta$ is the linear classifier, and $o_N(1)$ denotes vanishing terms (i.e., $o_N(1)\to0$ as $N\to\infty$). One can also consider the setting where the distribution of $\by$ is biased toward some task direction~\citep{montanari2019generalization}. Intuitively, $\alpha(\Phi)$ quantifies the number of patterns per neuron that a network can store and decode with linear readouts.

\begin{figure}[ht]
    \centering
    \includegraphics[width=12cm]{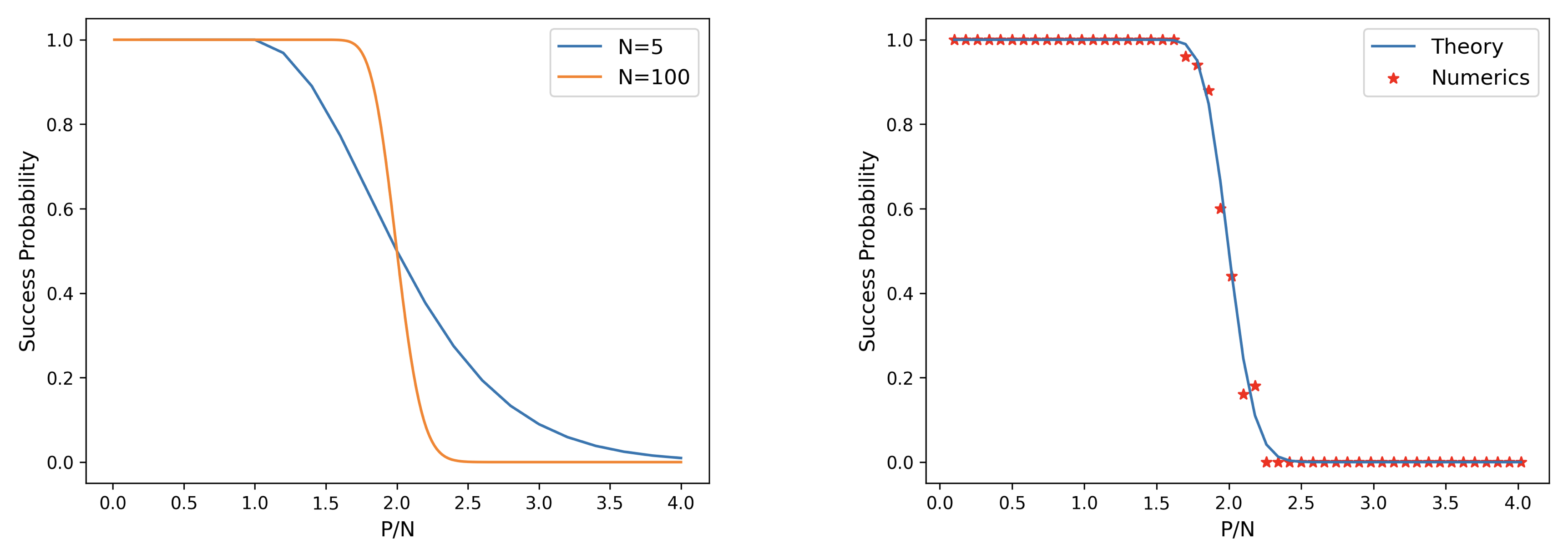}
    \caption{Storage capacity of random points and labels. Storage capacity is defined as the critical ration $P/N=2$ where the success probability undergoes a phase transition. Left: finite size success probability curves proved in Cover's theorem. Right: a numerical check for Cover's theorem.}
    \label{fig:cover}
\end{figure}

Recall that storage capacity is defined as the critical ratio between the number of stored patterns and the number of neurons (\cref{def:storage capacity}). Cover's theorem~\citep{cover1965geometrical} shows that the success probability of having a linear classifier for $P$ points with random binary labels in general position~\footnote{Meaning that every $N'\leq N$ points are linearly independent. Note that random points are in general position with probability $1-o(1)$.} is $p(N,P)=2^{1-P}\sum_{k=0}^{N-1}\binom{P-1}{k}$. In particular, for $P/N<2$ we have $\lim_{N\to\infty}p(N,P)=0$ and for $P/N>2$ we have $\lim_{N\to\infty}p(N,P)=1$. Namely, the storage capacity of points in general position with random binary label is $2$. See also~\autoref{fig:cover} for finite-size and numerical examples.

In the seminal works of Gardner and Derrida~\citep{gardner1988optimal,gardner1988space}, the storage capacity for random points with non-zero margin is analytically characterized using replica method. In the context of associative memory, the storage capacity of Hopfield networks~\citep{hopfield1982neural} is calculated by~\citep{amit1987statistical}.

\section{Manifold Capacity Theory and Effective Geometry}\label{app:MCT}
Manifold capacity  theory (MCT)\citep{chung2018classification,chung2021neural,wakhloo2023linear,chou2025glue} was originally developed for the study of manifold untangling~\citep{dicarlo2007untangling} in theoretical/computational neuroscience. Intuitively, manifold untangling refers to the increased separation of high-dimensional manifolds (e.g., point cloud manifolds) in the eyes of a downstream readout. MCT quantifies this intuition via modeling a downstream neuron as a linear classifier, and uses the \textit{packing efficiency} of the neural representational space to evaluate the degree of manifold untangling. Mathematically, such packing efficiency coincides with support vector machine (SVM) in an average-case setting.

\subsection{Neural manifolds as convex hulls of pre-readout representations}
As we are studying feature learning, we are interested in the neural representations that correspond to activations obtained from the pre-linear readout layer neurons. The readers can refer to~\autoref{app:2-layer} and~\autoref{app:dnn} for details on activation extraction. Notation wise, let $N$ be the number of neurons. Therefore, all neural representations live in $\Real^{N}$ space. 

Next, we group neural representations by their category labels assigned during training to obtain $P$ data manifolds. For $i\in\{1,\dots,P\}$, the $i$-th data manifold, denoted as $\cM_i$, is a convex set in $\Real^{N}$. To ensure convexity in practice, we take $M_i$ to be the convex hull of a collection of vectors $\cM_i=\{\bx_1^i,\dots,\bx_{M_{i}}^i\}$ where $M_{i}$ is the number of points in the $i$-th manifold. 

Notice that the each data manifold lives in its own subspace of dimension $D_i\leq N$. Therefore, we can rewrite each data manifold in its own coordinate system:
\begin{equation}\label{eq:manifold def}
\cM_i = \left\{\bu^i_0 + \sum_{j=1}^{D_i}s_j\bu^i_j\ \Bigg|\ \bs=(s_1,\dots,s_{D_i})\in\cS_i\right\}
\end{equation}
Here, $\bu_0^i$ is the center of the $i$-th manifold and $\{\bu_j^i\}_{j=1}^{D_i}$ is an orthonormal basis. The shape set $\cS_i\subset\Real^{D_i}$ is a convex set denoting coordinates of the manifold points in its subspace. In practice, the manifold axes and shape sets $\cS_i$ are completely data driven.

\subsection{A simulation definition for manifold capacity}
Recall from~\autoref{sec:method} that the simulation version of manifold capacity is defined as follows.
\simcapdef*
Intuitively, the simulated manifold capacity measures the \textit{packability}~\citep{chung2018classification} of manifolds by determining the smallest dimensional subspace needed to ensure they can be separated. Namely, manifolds that are more packable\footnote{The reason why this is called ``packing'' is that projecting manifolds into smaller dimensional subspace is like packing them into a smaller neural representational space.} (i.e., separable when projected to smaller dimensional subspaces) exhibit higher manifold capacity. Note that the simulated capacity can be estimated from data by empirically estimate $p_n$ and perform binary search to find the critical dimension $\min_{p_n\geq0.5}\{n\}$. This procedure is computationally expensive and requires some choices of hyperparameters (which makes the definition a little ad hoc). Nevertheless,~\cref{def:capacity simulated} provides good intuition on how to think about manifold capacity (and its connection to packing).

We remark that one don't necessarily need to sample $\by$ uniformly at random from $\{\pm1\}^P$. It is also reasonable to fix a choice of $\by$ or sample $\by$ from a subset of $\{\pm1\}^P$ (e.g., all the 1-versus-rest dichotomies) as discussed in~\citep{chou2025glue}.

\begin{algorithm}
\caption{Estimate simulated manifold capacity}\label{alg:simulation capacity}
\begin{flushleft}
    \textbf{Input:} $\{\mathcal{M}_i\}$: $P$ point clouds, each containing $M$ points in an $N$-dimensional ambient space.
    \\
    \textbf{Output:} $\alpha_\simcap$: Simulated manifold capacity.
\end{flushleft}
\begin{algorithmic}
    \State $n^*\gets \text{BinarySearch}(\{\cM_i\},1,N)$.
    \State $\alpha_\simcap\gets P/n^*$.
    \State \textbf{return}  $\alpha_\simcap$.
    
    \State
    \State \% Binary search for the smallest $n$ such that $p_n\geq0.5$.
    \Function{BinarySearch}{($\{\cM_i\},n_l,n_r$)}
        \State $n_m\gets\lfloor(n_l+n_r)/2\rfloor$.
        \If{$n_m=n_l$}
            \State \textbf{return} $n_m$.
        \Else
            \State $p_{n_m}\gets \text{EstProb}(\{\cM_i\},n_m)$.
            \If{$p_{n_m}>0.5$}
                \State \textbf{return} $\text{BinarySearch}(\{\cM_i\},n_l,n_m)$.
            \Else
                \State \textbf{return} $\text{BinarySearch}(\{\cM_i\},n_m,n_r)$.
            \EndIf
        \EndIf
    \EndFunction
    
    \State
    \State \% Estimate the probability of linear separability after random projection.
    \Function{EstProb}{($\{\cM_i\},n, m=1000$)}
        \State $\textsf{cnt}\gets0$
        \For{$i$ from $1$ to $m$}
            \State $\Pi\gets$ a random projection from $\Real^N$ to $\Real^n$.
            \State $\by\gets$ a random vector from $\{\pm1\}^P$.
            \State $\{\cM_i'\}\gets \{\Pi\cM_i\}$.
            \If{$\{\cM_i'\}$ is linearly separable}
                \State $\textsf{cnt}\gets\textsf{cnt}+1$.
            \EndIf
        \EndFor
        \State \textbf{return} $\textsf{cnt}/m$.
    \EndFunction
\end{algorithmic}

\end{algorithm}

\subsection{A mean-field definition for manifold capacity}
To overcome the above-mentioned drawbacks of simulated manifold capacity, previous work~\citep{chung2018classification,wakhloo2023linear,chou2025glue} defined a \textit{mean-field models} to enable a mathematical definition of manifold capacity while providing a good approximation to the simulated manifold capacity. In particular, in this paper we use the GLUE (Geometry Linked to Untangling Efficiency) theory developed in~\citep{chou2025glue}.

\paragraph{Mean-field model from GLUE~\citep{chou2025glue}.}
Given a collection of (finite) data manifolds \(\{\cM_i\}_{\mu=1}^{P}\). A mean-field model is to generate infinitely many ($P_{\mf}$) manifolds in an infinite-dimensional ($N_{\mf}$) space and characterizing the largest possible $P_{\mf}/N_{\mf}$ such that these ``mean-field'' manifolds are separable. The key idea is that if this generating process nicely preserve the structure in the data manifolds, then the packing property of these mean-field manifolds will be very similar

\begin{definition}[Mean-field model from~\citep{chou2025glue}]\label{def:mean-field model GCMC}
Let $\{\cM_i\}_{i\in[P]}$ be a collection of data manifolds in $\Real^{N}$ as defined in~\autoref{eq:manifold def}. 
Let $\alpha\in\Real_{\geq0}$ and $P_{\mf},N_{\mf}$ be integers with the following properties: (i) $P_{\mf},N_{\mf}\to\infty$ and (ii) $P_{\mf}/N_{\mf}=\alpha<\infty$, and $P_{\mf}$ be divisible by $P$. We define the mean-field manifolds $\cM_{\mf}(P_{\mf},N_{\mf})=\{\cM^{a,i}_{\mf}\}_{a\in[P_{\mf}/P],i\in[P]}$ as follows.
\begin{itemize}
\item First, find an orthogonal basis $\{\be_k\}_{k=1}^{N}$ in $\Real^{N}$ for the basis vectors of all the data manifolds. Namely, for each $i\in[P]$, there exists a linear transformation $Q^i\in\Real^{(D_{i}+1)\times N}$ such that $\bu^i_j=\sum_{k}Q^{i,j}_k\be_k$ for each $j\in\{0,1,\dots,D_{i}\}$.
\item Next, for each $a\in[P_{\mf}/P]$, generate $\bv^{a}_1,\dots,\bv^{a}_{N}\sim\cN(0,I_{N_{\mf}})$ independently and let $\bV^a$ be the $N_{\mf}\times N$ matrix with $\bv^a_j$ on its columns. 
\item Define $M^{a,i}_{\mf}=\left\{(\bV^aQ^i)_0 + \sum_{j=1}^{D_{i}}s_j(\bV^aQ^i)_j\ :\ \bs=(s_1,\dots,s_{D_{i}})\in\cS_i\right\}$ as the $i$-th manifold in the $a$-th cloud where $(\bV^aQ^i)_i=\sum_k\bv^a_{k}Q^{i,j}_k$ for every $a\in[P_{\mf}/P]$ and $i\in[P]$.
\end{itemize}
\end{definition}

Now, we are ready to formally define the mean-field version of manifold capacity.
\begin{definition}[Mean-field manifold capacity~\cite{chung2018classification,chou2025glue}]\label{def:capacity mean-field}
Let $\{\cM_i\}_{i\in[P]}$ be a collection of data manifolds in $\Real^{N}$ as defined in~\autoref{eq:manifold def}. 
The manifold capacity of $\{\cM_i\}_{i\in[P]}$ is defined as
\[
\alpha_{\mf} := \lim_{N_{\mf}\to\infty}\max_{P_{\mf}}\left\{ \frac{P_{\mf}}{N_{\mf}}\, :\, \Pr_{\by,\cM_{\mf}(P_{\mf},N_{\mf})}\left[\substack{\exists\theta\in\Real^{N_{\mf}},\ \forall a\in[P_{\mf}/P],\, i\in[P],\\ \min_{\bs\in\cM_{\mf}^{a,i}}y_i\langle\theta,\bs\rangle\geq0}\right]\geq1-o_{N_{\mf}}(1) \right\}
\]
where and $o_{N_{\mf}}(1)\to0$ as $N_{\mf}\to\infty$.
\end{definition}

Finally, previous work~\citep{chung2018classification,chou2025glue} derived a formula for mean-field manifold capacity as follows.
\begin{align}
\alpha^{-1}_{\mf} &= \frac{1}{P}\Exp_{\substack{\by\sim\{\pm1\}^P\\T\sim\cN(0,I_{N})}}\left[\max_{\bs_i\in\cM_i}\left\{\|\proj_{\textsf{cone}(\{y_i\bs_i\})}\bt\|_2^2\right\}\right] \label{eq:gcmc}\\
& = \frac{1}{P}\Exp_{\substack{\by\sim\{\pm1\}^P\\T\sim\cN(0,I_{N})}}\left[\max_{\substack{\bs_i\in\cM_i\\\lambda_i\geq0}}\left\{\left(\frac{-T\cdot\sum_{i}\lambda_iy_i\bs_i}{\|\sum_{i}\lambda_iy_i\bs_i\|_2}\right)_+^2\right\}\right] \nonumber
\end{align}
where $\cN(\mu,\Sigma)$ denotes the multivariate Gaussian distribution with mean $\mu$ and covariance $\Sigma$ and $\textsf{cone}(\cdot)$ is the convex cone spanned by the vectors, i.e., $\textsf{cone}(\{y_i\bs_i\})=\{\sum_i\lambda_iy_i\bs_i\, :\, \lambda_i\geq0\}$.

\subsection{Effective geometric measures from capacity formula}\label{app:geometry}
The advantages of mean-field manifold capacity are: (i) $\alpha_{\mf}$ can be estimated via solving a quadratic program (\autoref{alg:capacity}) and (ii)~\autoref{eq:capacity formula main text} connects manifold capacity to the structure of the manifolds $\{\cM_i\}$. Specifically, for each $\by,\bt$, define$\{\bs_i(\by,\bt)\} = y_i\cdot\arg\max_{\{\bs_i\}}\|\proj_{\textsf{cone}(\{y_i\bs_i\})}\bt\|_2^2$ as the \textit{anchor points} with respect to $\by$ and $T$.
Intuitively, these anchor points are the support vectors with respect to some random projection and dichotomy as in~\cref{def:capacity simulated}.
Specifically, these anchor points are analytically linked to manifold capacity via~\autoref{eq:capacity formula main text} and are distributed over the manifolds $\{\cM_i\}$.
This connection inspired the previous work~\citep{chung2018classification,chou2025glue} to define the following effective manifold geometric measures that capture the structure of manifolds while being analytically connected to capacity.

The first key idea of defining effective geometric measure is the segregation of anchor points into their \textit{center part} and their \textit{axis part}. Concretely, for each $i\in[P]$, define $\bs_i^0:=\Exp_{\by,\bt}[\bs_i(\by,\bt)]$ as the center of the $i$-th manifold and define $\bs_i^1(\by,\bt):=\bs_i(\by,\bt)-\bs_i^0$ to be the axis part of $\bs_i(\by,\bt)$ for each pair of $(\by,\bt)$.

Next,~\citep{chung2018classification} used an identity: $a=\frac{b}{1+\frac{b-a}{a}}$, and set $a=\|\proj_{\textsf{cone}(\{\bs_i(\by,\bt)\}_i)}\bt\|_2^2$ and $b=\|\proj_{\textsf{cone}(\{\bs_i^1(\by,\bt)\}_i)}\bt\|_2^2$ to rewrite the capacity formula (\autoref{eq:gcmc}) as follows.

\begin{align}
\alpha^{-1}_{\mf} &= \frac{1}{P}\Exp_{\by,\bt}\left[\|\proj_{\textsf{cone}(\{\bs_i(\by,\bt)\}_i)}\bt\|_2^2\right] \nonumber\\
& = \frac{1}{P}\Exp_{\by,\bt}\left[\frac{\|\proj_{\textsf{cone}(\{\bs_i^1(\by,\bt)\}_i)}\bt\|_2^2}{1+\frac{\|\proj_{\textsf{cone}(\{\bs_i^1(\by,\bt)\}_i)}\bt\|_2^2-\|\proj_{\textsf{cone}(\{\bs_i(\by,\bt)\}_i)}\bt\|_2^2}{\|\proj_{\textsf{cone}(\{\bs_i(\by,\bt)\}_i)}\bt\|_2^2}}\right] \, . \nonumber
\intertext{Then, they proceeded with the following approximation.}
&\approx \frac{\frac{1}{P}\Exp_{\by,\bt}\left[\|\proj_{\textsf{cone}(\{\bs_i^1(\by,\bt)\}_i)}\bt\|_2^2\right]}{\Exp_{\by,\bt}\left[1+\frac{\|\proj_{\textsf{cone}(\{\bs_i^1(\by,\bt)\}_i)}\bt\|_2^2-\|\proj_{\textsf{cone}(\{\bs_i(\by,\bt)\}_i)}\bt\|_2^2}{\|\proj_{\textsf{cone}(\{\bs_i(\by,\bt)\}_i)}\bt\|_2^2}\right]} \, . \label{eq:geometric approx}
\end{align}
\citep{chung2018classification,chou2025glue} found that the above approximation empirically performs well. Furthermore, as the numerator mimics the notion of Gaussian width of a convex body and the denominator behaves like (normalized) radius of a sphere, they defined effective manifold dimension and radius as follows.

\begin{definition}[Effective manifold geometric measures~\citep{chung2018classification,chou2025glue}]\label{def:geometry full}
For each $i\in[P]$, define $\bs_i^0:=\Exp_{\by,\bt}[\bs_i(\by,\bt)]$ as the \textbf{center} of the $i$-th manifold and define $\bs_i^1(\by,\bt):=\bs_i(\by,\bt)-\bs_i^0$ to be the \textbf{axis} part of $\bs_i(\by,\bt)$ for each pair of $(\by,\bt)$.
\begin{itemize}
\item \textbf{Manifold dimension} captures the degree of freedom of the noises/variations within the manifolds. Formally, it is defined as $D_\mf:=\Exp_{\by,\bt}[\|\proj_{\textsf{cone}(\{\bs_i^1(\by,\bt)\}_i)}\bt\|_2^2]$.
\item \textbf{Manifold radius} captures the noise-to-signal ratio of the manifolds. Formally, it is defiend as $R_\mf := \sqrt{\mathbb{E}_{\by,\bt}\left[\frac{\|\textsf{proj}_{\textsf{cone}(\{\mathbf{s}_i(\by,\bt)\}_{i})}\bt\|^2}{\|\textsf{proj}_{\textsf{cone}(\{\mathbf{s}^1_i(\by,\bt)\}_{i})}\bt\|^2-\|\textsf{proj}_{\textsf{cone}(\{\mathbf{s}_i(\by,\bt)\}_{i})}\bt\|^2}\right]}$.
\item \textbf{Center alignment} captures the correlation between the center of different manifolds. Formally, it is defined as $\rho^c_\mf:=\frac{1}{P(P-1)}\sum_{i\neq j}|\langle\bs_i^0,\bs_j^0\rangle|$.
\item \textbf{Axis alignment} captures the correlation between the axis of different manifolds. Formally, it is defined as $\rho^a_\mf:=\frac{1}{P(P-1)}\sum_{i\neq j}\Exp_{\by,\bt}[|\langle\bs_i^1(\by,\bt),\bs_j^1(\by,\bt)\rangle|]$.
\item \textbf{Center-axis alignment} captures the correlation between the center and axis of different manifolds. Formally, it is defined as $\psi_\mf:=\frac{1}{P(P-1)}\sum_{i\neq j}\Exp_{\by,\bt}[|\langle\bs_i^0,\bs_j^1(\by,\bt)\rangle|]$.
\end{itemize}
\end{definition}

\paragraph{A capacity approximation formula by dimension and radius.}
Recall that in~\autoref{eq:geometric approx} previous work~\citep{chung2018classification} used the identity $a=\frac{b}{1+\frac{b-a}{a}}$ to approximate the manifold capacity. After defining manifold dimension and radius, one can then plug them back to~\autoref{eq:geometric approx} and get the following approximation of manifold capacity via effective manifold dimension and radius.
\begin{equation}\label{eq:geometric approx 2}
\alpha_{\mf}\approx\frac{1+R^{-2}_{\mf}}{D_{\mf}} \, .
\end{equation}

\subsection{Connections between manifold capacity and its effective geometric measures}\label{app:connection between capacity and geometry}

Here, we demonstrate the connections between manifold capacity and its effective geometric measures by synthetic manifolds. In particular, we consider isotropic Gaussian clouds parametrized by a set of \textit{ground truth} latent parameters: dimension $D_{\text{ground}}$, radius $R_{\text{ground}}$, center correlations $\rho^c{\text{ground}}$, axis correlations $\rho^a_{\text{ground}}$, and center-axis correlations $\psi_{\text{ground}}$. See~\autoref{app:synthetic generation} for more details on the generative process. In this section, we focus on showing that the effective geometric measures $D_\mf, R_\mf, \rho^c_\mf,\rho^a_\mf,\psi_\mf$ capture the corresponding ground truth parameter.

\begin{figure}[ht]
    \centering
    \includegraphics[width=\linewidth]{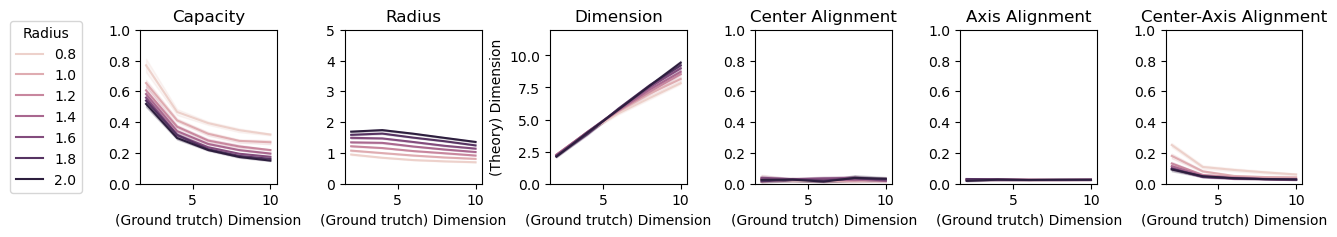}
    \caption{Effective manifold dimension tracks the ground truth dimension of uncorrelated isotropic Gaussian clouds. Note that the higher the dimension, the smaller capacity, as discussed in~\autoref{fig:method}c.}
    \label{fig:justification D}
\end{figure}

\begin{figure}[ht]
    \centering
    \includegraphics[width=\linewidth]{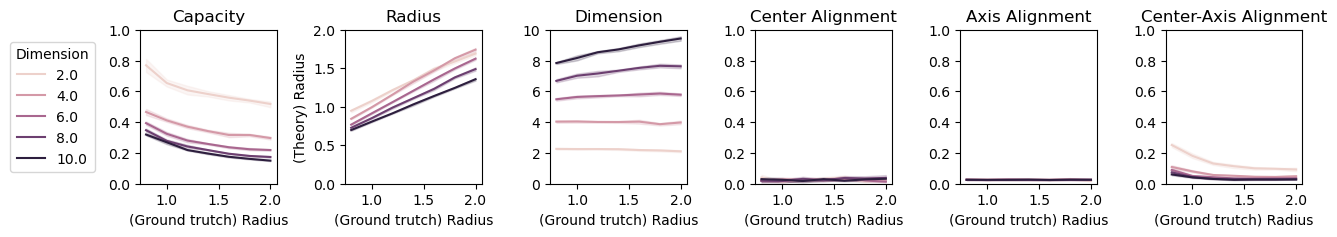}
    \caption{Effective manifold radius tracks the ground truth radius of uncorrelated isotropic Gaussian clouds. Note that the higher the radius, the smaller capacity, as discussed in~\autoref{fig:method}c.}
    \label{fig:justification R}
\end{figure}

\paragraph{Effective manifold dimension and radius.}
We first set all the manifold correlations to be zero and vary the ground truth radius and dimension. Here we pick $N=1000$ neurons, $P=2$ manifold, $M=200$ points per manifold, varying the underlying dimension from $2$ to $10$, and varying the underlying radius from $0.8$ to $2$. In~\autoref{fig:justification D}, we vary the ground truth dimension in the x-axis, and in~\autoref{fig:justification R}, we vary the ground truth radius in the x-axis.

\begin{figure}[ht]
    \centering
    \includegraphics[width=\linewidth]{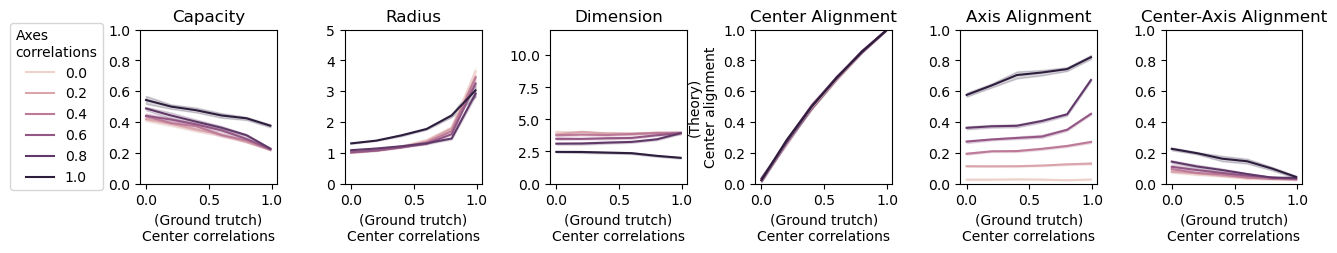}
    \caption{Effective manifold center alignment tracks the ground truth center correlations of isotropic Gaussian clouds. Note that the higher the center alignment, the smaller capacity, as discussed in~\autoref{fig:method}c. Also, in the large center correlations regime, the effective radius increases.}
    \label{fig:justification center}
\end{figure}

\begin{figure}[ht]
    \centering
    \includegraphics[width=\linewidth]{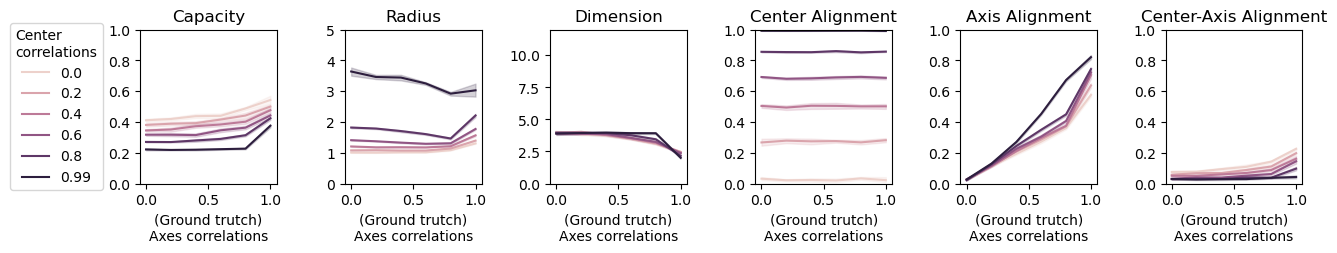}
    \caption{Effective manifold axis alignment tracks the ground truth axis correlations of isotropic Gaussian clouds. Note that the higher the axis alignment, the higher capacity, as discussed in~\autoref{fig:method}c. Also, in the large axis correlations regime, the effective dimension decreases.}
    \label{fig:justification axis}
\end{figure}

\paragraph{Effective alignment measures.}
Next, we fix the ground truth dimension to be $D_{\textsf{ground}}=4$ and radius to be $R_{\textsf{ground}}=1$ and vary $\rho^c_{\text{ground}},\rho^a_{\text{ground}},\psi_{\text{ground}}$ from $0$ to $0.8$. In~\autoref{fig:justification center}, we vary the center correlations, and in~\autoref{fig:justification axis}, we vary the axis correlations.

\subsection{Algorithms for estimating manifold capacity and effective geometric measure}\label{app:algorithms}
We provide pseudocodes for estimating  manifold capacity and effective geometric measure in~\autoref{alg:capacity}.

\begin{algorithm}
\caption{Estimate manifold capacity and effective geometric measures}\label{alg:capacity}
\begin{flushleft}
    \textbf{Input:} $\{\mathcal{M}_i\}$: $P$ point clouds, each containing $M$ points in an $N$-dimensional ambient space; $n_t$: number of samples for estimating the expectation.
    \\
    \textbf{Output:} $\alpha_\mf$: Manifold capacity; $D_\mf$: Effective dimension; $R_\mf$: Effective radius; $\rho^a_\mf$: Effective axis alignment; $\rho^c_\mf$: Effective center alignment; $\psi_\mf$: Effective center-axis alignment.
\end{flushleft}
\begin{algorithmic}
    \State \% Step 1: Sample anchor points.
    \For{$k$ from $1$ to $n_t$}
        \State $\bt_k\gets$ a vector sampled from isotropic $N$-dimensional Gaussian distribution.
        \State $\by\gets$ a random dichotomy vector from $\{\pm1\}^P$.
        \State $\mathbf{A}\gets I_{N}$; $\mathbf{q}\gets-\bt_k$; $\mathbf{h}\gets\mathbf{0}_{N}$. 
        \State $\mathbf{G}\gets (y\odot \{\mathcal{M}_i\}_{i=1}^P)$. \Comment{$\mathbf{G}_{i,j}=y_i\bs$ is a row vector where $\bs$ is the $j$-th point in $\cM_i$.}
        \State $\textsf{output}\gets qp(\mathbf{A},\mathbf{q},\mathbf{G},\mathbf{h})$. \Comment{$\min_{\mathbf{x}} \frac{1}{2}\mathbf{x}^\top \mathbf{A}\mathbf{x}+\mathbf{q}^\top \mathbf{x}$ s.t. $\mathbf{G}\mathbf{x}\leq h$.
        }
        \State $\mathbf{z}_\textsf{dual}\gets\textsf{output}[``\text{dual}'']$ \Comment{The support vectors
        }
        \For{$i$ from $1$ to $P$}
            \State $\mathbf{s}_{i}[k]\gets \sum_j(\mathbf{z}_\text{dual})_{i,j}^\top\mathbf{G} / \sum_j(\mathbf{z}_\text{dual})_{i,j}$
        \EndFor
    \EndFor
    
    \State
    \State \% Step 2: Estimate (anchor) manifold centers.
    \For{$i$ from $1$ to $P$}
        \State $\mathbf{s}_i^0\gets\frac{1}{n_t}\sum_{k=1}^{n_t}\mathbf{s}_i[k])$.
    \EndFor
    \State $\mathbf{G}^0 \gets \sum_i\bs_i^0(\bs_i^0)^\top$. \Comment{Anchor center gram matrix.}

    \State
    \State \% Step 3: Separate the center and axis part of anchor points.
    \For{$k$ from $1$ to $n_t$}
        \For{$i$ from $1$ to $P$}
            \State $\bs_i^1[k]\gets\bs_i[k]-\bs_i^0$. \Comment{The axis part of the anchor poitn in the $i$-th manifold.}
        \EndFor
        \State $\bt^1[k]\gets \sum_i\bs_i^1[k]\bt_k$.
        \State $\mathbf{G}^1[k]\gets \sum_i\bs^1_i[k](\bs^1_i[k])^\top$. \Comment{Anchor axis gram matrix.}
    \EndFor
    
    \State
    \State \% Step 4: Estimate manifold capacity and effective geometric measures.
    \State $\alpha_\mf\gets (\frac{1}{n_tP}\sum_{k=1}^{n_t}(\mathbf{s}_i[k] \bt_k)^\top(\mathbf{s}_i[k](\mathbf{s}_i[k]^\top)^\dagger(\mathbf{s}_i[k] \bt_k))^{-1}$.
    \State $D_\mf\gets\frac{1}{n_tP}\sum_{k=1}^{n_t}\bt^1[k]^\top \mathbf{G}^1[k]^\dagger \bt^1[k]$. 
    \State $R_\mf\gets\sqrt{\frac{1}{n_t}\sum_{k=1}^{n_t}\frac{\bt^1[k]^\top(\mathbf{G}^1[k]+\mathbf{G}^0)^\dagger \bt^1[k]}{\bt^1[k]^\top(\mathbf{G}^1[k]+\mathbf{G}^1[k](\mathbf{G}^0)^\dagger \mathbf{G}^1[k])^\dagger \bt^1[k]}}$. \Comment{Equivalent to the definition of radius after applying the Woodbury formula for numerical stabiltiy.}
    \State $\rho^c_\mf\gets\frac{1}{P(P-1)}\sum_{i=1}^P\sum_{i\neq j}\frac{(\mathbf{s}^0_i)^\top \mathbf{s}^0_j}{\|\mathbf{s}^0_i\|_2\cdot\|\mathbf{s}^0_j\|_2}$.
    \State $\rho^a_\mf\gets \frac{1}{P(P-1)}\sum_{i=1}^{P}\sum_{j\neq i}\frac{1}{n_k}\sum_{k=1}^{n_k} \frac{\bs^1_i[k]^\top\bs^1_j[k]}{\|\bs^1_i[k]\|_2\cdot\|\bs^1_j[k]\|_2}$. 
    \State $\psi_\mf\gets \frac{1}{P(P-1)}\sum_{i=1}^{P}\sum_{j\neq i}\frac{1}{n_k}\sum_{k=1}^{n_k} \frac{(\bs^0_i)^\top\bs^1_j[k]}{\|\bs^0_i\|_2\cdot\|\bs^1_j[k]\|_2}$. 
    \State \textbf{return}  $\alpha_\mf, D_\mf, R_\mf, \rho^a_\mf, \rho^c_\mf$, $\psi_\mf$.
\end{algorithmic}

\end{algorithm}


\section{Theoretical Results}\label{app:2-layer theory}

We consider the training of a fully-connected 2-layer network of the form $f(\bx)=\frac{1}{\sqrt{N}}\ba^\top\sigma(W^\top\bx)$, where $\bx\in\Real^d$ is an input, $W\in\Real^{N\times d}$ is the hidden layer matrix, $\ba\in\Real^N$ is the readout weight, and $\sigma:\Real\to\Real$ is the (non-linear) activation function. 
To study feature learning in this setting, it is common to consider $W$ to be randomly initialized (i.e., random feature model~\citep{rahimi2007random}) and update via gradient descent with squared loss.
Meanwhile, the readout weight $\ba$ is randomly initialized and fixed to avoid lazy learning (where the network minimally adjusts the hidden layer and focuses on learning a good readout weight) as well as enable mathematical analysis~\citep{ba2022high}.
Input data and label $(\bx_1,y_1),\dots,(\bx_{P_{\text{train}}},y_{P_{\text{train}}})$ were randomly generated by a teacher-student setting, where there is a hidden signal direction $\beta^*$ that correlates with the label (see~\cref{setting:2-layer} for the full setting). 
As previously proved in ~\cite{ba2022high} (see~\cref{prop:ba}), in the proportional asymptotic limit (i.e., $P_{\text{train}},d,N\to\infty$ at the same rate), the first-step gradient update can be approximated by a rank-1 matrix that contains label information, resulting in the updated weight to be more aligned with the hidden signal $\beta^*$. Hence, in this setting, the learning rate $\eta$ can be used as the ground-truth to measure the amount of task-relevant information (i.e., richness in learning) in the model representation after gradient updates.

We extend the previous results in~\citep{ba2022high} from a regression setting to a classification setting. Specifically, We prove that capacity correctly tracks the effective degree of richness after one gradient step\footnote{Here we follow the convention in~\citep{ba2022high} and study only the first gradient step as the key Gaussian equivalence step might not hold for more steps as remarked in footnote 2 of~\citep{ba2022high}.}. Moreover, we derive a monotone connection between capacity and prediction accuracy, thereby justifying the use of capacity as a measure of richness in feature learning within a well-studied theoretical setting.
Here, we provide an informal statement of our results and leave the formal version and proof in~\autoref{app:2-layer theory}.

\subsection{Formal theorem statement}
Let $d\in\N$ be the input dimension and $N\in\N$ be the number of hidden units.
Let $W_0\in\Real^{N\times d}$ be the weight matrix of a fully connected 2-layer neural network. The feature of an input vector is defined as $\Phi_0(\bx)=\sigma(W_0\bx)$ where $\sigma(\cdot):\Real\to\Real$ is a non-linear activation function, e.g., ReLU or tanh. The readout weight is denoted as $\ba\in\Real^N$. Finally, the output of the 2-layer NN is the sign of the readout, i.e., $f(\bx)=\sgn(\ba^\top\Phi(\bx))$.

Let $\{(\bx_i,y_i)\}_{i\in[\Ptrain]}$ be the collection of training data. We consider gradient descent over the mean square error (MSE) of the 2-layer NN, i.e., $\cL(f)=\frac{1}{\Ptrain}\sum_{i\in[\Ptrain]}\ell(f(\bx_i),y_i)$ where $\ell(z_i,y_i)=\frac{1}{2}(z-y)^2$. The gradient update with learning rate $\eta>0$ is $W_{t+1}=W_t+\eta G_t$ where
\[
G_t = \frac{1}{\Ptrain}\sum_{i\in[\Ptrain]}\left[(y_i-\ba^\top\sigma(W_t\bx_i))\ba\odot\sigma'(W_t\bx_i)\right]\bx_i^\top
\]
and $\sigma'(\cdot)$ denotes the first order derivative of $\sigma(\cdot)$. 

\begin{assumption}\label{assumption:2-layer}
We adopt the following assumptions used in~\citep{montanari2019generalization,ba2022high}.
\begin{enumerate}
\item (Proportional limit) $P_{\text{train}},d,N\to\infty$ with $\psi_1=N/d$, $\psi_2=P_{\text{train}}/d$, and $0<\psi_1,\psi_2<\infty$.
\item (Gaussian initialization) $[W_0]_{kj}\sim\cN(0,1/N)$ for each $k\in[N]$ and $j\in[d]$.
\item (Gaussian readout) $a_k\sim\cN(0,1/N)$ for each $k\in[N]$.
\item (Normalized activation) The non-linear activation function $\sigma(\cdot)$ has $O(1)$-bounded first three derivatives almost surely. In addition, $\Exp[\sigma(G)]=0$ and $\Exp[G\sigma(G)]\neq0$ for $G\sim\cN(0,1)$.
\item (Non-degenerate label function) Let $F:\Real\to[0,1]$ be a continuous function satisfying
\[
\text{inf}\left\{x:\Pr[T<x]>0\right\} = -\infty\ \ \text{and}\ \ \text{sup}\left\{x:\Pr[T>x]>0\right\} = \infty
\]
where $T=YG$, $G\sim\cN(0,1)$, and $\Pr[Y=1\, |\, G]=1-\Pr[Y=-1\, |\, G]=F(G)$.
\end{enumerate}
\end{assumption}

\begin{setting}\label{setting:2-layer}
We consider the following data generation process. Let $F:\Real\to[0,1]$ be a function satisfying~\cref{assumption:2-layer}. Let $\beta_*\in\Real^d$ be a hidden vector with $\|\beta_*\|_2=1$. The data distribution $\cD_F(\beta_*)$ is defined by the following two steps: (i) sample $\bx\sim\cN(0,I_d)$, and (ii) sample $y$ with $\Pr[y=1]=1-\Pr[y=-1]=F(\langle\beta_*,\bx\rangle)$.
Finally, the prediction accuracy of a network is defined as the expected accuracy of a fresh sample, i.e., $\Pr_{(\bx,y)\sim\cD_F(\beta_*)}[yf(\bx)\geq0]$.
\end{setting}

\begin{parameter}\label{param:2-layer}
Given $\psi_1,\psi_2,F,\beta_*$ from~\cref{assumption:2-layer} and~\cref{setting:2-layer}. We define the following parameters.
\begin{align*}
\gamma_1 &= \Exp_{G\sim\cN(0,1)}[G\sigma(G)]\\
\gamma_2^2 &= \Exp_{G\sim\cN(0,1)}[\sigma(G)^2]-\Exp_{G\sim\cN(0,1)}[G\sigma(G)]^2\\
\theta_1 &= \Exp_{X\sim\mu_{\psi_1}}\left[\frac{\gamma_1^2}{\gamma_1^2X+\gamma_2^2}\right]\\
\theta_2 &= \psi_1\Exp_{X\sim\mu_{\psi_1}}\left[\frac{\gamma_1^2X}{\gamma_1^2X+\gamma_2^2}\right]\\
\theta_3 &= \Exp_{(G,Y)\sim\cD_F}[YG]\\
\theta_4 &= \left(\frac{1}{\psi_2}+\Exp_{(G,Y),(G',Y')\overset{\text{i.i.d.}}{\sim}\cD_F}[YY'GG']\right)
\end{align*}
where $\mu_{\psi_1}$ is the Marchenko-Pastur distribution with the ratio parameter being $\psi_1$ and $(G,Y)\sim\cD_F$ is defined as the sampling process: $G\sim\cN(0,1)$ and $\Pr[Y=1]=1-\Pr[Y=-1]=F(G)$.
\end{parameter}

\begin{theorem}\label{thm:storage capacity formal}
Given~\cref{assumption:2-layer} and consider $0<\psi_1,\psi_2,\eta<\infty$. 
\begin{enumerate}
\item (Capacity tracks the degree of feature learning) The storage capacity of 2-layer network trained with synthetic data defined in~\cref{setting:2-layer} after one gradient step is $\alpha_{P_{\text{train}},d,N}(\psi_1,\psi_2,\eta)$ and
\[
\alpha_{P_{\text{train}},d,N}(\psi_1,\psi_2,\eta) \xrightarrow{P_{\text{train}},d,N\to\infty} \alpha(\psi_1,\psi_2,\eta)
\]
Here the function $\alpha(\cdot)$ is defined as
\[
\alpha(\psi_1,\psi_2,\eta) = \left(\min_{c\in\Real}\Exp_{(Z,G,Y)\sim\cD_{\psi_1,\psi_2,\eta}}\left[(-cYG-Z)_+^2\right]\right)^{-1}
\]
where $(Z,G,Y)\sim\cD_{\psi_1,\psi_2,\eta}$ is defined as the following sampling process
\[
Z\sim\cN(0,1),\ G\sim\cN(0,1),\ \Pr[Y=1]=1-\Pr[Y=-1]=f_{\tau(\psi_1,\psi_2,\eta)}(G)
\]
and the scalar function $f_{\tau}(\cdot)$ and $\tau(\psi_1,\psi_2,\eta)$ are defined as 
\[
f_\tau(G) = \Exp_{G'\sim\cN(0,1)}\left[F(\sqrt{1-\tau^2} G+\tau G')\right]
\]
and
\[
\tau=\tau(\psi_1,\psi_2,\eta) = \sqrt{\tau_0(\psi_1,\psi_2)^2 - \tau_\Delta(\psi_1,\psi_2,\eta)^2}
\]
where $\tau_0(\cdot)$ and $\tau_\Delta(\cdot)$ are scalar functions defined as
\[
\tau_0(\psi_1,\psi_2)^2 = 1-\theta_2
\]
and
\[
\tau_\Delta(\psi_1,\psi_2,\eta)^2 = \frac{\eta^2\theta_1(1-\theta_2)^2\theta_3^2}{1+\eta^2\theta_1(1-\theta_2)\theta_4}
\]
where the parameters $\theta_i$'s are defined in~\cref{param:2-layer}.
In particular, $0<\alpha(\psi_1,\psi_2,\eta)<\alpha(\psi_1,\psi_2,\eta')$ for all $0<\eta<\eta'$.

\item (Capacity analytically links to prediction accuracy) The prediction accuracy of 2-layer network trained with synthetic data defined in~\cref{setting:2-layer} after one gradient step is $\Acc_{P_{\text{train}},d,N}(\psi_1,\psi_2,\eta)$ and
\[
\Acc_{P_{\text{train}},d,N}(\psi_1,\psi_2,\eta) \xrightarrow{P_{\text{train}},d,N\to\infty} \Acc_(\psi_1,\psi_2,\eta)
\]
Here the function $\Acc(\cdot)$ is defined as
\[
\Acc(\psi_1,\psi_2,\eta) = \Exp_{\substack{(G,Y)\sim\cD_F}}\left[\Phi\left(\frac{\eta\gamma_1^2\theta_3}{\sqrt{\frac{\eta^2\gamma_1^4}{\psi_2}+\gamma_1^2+\gamma_*^2}}YG\right)\right]
\]
In particular, there exists an increasing and invertible function $g_{\psi_1,\psi_2}:[0,1]\to\Real_+$ such that
\[
\Acc(\psi_1,\psi_2,\eta) = g_{\psi_1,\psi_2}(\alpha(\psi_1,\psi_2,\eta)) \, .
\]
\end{enumerate}
\end{theorem}

\subsection{Proof for~\autoref{thm:storage capacity formal}}

\paragraph{Step 1: Rank-1 approximation of gradient descent in 2-layer networks by ref.~\citep{ba2022high}.}
When the learning rate is constant, i.e., $\eta=O(1)$, ref.~\citep{ba2022high} shows that the gradient update matrix can be approximated by a rank-1 matrix. In particular, the following is a restatement of Proposition 2 in~\citep{ba2022high}.
\begin{proposition}[Proposition 2 in~\citep{ba2022high}]\label{prop:ba}
Given~\cref{assumption:2-layer} and~\cref{setting:2-layer}, there exist some constants $c,C>0$ such that for all large $P_{\text{train}},N,d$, the following holds
\[
\left\|G_0-\gamma_1\ba\left(\frac{\sum_iy_i\bx_i^\top}{P_{\text{train}}}\right)\right\|\leq\frac{C\log^2P_{\text{train}}}{\sqrt{P_{\text{train}}}}\cdot\|G_0\|
\]
with probability at least $1-P_{\text{train}}e^{-c\log^2 P_{\text{train}}}$ and $\|\cdot\|$ denotes the operator norm.
\end{proposition}

\paragraph{Step 2: A formula for the storage capacity of a Gaussian model by ref.~\citep{montanari2019generalization}.}
The storage capacity of a Gaussian model is proven in~\citep{montanari2019generalization}. In particular, the following is a restatement of the Proposition 5.1 in~\citep{montanari2019generalization}.

\begin{definition}[Gaussian model]\label{def:gaussian model}
Let $\theta_*\in\Real^N$ be some latent vector. A sample $(\bx_i,y_i)\in\Real^N\times\{\pm1\}$ is i.i.d. sampled as follows. First, sample $\bx_i$ from $\cN(0,\Sigma)$ where $\Sigma$ is a covariance matrix satisfying certain technical condition as defined in Assumption 1-2 in~\citep{montanari2019generalization}. Next, let $y_i=+1$ with probability $f(\langle\theta_*,\bx_i\rangle)$ for some function $f$ satisfying Assumption 3 in~\citep{montanari2019generalization}.
\end{definition}

\begin{proposition}[Theorem 3 in~\citep{montanari2019generalization}]\label{prop:MRSY}
Consider a Gaussian model satisfying~\cref{def:gaussian model}. As $P_{\text{train}},N,d\to\infty$, the storage capacity converges to
\[
\alpha^* = \left(\min_{c\in\Real}\Exp_{(Z,G,Y)\sim\cD_{f}}\left[(-cYG-Z)_+^2\right]\right)^{-1}
\]
where $(Z,G,Y)\sim\cD_{f}$ is defined as the following sampling process
\[
Z\sim\cN(0,1),\ G\sim\cN(0,1),\ \Pr[Y=1]=1-\Pr[Y=-1]=f(\rho\cdot G) \, .
\]
where $\rho$ is some scalar related to the Gaussian model as defined in Assumption 2 of~\citep{montanari2019generalization}.
\end{proposition}

Note that the capacity only depends on the alignment between data and task (as encoded in $f$) and does not depend on the covariance structure. The dependence on the covariance structure will appear when one considers the non-zero margin version of capacity.

\paragraph{Step 3: A Gaussian equivalent model for 2-layer NNs after one gradient step.}
Next, we combine a Gaussian equivalent model for random feature 2-layer NNs in~\citep{montanari2019generalization} (Theorem 3) and the rank-1 approximation of gradient step in~\cref{prop:ba} to get a Gaussian equivalent model for 2-layer NNs after one gradient step.

\begin{proposition}\label{prop:GET}
Given~\cref{assumption:2-layer} and $0<\psi_1,\psi_2,\eta<\infty$. Let $d\in\N$ and $(W_1,\beta_*,F)$ be the weight matrix, hidden vector, and label function from~\cref{setting:2-layer}. Let $\alpha^{\textsf{GM}}_{P_{\text{train}},d,N}(\psi_1,\psi_2,\eta)$ be the capacity of the following Gaussian model:
\begin{align}
\Sigma_{d,\eta} &= \gamma_1^2W_1W_1^\top + \gamma_*^2I \nonumber\\
\theta_{*,d,\eta} &= \alpha_{d,\eta}^{-1}\gamma_1(\gamma_1^2W_1W_1^\top+\gamma_*^2I)^{-1}W_1\beta_* \nonumber\\
\alpha_{d,\eta}^2 &= \gamma_1^2\beta_*^\top W_1^\top(\gamma_1^2W_1W_1^\top+\gamma_*^2I)^{-1}W_1\beta_* \nonumber\\
\tau_{d,\eta}^2 &= 1-\alpha_{d,\eta}^2 \label{eq:tau}\\
f_{d,\eta}(x) &= \Exp_{G\sim\cN(0,1)}[F(\alpha_{d,\eta}x+\tau_{d,\eta}G)]\, . \nonumber
\end{align}

We have that
\[
\lim_{P_{\text{train}},d,N\to\infty}|\alpha_{P_{\text{train}},d,N}(\psi_1,\psi_2,\eta)-\alpha^{\textsf{GM}}_{P_{\text{train}},d,N}(\psi_1,\psi_2,\eta)|=0 \label{eq:GET}
\]
and
\[
\alpha^{\textsf{GM}}_{P_{\text{train}},d,N}(\psi_1,\psi_2,\eta)\xrightarrow{P_{\text{train}},d,N\to\infty} \alpha(\psi_1,\psi_2,\eta) . \label{eq:capacity_limit}
\]
Here the function $\alpha(\cdot)$ is defined as
\[
\alpha(\psi_1,\psi_2,\eta) = \left(\min_{c\in\Real}\Exp_{(Z,G,Y)\sim\cD_{f_\tau(\psi_1,\psi_2,\eta)}}\left[(-cYG-Z)_+^2\right]\right)^{-1}
\]
where the scalar function $f_{\tau}(\cdot)$ and $\tau(\psi_1,\psi_2,\eta)$ are defined as 
\[
f_\tau(G) = \Exp_{G'\sim\cN(0,1)}\left[F(\sqrt{1-\tau^2} G+\tau G')\right]
\]
and
\[
\tau=\tau(\psi_1,\psi_2,\eta) = \lim_{d\to\infty}\tau_{d,\eta} = \sqrt{\tau_0(\psi_1,\psi_2)^2 - \tau_\Delta(\psi_1,\psi_2,\eta)^2} \, .
\]
where $\tau_0(\psi_1,\psi_2)=\lim_{d\to\infty}\tau_{d,0}$.
\end{proposition}


To derive the Gaussian equivalent model in~\cref{prop:GET} of the random features model after one gradient step defined in~\cref{setting:2-layer}, we analyze the following random features and their associated labels:
\[
\Phi_0(\bx_i)=\sigma(W_1\bx_i), \quad \Pr[y_i=1|\bx_i]=1-\Pr[y_i=-1|\bx_i]=F(\langle\beta_*,\bx_i\rangle), \quad \|\beta_*\|_2=1 \label{eq:rf_1step}
\]
where $\bx_i\sim\cN(0,I_d)$ and $W_1 = W_0 + \eta G_0$ while $G_0$ satisfies the bound given in~\cref{prop:ba}. 
Given the assumptions in~\cref{assumption:2-layer}, we can decompose the nonlinear activation function $\sigma$ into Hermite polynomials. Following our parameters in~\cref{param:2-layer}, we define the Gaussian equivalent features of our model as the linearization of~\autoref{eq:rf_1step}:
\[
\bg_i = \gamma_1 W_1 \bx_i + \gamma_2 \bh_i \label{eq:gaussian_linearization}
\]
where $\bh_i\sim \cN(0,I_N)$ are independent from everything else. 
Now, we wish to find a similar linearized Gaussian model for the labels $y_i$ given the Gaussian equivalent features $\bg_i$. It is easy to check that the Gaussian features has the following covariance:
\[
\bg_i \sim \cN(0, \Sigma_{d,\eta}),\quad \Sigma_{d,\eta} = \gamma_1^2W_1W_1^\top + \gamma_*^2I
\]
By matching covariance through~\autoref{eq:gaussian_linearization}, we obtain
\[
\bx_i = \gamma_1 W_1^\top \Sigma_{d,\eta}^{-1}\bg_i + Q^{1/2}\tilde{\bh_i}
\]
where $Q = \gamma_2^2(\gamma_2^2 I_N + \gamma_1^2 W_1^\top W_1)^{-1}$ and $\tilde{\bh_i} \sim \cN(0, I_N)$ are independent of $\bx_i$. Therefore, we can rewrite the label function parameter as
\[
\langle \beta_*, \bx_i \rangle = \alpha_{d,\eta} \langle \theta_{*,d,\eta}, \bg_i \rangle + \varepsilon_i
\]
where $\varepsilon_i \sim \cN(0, \tau_{d,\eta}^2)$ are independent of $\bg_i$. Effectively, we obtain an equivalent label function 
\[
f_{d,\eta}(x) = \Exp_{G\sim\cN(0,1)}[F(\alpha_{d,\eta}x+\tau_{d,\eta}G)]
\]
such that $\Pr[y_i=1|\bx_i]=1-\Pr[y_i=-1|\bx_i]=f_{d,\eta}(\langle \theta_{*,d,\eta}, \bg_i \rangle)$. It is easy to verify that this Gaussian model satisfies the assumptions in~\cref{def:gaussian model}. 


\paragraph{Step 4: Analysis of $\tau$.}
Finally, we combine~\cref{prop:ba} and~\cref{prop:GET} to get the formula for the right hand side of~\autoref{eq:tau}. From~\cref{prop:ba}, we approximate $W_1$ as $W_1 = W_0 + \ba\bu^\top$
where $\bu = \eta\sum_iy_i\bx_i^\top/P_{\text{train}}$.
To rewrite the right hand side of~\autoref{eq:tau}, we first deal with the matrix inverse term using the same trick as in ref.~\citep{ba2022high}. Let $\Sigma_t=\gamma_1^2W_tW_t^\top+\gamma_*^2I$. Observe that
\[
\Sigma_1 = \Sigma_0 + \gamma_1^2\begin{bmatrix}
\ba&\bc
\end{bmatrix}
\begin{bmatrix}
L_1&1\\1&0
\end{bmatrix}
\begin{bmatrix}
\ba^\top\\\bc^\top
\end{bmatrix}
\]
where $\bc=W_0\bu$. By Sherman-Morrison-Woodbury formula, we have
\begin{align*}
\Sigma_1^{-1} &= \Sigma_0^{-1} - \gamma_1^2\Sigma_0^{-1}\begin{bmatrix}
\ba&\bc
\end{bmatrix}\left(
\begin{bmatrix}
L_1&1\\1&0
\end{bmatrix}^{-1}
+\gamma_1^2\begin{bmatrix}
\ba^\top\\\bc^\top
\end{bmatrix}\Sigma_0^{-1}\begin{bmatrix}
\ba&\bc
\end{bmatrix}
\right)^{-1}\begin{bmatrix}
\ba^\top\\\bc^\top
\end{bmatrix}\Sigma_0^{-1}\\
&= \Sigma_0^{-1} - \Delta_{aa} - \Delta_{cc} + \Delta_{ac} + \Delta_{ca}
\end{align*}
where
\begin{align*}
\Delta_{aa} &= \gamma_1^2\frac{L_4-L_1}{D}\Sigma_0^{-1}\ba\ba^\top\Sigma_0^{-1}\\
\Delta_{cc} &= \gamma_1^2\frac{L_3}{D}\Sigma_0^{-1}\bc\bc^\top\Sigma_0^{-1}\\
\Delta_{ac} &= \gamma_1^2\frac{1+L_6}{D}\Sigma_0^{-1}\ba\bc^\top\Sigma_0^{-1}\\
\Delta_{ca} &= \gamma_1^2\frac{1+L_6}{D}\Sigma_0^{-1}\bc\ba^\top\Sigma_0^{-1}
\end{align*}
and
\begin{align*}
L_0 &= \gamma_1^2\beta_*^\top W_0^\top \Sigma_0^{-1} W_0\beta_* \\
L_1 &= \bu^\top\bu\\
L_2 &= \bu^\top\beta_* \\
L_3 &= \gamma_1^2\ba^\top\Sigma_0^{-1}\ba \\
L_4 &= \gamma_1^2\bc^\top\Sigma_0^{-1}\bc \\
L_5 &= \gamma_1^2\bc^\top\Sigma_0^{-1}W_0\beta_* \\
L_6 &= \gamma_1^2\ba^\top\Sigma_0^{-1}\bc \\
L_7 &= \ba^\top\bc\\
L_8 &= \gamma_1^2\ba^\top\Sigma_0^{-1}W_0\beta_* \\
D &= L_3(L_4-L_1)-(1+L_6)^2
\end{align*}

Thus, we can rewrite the right hand side of~\autoref{eq:tau} as follows.
\begin{align*}
\tau_{d,\eta} =&\ 1 - \gamma_1^2\beta_*^\top  (W_0+\ba\bu^\top)^\top \Sigma_0^{-1} (W_0+\ba\bu^\top)\beta_* \\
&+\gamma_1^2\beta_*^\top (W_0+\ba\bu^\top)^\top\Delta_{aa}(W_0+\ba\bu^\top)\beta_* \\
&+\gamma_1^2\beta_*^\top (W_0+\ba\bu^\top)^\top\Delta_{cc}(W_0+\ba\bu^\top)\beta_* \\
&-\gamma_1^2\beta_*^\top (W_0+\ba\bu^\top)^\top\Delta_{ac}(W_0+\ba\bu^\top)\beta_* \\
&-\gamma_1^2\beta_*^\top (W_0+\ba\bu^\top)^\top\Delta_{ca}(W_0+\ba\bu^\top)\beta_* \\
=&\ 1 - L_0 - L_2^2L_3 - 2L_2L_8\\
&+ \frac{L_4-L_1}{D}(L_2L_3+L_8)^2 \\
&+ \frac{L_3}{D}(L_5+L_2L_6)^2 \\
&-2\frac{1+L_6}{D}(L_2L_3+L_8)(L_5+L_2L_6) \, .
\end{align*}
Similar to Proposition 29 in~\citep{ba2022high}, by Hanson-Wright inequality, we have that $L_6,L_8,L_7\to0$.
\begin{align*}
L_0 &\to \theta_2 \\
L_1 &\to \eta^2\theta_4\\
L_2 &= \eta\theta_3 \\
L_3 &\to \gamma_1^2\Exp_{X\sim\mu_{\psi_1}}\left[\frac{1}{\gamma_1^2X+\gamma_2^2}\right] = \theta_1\\
L_4 &\to \gamma_1^2\eta^2\theta_4\cdot\psi_1\Exp_{X\sim\mu_{\psi_1}}\left[\frac{X}{\gamma_1^2X+\gamma_2^2}\right] = \eta^2\theta_2\theta_4\\
L_5 &\to \gamma_1^2\eta\theta_3\cdot\psi_1\Exp_{X\sim\mu_{\psi_1}}\left[\frac{X}{\gamma_1^2X+\gamma_2^2}\right] = \eta\theta_2\theta_3\\
L_6, L_7, L_8 &\to 0 \\
D &\to L_3(L_4-L_1)-1 \to \eta^2\theta_1(\theta_2-1)\theta_4-1
\end{align*}
To sum up, we have
\begin{align*}
\lim_{d\to\infty}\tau_{d,\eta} &=\ 1 -\theta_2 - \frac{\eta^2\theta_1\theta_3^2(\eta^2\theta_1(\theta_2-1)\theta_4-1)}{\eta^2\theta_1(\theta_2-1)\theta_4-1} \\
&+ \frac{\eta^4\theta_1^2(\theta_2-1)\theta_3^2\theta_4}{{\eta^2\theta_1(\theta_2-1)\theta_4-1}} \\
&+ \frac{\theta_1\theta_2^2\theta_3^2}{\eta^2\theta_1(\theta_2-1)\theta_4-1} \\
&- 2\frac{\eta^2\theta_1\theta_2\theta_3^2}{\eta^2\theta_1(\theta_2-1)\theta_4-1}\\
&= 1-\theta_2 - \frac{\eta^2\theta_1(1-\theta_2)^2\theta_3^2}{1+\eta^2\theta_1(1-\theta_2)\theta_4} \, .
\end{align*}
This completes the proof for the first part of~\cref{thm:storage capacity formal}.

\begin{figure}[ht]
    \centering
    \includegraphics[width=12cm]{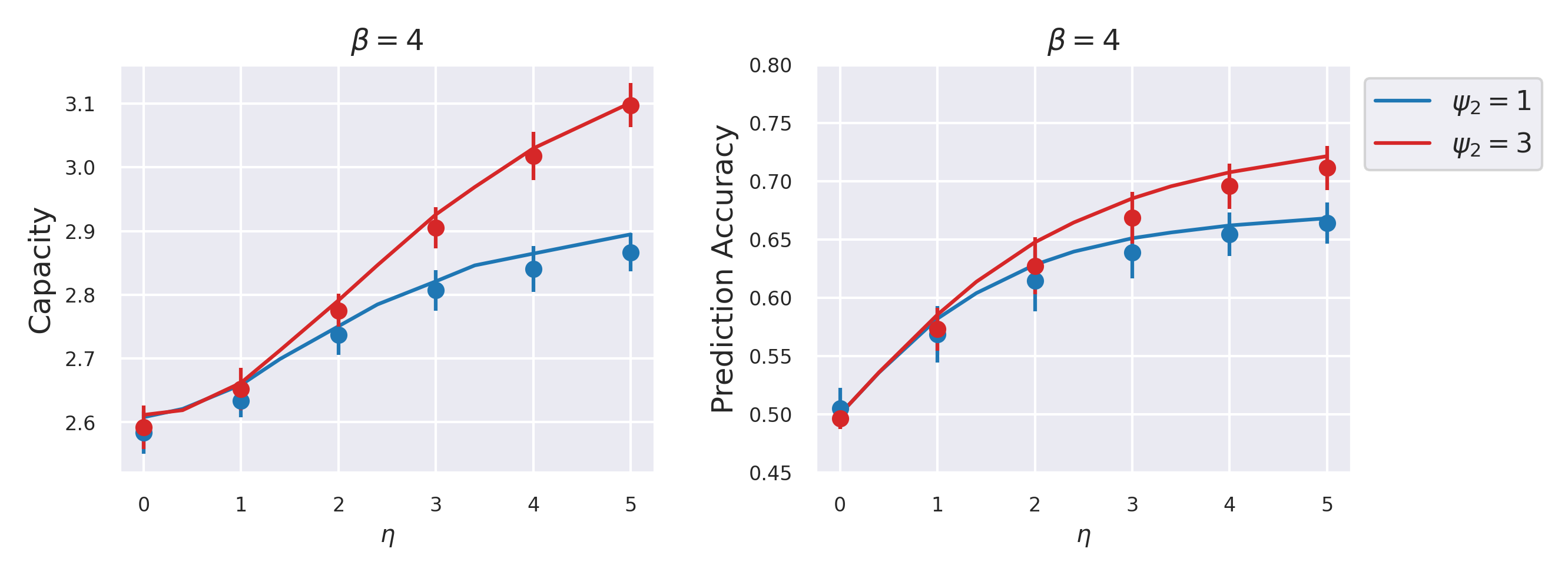}
    \caption{Numerical checks for the formulas in~\cref{thm:storage capacity formal}. We run the simulation with $d=2000$, $\psi_1=1$, ReLU activation, and label function $f(x)=\frac{1}{1+e^{-4x}}$ for 50 repetitions. Left: numerical checks for the capacity formula. Right: numerical checks for the prediction accuracy formula.}
    \label{fig:numerical check}
\end{figure}

\paragraph{Step 5: Analysis for prediction accuracy.}
Recall from~\cref{setting:2-layer} the definition of prediction accuracy of the network after a gradient step is $\Pr_{(\bx,y)\sim\cD_F(\beta_*)}[y\ba^\top\sigma(W_1\bx)\geq0]$. By Gaussian equivalence and~\cref{prop:ba}, we have that the following.
\begin{align*}
&\Acc_{P_{\text{train}},d,N}(\psi_1,\psi_2,\eta) \\ &=\Pr_{\substack{(\bx,y)\sim\cD_F(\beta_*)\\\ba,W_1}}[y\ba^\top\sigma(W_1\bx)\geq0] \, . \\
\intertext{By~\cref{prop:ba}, we can further approximate the equation as follows.}
&= \Pr_{\substack{(\bx,y)\sim\cD_F(\beta_*)\\\ba,W_0,\bu}}[y\ba^\top\sigma((W_0+\ba\bu^\top)\bx)\geq0] + o(1) \, . \\
\intertext{By Gaussian equivalence, we can further approximate the equation as follows.}
&= \Pr_{\substack{(\bx,y)\sim\cD_F(\beta_*)\\\ba,W_0,W_*,\bu}}[y\ba^\top(\gamma_1(W_0+\ba\bu^\top)+\gamma_*W_*)\bx)\geq0]  + o(1) \\
\intertext{where $W_*\in\Real^{N\times d}$ and $([W_*]_{kj}\sim\cN(0,1/N))$ for each $k\in[N], j\in[d]$. Note that as $\ba,W_0,W_*$ are independent, we can further simplify the equation as follows.}
&= \Pr_{\substack{(\bx,y)\sim\cD_F(\beta_*)\\\ba,W_*',\bu}}[y\gamma_1\bu^\top\bx + \sqrt{\gamma_1^2+\gamma_*^2}\cdot y\ba^\top W_*'\bx + o(1)\geq0]  + o(1) \\
\intertext{where $W_*'\in\Real^{N\times d}$ and $([W_*']_{kj}\sim\cN(0,1/N))$ for each $k\in[N], j\in[d]$. Note that as $\ba,W_*'$ are independent, we can further simplify the equation as follows.}
&= \Pr_{\substack{(\bx,y)\sim\cD_F(\beta_*)\\Z\sim\cN(0,1)}}\left[\eta\gamma_1^2\Exp_{(\bx',y')\sim\cD_F(\beta_*)}[yy'\bx'^\top\bx] +\sqrt{\gamma_1^2+\gamma_*^2}\cdot Z + o(1)\geq0\right] + o(1) \, . \\
\intertext{Note that by decomposing $\bx$ and $\bx'$ to direction that's parallel to $\beta_*$ and orthogonal to $\beta_*$, we can further simplify the equation as follows.}
&= \Pr_{\substack{(G,Y)\sim\cD_F\\Z,Z'\sim\cN(0,1)}}\left[\eta\gamma_1^2\left(\Exp_{(G',Y')\sim\cD_F}[YY'GG'] + \sqrt{1/\psi_2}Z'\right) +\sqrt{\gamma_1^2+\gamma_*^2}\cdot Z + o(1)\geq0\right] + o(1) \\
&= \Pr_{\substack{(G,Y)\sim\cD_F\\Z\sim\cN(0,1)}}\left[\eta\gamma_1^2\theta_3YG +\sqrt{\frac{\eta^2\gamma_1^4}{\psi_2}+\gamma_1^2+\gamma_*^2}\cdot Z + o(1)\geq0\right] + o(1) \\
&= \Exp_{\substack{(G,Y)\sim\cD_F}}\left[\Phi\left(\frac{\eta\gamma_1^2\theta_3}{\sqrt{\frac{\eta^2\gamma_1^4}{\psi_2}+\gamma_1^2+\gamma_*^2}}YG\right)\right] +o(1) \, .
\end{align*}
Note that when fixing $\psi_1,\psi_2$ and non-trivial $F$, both capacity formula and prediction accuracy formula are increasing and invertible with respect to $\eta$. As a consequence, the two quantities are also analytically connected by an increasing and invertible function.
This completes the proof for the second part of~\cref{thm:storage capacity formal}. We also provide numeric checks for the formulas in~\autoref{fig:numerical check}.

\section{2-Layer Non-linear Neural Networks}\label{app:2-layer}
In this paper, we use 2-layer non-linear neural networks and Gaussian mixture models (for input data generation) as a convenient experimental setup to systematically explore different regimes in feature learning. Moreover, given its medium level of complexity, it might be possible to have an analytical characterization of our numerical findings, and we leave it as an interesting future direction.

\subsection{Experimental setup}\label{app:2l_experiment_setup}

\subsubsection{Synthetic data generation}\label{app:synthetic generation}
We focus on point manifold, which consists of data points associated with the same label. As discussed in the previous section, we are particularly interested in the effective radius, dimension, center alignment, axes alignment, and center-axes alignment of the representation manifolds. Therefore, we consider a synthetic model to generate training and test data with relevant geometric interpretations. Namely, construct $P\in\N$ synthetic data manifolds with radius $R\in\Real_+$, intrinsic dimension $D\in\N$, size $M\in\N$. The manifold layouts are further determined by center correlation strength $\rho_C\in[0,1)$, axes correlation strength $\rho_A\in[0,1)$, and center-axes correlation strength $\psi\in[0,1)$, all of which we would detail in the following subsections. 

\paragraph{Isotropic spherical manifolds.}\label{para:iso_sph_manifolds}
First, we consider the simplest case: manifolds with isotropic Gaussian center distribution and axes distribution with no correlations. This is the scenario considered in ~\autoref{sec:capacity quantification} and \autoref{sec:effective geometry}. 

Let $d\in\N$ be the dimension of the data. We consider $P$ point manifolds $\{\mathcal{M}_i\}_{i\in[P]}$ with manifold size $M\in\N$ and radius $R$ that lies in a subspace of dimension $D$.
Each manifold is defined as
\[
    \mathcal{M}_i=\{\bu_0 + R \cdot \sum_{j=1}^D s^k_j\bu_j + \epsilon \bv_k\}_{k\in[M]}
\]
where the axes $\bu_j\sim N(0,I_d/d)$, the coordinates $s^k_j\sim N(0,1)$, the noise vectors $\bv_k\sim N(0, I_d/d)$, and $\epsilon=10^{-2}$. The pre-scaled points in the manifolds $\{\sum_{j=1}^D s_j^ku_j\}_{k\in[M]}$ are well-normalized to unit norm. 

Test manifolds share the same model except that the noise vectors $\bv_j$ are sampled again in the same distribution. 

\paragraph{Isotropic Gaussian manifolds.}
In certain experiments, we drop in the intrinsic dimension $D$ and directly consider manifolds defined as
\[
    \mathcal{M}_i=\{\bu_0 + R \cdot \bv_k\}_{k\in[M]}
\]
where the noise vectors are $\bv_k\sim N(0,I_d/d)$. Test manifolds share the same model except that the noise vectors $\bv_k$ are sampled again in the same distribution. 

\paragraph{Correlated spherical manifolds.}
To generated correlated manifolds, we consider an auto-regressive model described by the covariance matrix $C=(\rho^{|i-j|})_{ij}\in\Real^{P\times P}$, where $\rho\in[0,1)$ is either the center correlation strength $\rho_C$ or axes correlation strength $\rho_A$. The center covariance $C_C$ is then mixed into the isotropic manifold centers $\{\bu_0^j\sim N(0,I_d/d)\}_{j\in [M]}$. The axes covariance matrices $C_A^i$ is mixed into the isotropic axes $\{\bu_i^j\sim N(0,I_d/d)\}_{j\in[M]}$ for each $i=1,2,\ldots,D$ respectively. The mixing is performed through multiplying the column matrix $M_C$ or $M_A^j\in \Real^{P\times d}$ of centers or each axes with the Cholesky decomposition of $C_C$ or $C_A^i$.  To incorporate center-axes correlation, we scale each center vector $\bu_0$ by a factor of $(1+\psi\cdot q)$ where $q\sim N(0,1)$. 

\paragraph{Labels.}
For $P$ manifolds with manifold size $M$, the $P$ labels are randomly sampled from a uniform distribution on $\{\pm1\}$. Each label is associated with $M$ data points in the individual manifold. When learning with binary cross entropy, the labels are reassigned as $\{0,1\}$ during loss and gradient computation. 

\subsubsection{2-Layer neural network architecture}
The model architecture we consider is similar to the architecture mentioned in ~\autoref{app:2-layer theory}. 

Let $d\in\N$ be the input data dimension, $N\in\N$ be the number of hidden neurons, $K\in\N$ be the number of linear readouts, $\alpha\in\Real_+$ be the scaling factor of the readout weights. 

Let $W=W_0\in\Real^{N\times d}$ be the initial weight matrix of a fully connected 2-layer neural network. 
Let $\{a^i_0\}_{i\in[K]}$ be a list of initial readout weights where $a^i_0\in \Real^N$. 
Let $\sigma(\cdot):\Real\to\Real$ be a non-linear activation function, e.g. $\relu$ or $\tanh$. 

The feature of an input vector is defined as $\phi(\bx)=\sigma(W\bx)$. The 2-layer neural network parameterized by $W$ and $a^i$ is defined as
\[
f(W,a^i;\bx)=\frac{\alpha}{\sqrt{N}}\ba^\top\phi(\bx)
\]
where the label prediction for data point $\bx$ is $\sgn(f(\bx))$ when learning with the mean squared error loss function. When learning with binary cross entropy loss function, we use $\{0,1\}$ as labels and $\varsigma(f(\bx))$ as prediction instead, where $\varsigma$ is the standard sigmoid function. 

\subsubsection{Learning rule}

\paragraph{Loss function and gradient update.}
Let $\eta\in\Real_+$ be the learning rate of the weight matrix, $c\in\Real_+$ be the scaling factor of the readout learning rate, and let $\{(\bx_i,y_i)\}_{i\in[PM]}$ be the collection of training data, where $P$ is the number of manifolds and $M$ is the manifold size.

We consider gradient descent over the loss function 
$$\cL(f)=\frac{1}{\alpha^2}\frac{1}{PM}\sum_{i\in[PM]}\ell(f(\bx_i),y_i)$$
where $\ell:\Real\times\{\pm1\}\to\Real$ is either the mean squared error (MSE)
\[
\ell_{MSE}(z, y)=\frac{1}{2}(z -y)^2
\]
or $l:\Real\times\{0,1\}\to\Real$ is the binary cross entropy (BCE)
\[
\ell_{BCE}(z,y) = y \cdot \log(1 + e^{-z}) + (1 - y) \cdot \log(1 + e^z)
\]

\paragraph{Mean squared error.}
For the weight matrix, the gradient update with learning rate $\eta>0$ is $W_{t+1}=W_t+\eta G_t$ where
\[
G_t = 
\frac{1}{\alpha^2} 
\frac{1}{PM}\sum_{i\in[PM]} 
\frac{1}{K}\sum_{j\in[K]}
\left[(y_i-\frac{\alpha}{\sqrt{N}}
{\ba^j_t}^\top\sigma(W_t\bx_i))\frac{\alpha}{\sqrt{N}}\ba^j_t
\odot\sigma'(W_t\bx_i)\right]\bx_i^\top
\]
and $\sigma'(\cdot)$ denotes the first order derivative of $\sigma(\cdot)$. For each linear readout, the gradient update is $a_{t+1}=a_t+c\eta g_t$ where 
\[
g_t = \frac{1}{\alpha^2} \frac{1}{PM} \sum_{i\in[PM]} 
\left[ y_i - \frac{\alpha}{\sqrt{N}}\ba_t^\top\sigma(W_t\bx_i) \right]
\frac{\alpha}{\sqrt{N}} \sigma(W_t\bx_i) 
\]
Note that the $\alpha^{-2}$ multiplier on the loss function to ensure common convergence time when $\alpha\to\infty$ as mentioned in ~\citep{geiger2020disentangling}.

\paragraph{Binary cross entropy.}
For the weight matrix, the gradient update with learning rate $\eta>0$ is $W_{t+1}=W_t+\eta G_t$ where
\[
G_t = 
\frac{1}{\alpha^2} 
\frac{1}{PM}\sum_{i\in[PM]} 
\frac{1}{K}\sum_{j\in[K]}
\left[(y_i-\varsigma[\frac{\alpha}{\sqrt{N}}
{\ba^j_t}^\top\sigma(W_t\bx_i)])\frac{\alpha}{\sqrt{N}}\ba^j_t
\odot\sigma'(W_t\bx_i)\right]\bx_i^\top
\]
where $\varsigma$ denotes the standard sigmoid function and $\sigma$ denotes the activation function. For each linear readout, the gradient update is $a_{t+1}=a_t+c\eta g_t$ where 
\[
g_t = \frac{1}{\alpha^2} \frac{1}{PM} \sum_{i\in[PM]} 
\left[ y_i - \varsigma[\frac{\alpha}{\sqrt{N}}\ba_t^\top\sigma(W_t\bx_i)] \right]
\frac{\alpha}{\sqrt{N}} \sigma(W_t\bx_i) 
\]

If not otherwise noted, we conduct experiments with the MSE loss function and $\relu$ activation function by default. 

\paragraph{A Note on Learning rate.}
We define $\bar\eta=\eta\alpha^{-1}$ as the normalized effective learning rate. During training, We implicitly scale the learning rate $\eta$ by a factor of $\sqrt{N}$ in the experiments to enter the rich regime as mentioned in ~\citep{ba2022high}.

\subsubsection{Training}
For each 2-layer neural network experiment conducted in the paper, forty random seeds are chosen from $0$ to $39000$ with an interval of $1000$ to train forty models in parallel for $10^5$ epochs. All training are conducted on the Flatiron Institute high performance computing clusters. 

\subsubsection{Feature extraction}
During analysis, fifty epochs are sampled uniformly in log-scale. For each model at checkpoint epoch $t$, we extract total $P$ size $M$ manifold representations $\{\Phi_t(\bx_i)\}_{i\in[PM]}$ associated with labels $\{y_i\}_{i\in[PM]}$. We perform conventional analysis and manifold capacity analysis described in ~\autoref{app:related work} and ~\autoref{app:MCT} respectively. We will present more details in the following experiment sections. 

\subsection{Capacity is a robust measure of feature learning across architecture, data, and learning rule variations}\label{app:2l:arch_variation}

The purpose of this section is to support ~\autoref{sec:capacity quantification} by showcasing that capacity is able to quantify feature learning even when model architecture, data distribution, and learning rule varies. 

\subsubsection{Feature analysis methods}

Here, we briefly present the conventional feature analysis methods and capacity analysis method and how they are computed in the experimental setup. 

\paragraph{Representation level analysis.}
Activation stability is a representation level metric that intuitively captures how much neurons are activated in hidden units. Formally, we define it as
\[
\frac{\sum_{i=1}^{PM}\sum_{j=1}^N\mathbf{1}_{>0}(\phi_j(\bx_i))}{PMN}
\]
Another conventional method to disentangle feature learning at representation level is tracking the norm of deviation from initial weights ~\citep{jacot2018neural}
$$\frac{\|W_t-W_0\|}{\|W_0\|}$$
On the other hand, the cosine similarity ~\citep{liu2024connectivity} can be used to study alignment at representation level
\[
\frac{\Phi_t\Phi_0}{\|\Phi_t\|\|\Phi_0\|}
\]
where $(\Phi_t)_{ij} = \phi_t(\bx_i)\cdot\phi_t(\bx_j)\in\Real^{PM\times PM}$ is the gram matrix of features over the test data. 

\paragraph{Kernel methods.}
The kernel methods for quantifying feature learning involves computing the Neural Tangent Kernel (NTK) ~\citep{jacot2018neural} for each pair of test data points:
\[
\Theta_t(\bx_1,\bx_2) = \nabla_{w_t} f(\bx_1) \cdot \nabla_w f(\bx_2) 
\]
where $\nabla_{w_t} f$ denotes the total gradient of the neural network at epoch $t$ with respect to the hidden weights $W_t$ and readout weights $\{a^j_t\}$. Note that we scale the readout contribution to the total gradient by the readout learning rate factor $c\in\Real_+$ aforementioned. Hence,
\[
\nabla_w f(\bx) = \nabla_{W_t} f(\bx) + \frac{1}{K}\sum_{j=1}^K \nabla_{a_t^j} f(\bx)
\]
After obtaining the gram matrix $\Theta_t=\Theta_t(\bx_i,\bx_j)_{ij}\in\Real^{PM\times PM}$ from the test data, we can compute the \emph{NTK change} defined as
$$\frac{\|\Theta_t-\Theta_0\|}{\|\Theta_0\|}$$
which can be interpreted as the relative deviation of the the kernel from initialization in the Frobenius norm metric. Conventionally studied, NTK change disentangles lazy and feature learning, as detailed in ~\citep{jacot2018neural}. We present NTK change in ~\autoref{sec:capacity quantification} ~\autoref{fig:capacity} to compare it with capacity as the metric to track feature learning. 

The \emph{kernel alignment} can be similarly defined as the cosine similarity of initial and current NTK gram matrices:
\[
\frac{\Theta_t\Theta_0}{\|\Theta_t\|\|\Theta_0\|}
\]
which can be interpreted as the relative deviation of the kernel from initialization in terms of alignment. Kernel alignment is also studied in ~\citep{liu2024connectivity} to disentangle lazy and feature learning.

The \emph{centered kernel alignment} ~\citep{kornblith2019similarity} is another approximation method to study kernel evolution when the gram matrices is large:
\[
\frac{HSIC(\Theta_t, \Theta_0)}{\sqrt{HSIC(\Theta_t, \Theta_t)HSIC(\Theta_0, \Theta_0)}}
\]
where
\[
HSIC=\frac{\Tr(\Theta_tL\Theta_0L)}{(n-1)^2}
\]
These kernel metrics can be readily computed from the trained models and extracted features. 

\paragraph{Capacity and effective geometry.}

For more details on data-driven manifold capacity analysis, please refer to \autoref{app:MCT}.



\paragraph{Setup of~\autoref{fig:capacity}a.} In \autoref{fig:capacity}a, we showcase that the degree of feature learning is controlled by the effective learning rate $\bar\eta$ with the following standard setup:
\begin{itemize}
    \item Data: Isotropic Gaussian manifolds with $R=0.5, M=15$.
    \item Model: We set $\sigma=\relu, N=1500, d=1000, P=100, K=1$.
    \item Learning rule: We set $\ell=\ell_{MSE}, \eta=50, c=0$ and $$\alpha=10/128, 10/112, 10/96, 10/80, 10/64, 10/16, 10/4, 10/1$$ so that the normalized effective learning rates are $$\bar\eta=128, 112, 96, 80, 64, 16, 4, 1$$ which is computed by $\bar\eta=\frac{\eta\alpha^{-1}}{5}$ where the division by $5$ normalizes the smallest $\eta\alpha^{-1}$ to be $1$.
    \item Training: We trained the models for $100000$ epochs with $40$ repetitions per parameter combination.
    \item Plotting: We use sample mean and 95\% confidence interval for each data point.
\end{itemize}

\subsection{Effective geometry reveals distinct learning dynamics}\label{app:2l:task_variation}

\subsubsection{Learning strategies}

\paragraph{Compression strategy setup} In \autoref{fig:learning geometry}b where the networks performs the compression strategy, we use a difficult-task setup with higher data manifold radius and more readout tasks:
\begin{itemize}
    \item Data: Isotropic spherical manifolds with $R=1.0, D=8, M=15$.
    \item Model: We set $\sigma=\relu, N=300, d=200, P=20, K=27$.
    \item Learning rule: we set $\ell=\ell_{MSE}, \alpha=1, c=0$ and $$\eta=1, 5, 10, 20, 30, 40, 50, 60, 70, 80, 90, 100, 110, 120, 130, 140, 150$$ so that the normalized effective learning rates are $$\bar\eta=1, 5, 10, 20, 30, 40, 50, 60, 70, 80, 90, 100, 110, 120, 130, 140, 150.$$
    \item Training: We trained the models for $100000$ epochs with $40$ repetitions per parameter combination.
    \item Plotting: We use sample mean for each data point.
\end{itemize}

\paragraph{Flattening strategy setup.} In \autoref{fig:learning geometry}b where the networks performs the flattening strategy, we use an easy-task setup with smaller data manifold radius and very few readout tasks:
\begin{itemize}
    \item Data: Isotropic spherical manifolds with $R=0.5, D=8, M=15$.
    \item Model: We set $\sigma=\relu, N=300, d=200, P=20, K=3$.
    \item Learning rule: we set $\ell=\ell_{MSE}, \alpha=1, c=0$ and $$\eta=80, 90, 100, 110, 120, 130, 140, 150, 160, 170$$ so that the normalized effective learning rates are $$\bar\eta=80, 90, 100, 110, 120, 130, 140, 150, 160, 170.$$
    \item Training: We trained the models for $100000$ epochs with $40$ repetitions per parameter combination.
    \item Plotting: We use sample mean for each data point.
\end{itemize}

\paragraph{Contour plot of learning strategies.}

In \autoref{fig:learning geometry}b and c, we use contour plots to visualize the different learning strategies adopted by the network. We use Equation 34 in \citep{chung2018classification} to approximate capacity using effective radius and dimension:
\[
\alpha=\frac{1 + \left(\frac{1}{R_M^2}\right)}{D_M}
\]
The scatter points with the same color correspond to a model trained with the same normalized effective learning rate $\bar\eta$ over different epochs. 

\subsubsection{Learning stages}

\paragraph{Setup.} 

In \autoref{fig:learning geometry}a, we adopt a setup with moderate radius and number of readout tasks that shows clean learning stages:
\begin{itemize}
    \item Data: Isotropic spherical manifolds with $R=1, D=8, M=15$.
    \item Model: We set $\sigma=\relu, N=300, d=200, P=20, K=5$.
    \item Learning rule: we set $\ell=\ell_{MSE}, \eta=10, \alpha=1, c=0$ so that the normalized effective learning rate is $\bar\eta=10$. 
    \item Training: We trained the models for $100000$ epochs with $40$ repetitions per parameter combination.
    \item Plotting: We use sample mean for each data point.
\end{itemize}

\section{Deep Neural Networks}\label{app:dnn}
\subsection{Experimental setup}\label{app:dnn_setting}
In this section, we provide detailed information about the experimental setup for deep neural networks, including model architectures, datasets, training procedure, and manifold capacity measurements.

\subsubsection{Models}
We use the VGG-11 models ~\citep{simonyan2015very} for experimental results in the main paper. We also repeat these experiments on ResNet-18 ~\citep{he2016deep}. The specific implementation follows a similar setting in ~\citep{chizat2019lazy} and is adapted from \texttt{https://github.com/edouardoyallon/lazy-training-CNN}.

\paragraph{Output rescaling \label{def:DNN_etabar}.} 
As previously studied in ~\citep{chizat2019lazy}, multiplying the model outputs by a large scaling factor $\beta$ can induce lazy learning (we use the notation $\beta$ instead of $\alpha$ in ~\citep{chizat2019lazy} to avoid confusion with the notation $\alpha$ as capacity in ~\cref{def:storage capacity} ). In this section, we use the inverse scaling factor $\beta^{-1}$ as the parameter to control the degree of feature learning. We define the \textit{normalized effective learning rate} $\overline{\eta} = \beta^{-1}$. We also note several adjustments to the common training framework to adapt to using the inverse scaling factor $\beta^{-1}$ as the parameter to control the degree of feature learning.
\begin{itemize}
    \item Rescaled loss function: To adjust for using the scaling factor $\beta$, we use the rescaled loss function $L_{\beta} = \frac{L}{\beta^2}$ with $L$ denotes the loss function to accommodate for the time parameterization of the loss dynamic for large $\beta$ as previously indicated in ~\citep{chizat2019lazy} and ~\citep{geiger2020disentangling}.
    \item Model's initial outputs as $0$: As mentioned in ~\citep{chizat2019lazy}, for the scaling factor $\beta$ to be able to control the rate of feature learning, the model output as initialization $f(W_0)$ must be equal $0$. To ensure this condition, we set $f(W_t) = h(W_t) - h(W_0)$ with $W_t$ be the model's weight at training step $t$, $h$ be the output of the network, and $f$ be the final adjusted network output.
\end{itemize}

\paragraph{Number of repetitions.} 
All model measurements (train accuracy, test accuracy, activation stability, etc.) are reported as the mean of 5 independently trained model (with different random seeds). The error bar indicates the bootstraped 95\% confidence interval calculated using \texttt{seaborn.lineplot(errorbar=('ci', 95))}.

\subsubsection{Dataset}
In this section, we list detailed information about the dataset used in the paper.
\paragraph{CIfAR-10.} The CIfAR-10 dataset ~\citep{cifardataset} consists of 60000 32x32 colour images in 10 classes, with 6000 images per class. There are 50000 training images and 10000 test images.
\paragraph{CIfAR-100.} The CIfAR-100 dataset ~\citep{cifardataset} is similar to CIfAR-10, except that it has 100 classes containing 600 images each. There are 500 training images and 100 testing images per class. Note that the images in CIfAR-10 and CIfAR-100 are mutually exclusive.
\paragraph{CIfAR-10C.} The CIfAR-10C dataset ~\citep{hendrycks2018benchmarking} includes images from the CIfAR-10 evaluation set with common corruptions such as Gaussian noise, fog, motion blur, etc. The dataset has 15 different common corruption types, and 5 different severity levels for each corruption type.

\subsubsection{Training procedure}
\begin{itemize}
    \item Loss function: We follow the theoretical results and practice used in ~\citep{chizat2019lazy} to use mean-squared error loss to train all DNNs mentioned in the paper.
    \item Optimizer: We use Stochastic Gradient Descent with momentum (implemented as \texttt{torch.optim.SGD(momentum=0.9)}) to train the models.
    \item Data augmentation: We apply the following data augmentation during training: \texttt{RandomCrop(32, padding=4)}, \texttt{RandomHorizontalFlip}.
    \item Learning rate and learning schedule: We follow the practice in ~\citep{chizat2019lazy} and set initial learning rate $\eta_0 = 1.0$ for VGG-11 and $\eta_0 = 0.2$ for ResNet-18. The learning rate schedule is defined as $\eta_t = \frac{\eta_0}{1+ \frac{1}{3}t}$.
    \item Initialization: We follow the practice in ~\citep{chizat2019lazy} to initialize the model's weight using Xavier initialization ~\citep{glorot2010understanding} and the bias to be $0$.
    \item Batch size: We use batch size of 128 during training and batch size of 100 during evaluation.
\end{itemize}

\subsubsection{Manifold capacity measurements}\label{app:dnn_capacity}
In this section, we provide detailed information about how we define object manifolds from the model's representations and measure the manifold capacity and geometric properties ~\citep{chung2018classification}.
\begin{itemize}
    \item Features extraction: For each image, we extract the object representation from the last linear layer (dimension $512$) before the classification layer (dimension $10$).
    \item Number of manifolds: We use 10 object manifolds for each measurement.
    \item Number of points per manifold: For each object manifold, we randomly sample 50 images from the interested class.
    \item Number of repetitions: Every capacity and geometry measurement is repeated 10 times per model instance (50 times if we have 5 model repetitions) and we report the mean and the error bar as the bootstraped 95\% confidence interval calculated using \texttt{seaborn.lineplot(errorbar=('ci', 95))}.
\end{itemize}

\subsection{Capacity quantifies the degree of feature learning in deep neural networks}

\begin{figure}[ht]
    \centering
    \includegraphics[width=\textwidth]{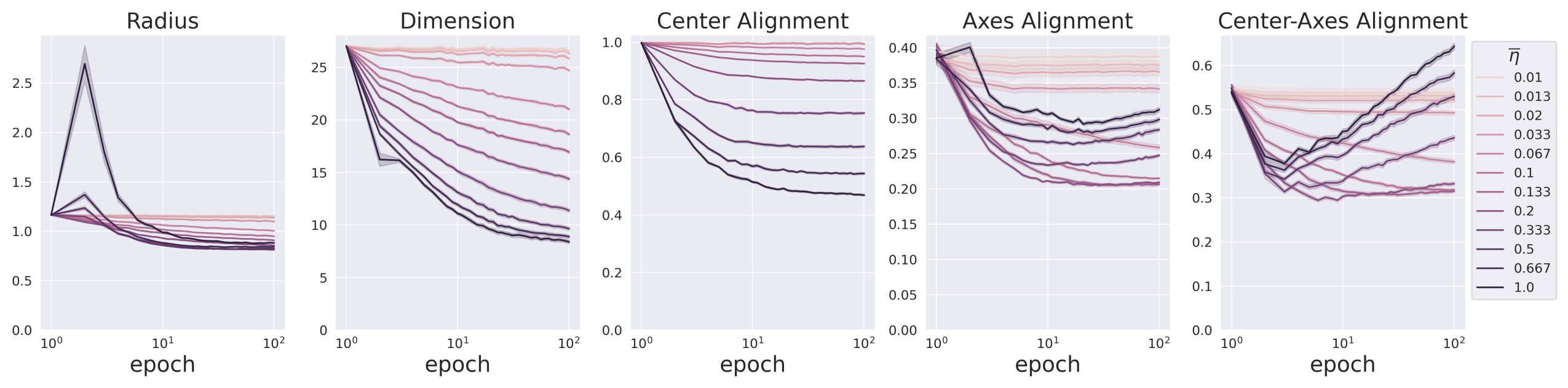}
    \caption{Manifold capacity and geometry for VGG-11 models trained with different $\overline{\eta}$}
    \label{fig:app_vgg_capacity}
\end{figure}

\paragraph{Capacity and manifold geometry for VGG-11 models.} 
In ~\autoref{fig:capacity}, we show manifold capacity along with other common metrics used to identify feature learning such as train accuracy, test accuracy, relative weight norm change, and activation stability. In this section, we provide other manifold geometric measurements along with manifold capacity in ~\autoref{fig:app_vgg_capacity}.

\begin{figure}[ht!]
    \centering
    \includegraphics[width=\textwidth]{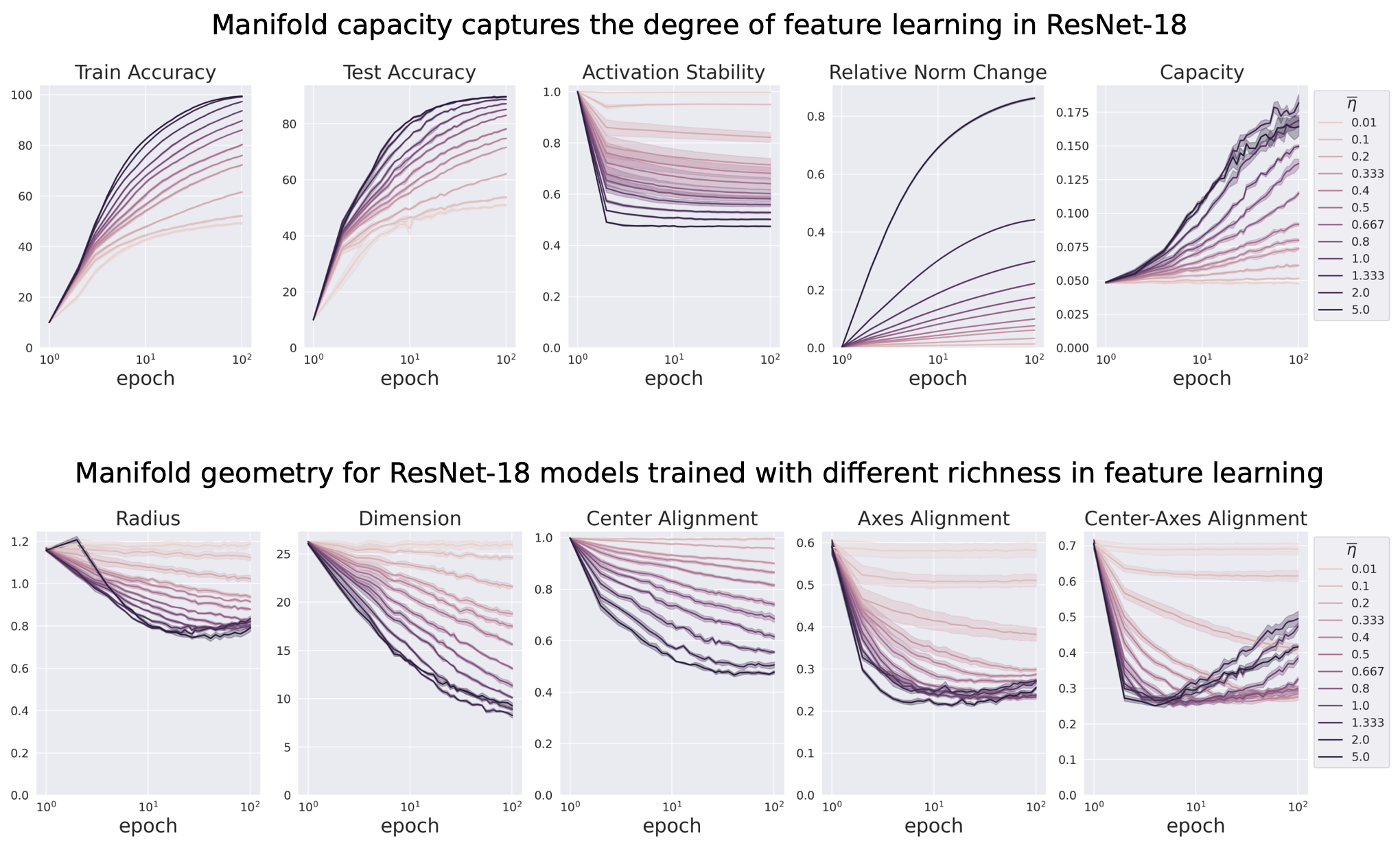}
    \caption{Manifold capacity and geometry of ResNet-18 models trained with different scale factor.}
    \label{fig:app_resnet_capacity}
\end{figure}

\paragraph{Capacity quantifies the degree of feature learning in ResNet-18 models.}
In section ~\autoref{sec:capacity quantification}, we show that manifold capacity can capture the degree of feature learning in DNNs, specifically in VGG models. In this section, we empirically show this statement can also be extended to other model architectures, specifically ResNets, in ~\autoref{fig:app_resnet_capacity}.

\begin{figure}[ht]
    \centering
    \includegraphics[width=\textwidth]{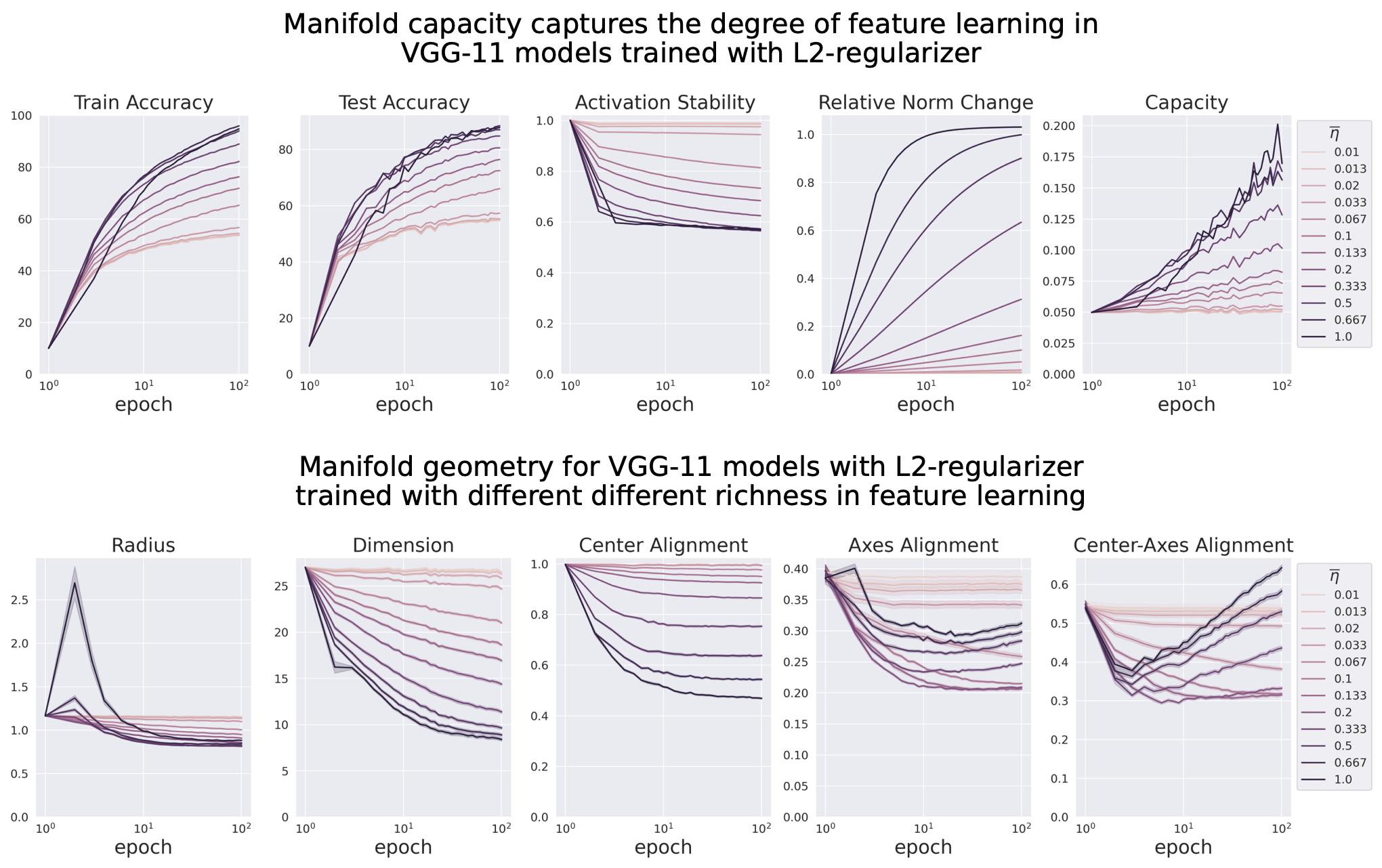}
    \caption{Manifold capacity and geometry of VGG-11 models with L2-regularizer trained with different scale factor.}
    \label{fig:app_vgg_capacity_reg}
\end{figure}

\paragraph{Capacity quantifies the degree of feature learning in VGG-11 models trained with weight regularizer.}\label{app:vgg_capacity_reg}
While most theoretical work in the lazy vs rich learning literature are formulated with vanilla mean squared error (MSE) loss~\citep{jacot2018neural}~\citep{chizat2019lazy}, in practice, MSE with weight regularizer (or weight decay) is used widely to prevent over-fitting and improve model generalization. In ~\autoref{fig:app_vgg_capacity_reg}, we explore the effect of weight decay to feature learning and demonstrate empirically that capacity can still quantify the degree of feature learning in models trained with L2-regularizer. We implemented L2-regularizer by setting \texttt{torch.optim.SGD(weight\_decay=0.0002)}. We leave further study about the impact between the magnitude of weight regularizer and effective learning rate (and/or scaling factor) to the degree of feature learning as a potential future direction.

\subsection{Manifold capacity and manifold geometry delineate learning stages in deep neural networks}
In section ~\cref{subsec:learning_stages}, we have demonstrated the use of effective manifold geometry to uncover hidden learning stages in 2-layer neural networks. In this section, we showed that using similar technique, we can also discover geometric learning stages in deep neural networks as well.

\paragraph{Experiment setup} We used similar setup mentioned in ~\autoref{app:dnn_setting}. In this section, to give a higher resolution into the learning dynamic, we extracted the model checkpoint at each training step (after each training batch, with \texttt{batch\_size=100}) instead of each training epoch (after a whole train dataset iteration).

\subsection{Feature learning and downstream task: out-of-distribution generalization}\label{app:OOD}
In this section, we measure the performance of the models trained with different degree of feature learning (quantified by effective learning rate $\overline{\eta}$) on the downstream tasks for OOD using CIfAR-100, a dataset with no overlap with CIfAR-10, the dataset used to train the model.

\subsubsection{Experimental setup}
We use linear probe ~\citep{alain2016understanding} on representation from the last linear layer (dimension $512$) to measure the performance of models trained on CIfAR-10 on the out-of-distribution dataset, CIfAR-100. Linear probes are linear classifiers trained on top of the representation to probe how much information the representations encode about a particular task or characteristic. This approach has been used widely in different fields including natural language processing ~\citep{belinkov2017} and computer vision ~\citep{raghu2021vit}.

Here we provide detailed information about how we construct the linear probes.


\paragraph{Optimizer.} 
We use \textit{Adam} optimizer with initial learning rate $\eta_0 = 0.1$ and learning rate schedule is defined as $\eta_t = \frac{\eta_0}{1+ \frac{1}{3}t}$. Other parameters are default \texttt{Pytorch} parameters.

\paragraph{Number of epochs.} 
The linear probe is trained for 50 epochs, unless it is stopped early, as described by the early stop method below.

\paragraph{Early stop.} 
During training, if the validation loss is greater than the minimum validation loss so far for more than $N_{patience}$ epoch, then training is stopped. We set $N_{patience} = 3$.

\subsubsection{OOD performance for ResNet-18}
\begin{figure}[ht]
    \centering
    \includegraphics[width=\textwidth]{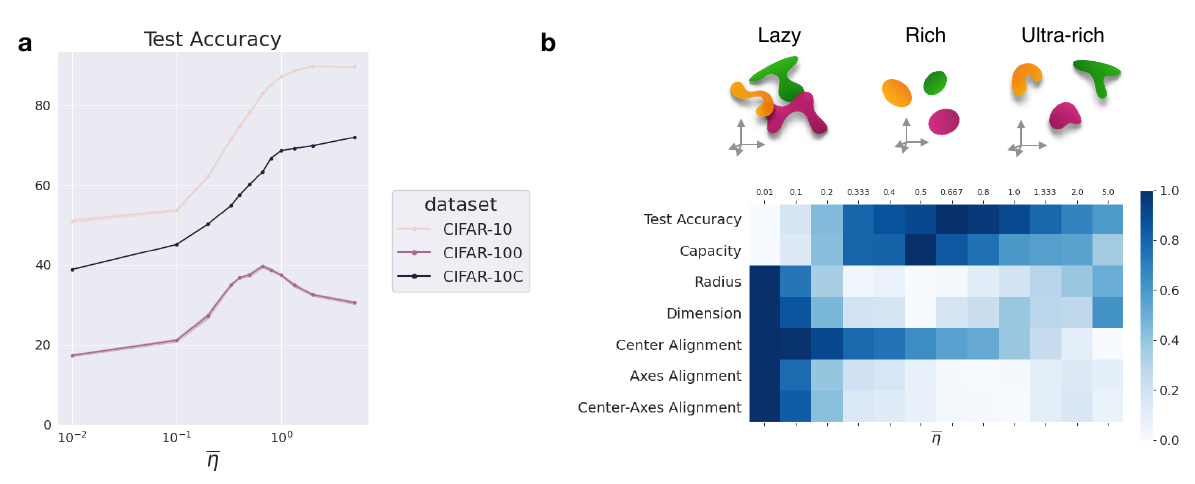}
    \caption{OOD performance and effective geometric measure of ResNet-18 models trained with different scale factor.}
    \label{fig:app_ood_resnet}
\end{figure}

In~\cref{subsec:ood_few_shot}, we demonstrate how capacity and effective manifold geometry can be used to characterize the OOD performance of VGG-11 models trained with different 
effective learning rate $\overline{\eta}$. In this section, we show OOD performance and effective geometry of ResNet-18 models trained with different effective learning rate $\overline{\eta}$ in~\autoref{fig:app_ood_resnet}. Interestingly, unlike VGG-11, for ResNet-18, the failure of models in the ultra-rich regime is characterized by the expansion of manifold dimension, not manifold radius.

\section{Recurrent Neural Networks}\label{app:rnn}
\subsection{Experimental Setup}
In this section, we provide detailed information about the experimental setup for recurrent neural network in~\cref{subsec:rnn}, including model architectures, datasets, training procedure, and manifold capacity measurements.
\subsubsection{Dataset}
We used the package \texttt{neurogym} ~\citep{neurogym} to simulate common cognitive tasks. In this paper, we trained recurrent neural networks to perform the following cognitive tasks: perceptual decision making, context decision making, and delay match sample. We followed the task configuration used in ~\citep{liu2024connectivity}. We list detailed information of task configuration and descriptions below.
\paragraph{Perceptual decision making~\citep{britten1992} (\href{https://neurogym.github.io/envs/PerceptualDecisionMaking-v0.html}{documentation page})}
\begin{itemize}
    \item Task description: In each trial, given two noisy stimulus, the agent needs to integrate the stimulus over time to determine which stimuli has stronger signal.
    \item Task configuration: We set up the task using the following parameters: \texttt{\{timing: \{fixation: 0, stimulus: 700, delay: 0, decision: 100\}, dt: 100, seq\_len: 8\}}
\end{itemize}

\paragraph{Context decision making ~\citep{mante2013context} (\href{https://neurogym.github.io/envs/ContextDecisionMaking-v0.html}{documentation page})} 
\begin{itemize}
    \item Task description: In each trial, given two noisy stimulus, each has two modalities, the agent needs to integrate the stimulus in one specific modal while ignoring the other modal. The interested modal is given by the context.
    \item Task configuration: We set up the task using the following parameters: \texttt{\{timing: \{fixation: 0, stimulus: 200, delay: 500, decision: 100\}, dt: 100, seq\_len: 8\}}
\end{itemize}

\paragraph{Delay match sample ~\citep{miller1996neural} (\href{https://neurogym.github.io/envs/DelayMatchSample-v0.html}{documentation page})}
\begin{itemize}
    \item Task description: In each trial, a sample stimulus is shown during the sample period, which followed by a delay period. Afterwards, the test stimulus is shown. The agent needs to determine whether the sample and the test stimuli are matched.
    \item Task configuration: We set up the task using the following parameters: \texttt{\{timing: \{fixation: 0, sample: 100, delay: 500, test: 100, decision: 100\}, dt: 100, seq\_len: 8\}}
\end{itemize}

\subsubsection{Models}

\paragraph{Model architecture}
We consider time-continuous recurrent neural networks (RNNs) architecture that are commonly used to model neural circuits ~\citep{liu2024connectivity, ehrlich2021psychrnn}. Specifically, we consider RNNs with 1 hidden layer, ReLU activation, $N_{in}$ input units, $N_{hidden}$ hidden units, and $N_{out}$ output unit. Let $x_t \in \Real^{N_{in}}$, $y_t \in \Real^{N_{out}}$ be the corresponding input and output at time-step $t$. The model's hidden representation $h_t$ and outputs $\hat{y}_t$ at time step $t$ can be defined by the given equations:
\begin{equation}\label{eq:hiddenrnn}
    h_{t+1} = \rho h_t + (1 - \rho)(W_h\sigma(h_t) + W_i x_t) 
\end{equation}
\begin{equation}
    \hat{y}_t = W_o \sigma(h_t)
\end{equation}
In the above equation, $W_i \in \Real^{N_{in} \times N_{hidden}}$, $W_h \in \Real^{N_{hidden} \times N_{hidden}}$, $W_o \in \Real^{N_{hidden} \times N_{out}}$. $\sigma(.)$ is the non-linear activation function, in which we used ReLU, and $\rho$ is the decay factor which is defined by $\rho = e^{\frac{-dt}{\tau}}$ with time step $dt$ and time constant $\tau$. We use $N_{hidden} = 300$ for all RNNs models.
\paragraph{Weight rank initialization}
Following the practice in \citep{liu2024connectivity}, we initialize the recurrence weight $W_h$ by initializing an initial full-ranked random Gaussian matrix, and then use Singular Value Decomposition to truncate the weight rank to the desired rank. The truncated weight matrix is then re-scaled to ensure that weight matrices with varying ranks have the same weight norm. 
\subsubsection{Training Procedure}
\begin{itemize}
    \item Loss function: Since all three tasks that we consider are classification tasks, we use cross entropy loss.
    \item Optimizer: We use Stochastic Gradient Descent with momentum (implemented as \texttt{torch.optim.SGD(lr=0.003, momentum=0.9)}) to train the models.
    \item Batch size: We use batch size of 32 for each training step.
\end{itemize}
The models are trained for 10000 iterations and all models being compared achieved similar loss and accuracy after training (see Figure ~\autoref{fig:app_rnn_2af}, ~\autoref{fig:app_rnn_cxt}, ~\autoref{fig:app_rnn_dms} for more details).

\subsubsection{Manifold Capacity Measurements}
In this section, we provide detailed information about how we define object manifolds from the model's representations and measure the manifold capacity and geometric properties ~\citep{chung2018classification}.
\begin{itemize}
    \item Features extraction: We extract the  representation $h_t$ (in Equation ~\autoref{eq:hiddenrnn}) from the hidden layer (dimension $300$) with $t$ being the decision period of the trial.
    \item Number of manifolds: The number of possible choices in the decision period of all the three tasks that we consider is 2, so the number of manifolds are 2.
    \item Number of points per manifold: For each task-relevant manifold, we randomly sample 50 trials of the corresponding ground truth choices.
    \item Number of repetitions: Every capacity and geometry measurement is repeated 50 times and we report the mean and the error bar as the bootstraped 95\% confidence interval calculated using \texttt{seaborn.lineplot(errorbar=('ci', 95))}.
\end{itemize}
\subsection{Additional results on other cognitive tasks}
\begin{figure}[ht!]
    \centering
    \includegraphics[width=\textwidth]{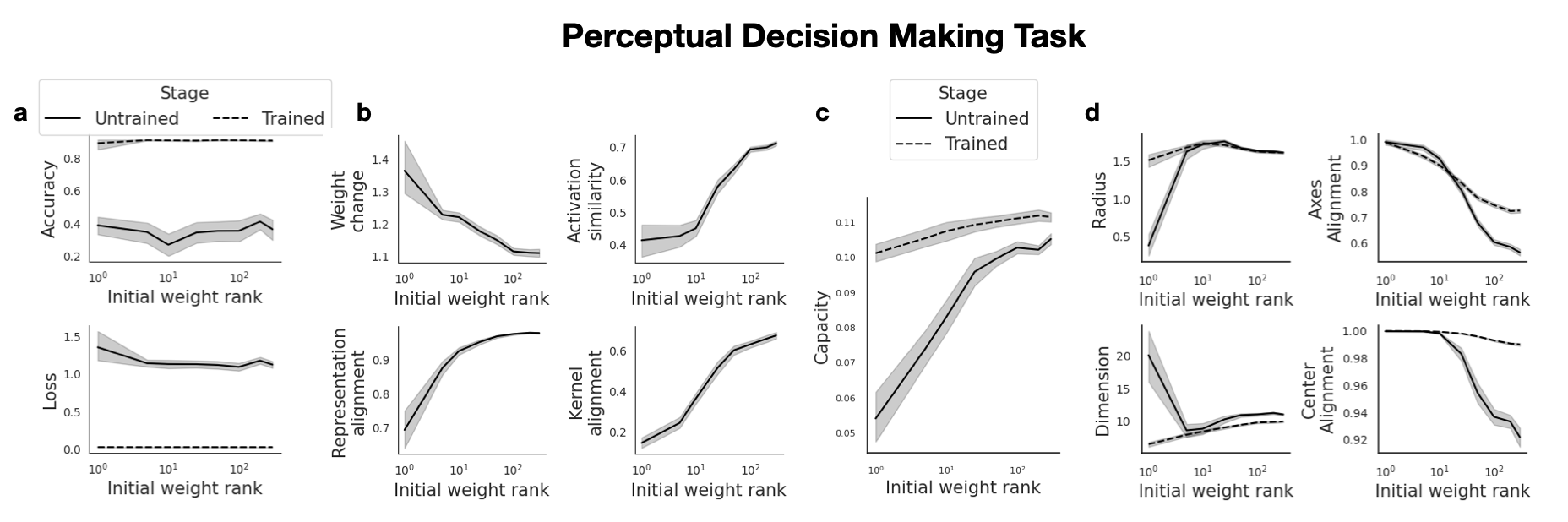}
    \caption{Structural connectivity bias in the two-alternative forced choice task. \textbf{a.} Model train and loss accuracy \textbf{b.} Weight change and alignment measurements \textbf{c.} Manifold capacity measurements \textbf{d.} Effective manifold geometry measurements.}
    \label{fig:app_rnn_2af}
\end{figure}

\begin{figure}[ht!]
    \centering
    \includegraphics[width=\textwidth]{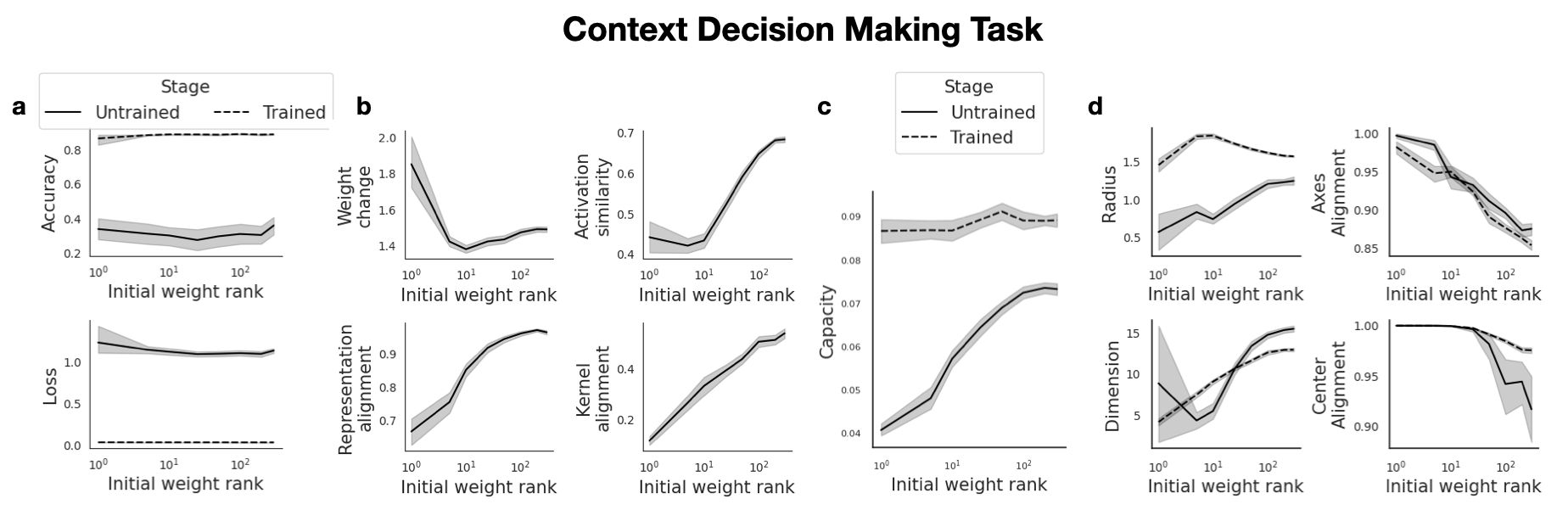}
    \caption{Structural connectivity bias in the context decision making task \textbf{a.} Model train and loss accuracy \textbf{b.} Weight change and alignment measurements \textbf{c.} Manifold capacity measurements \textbf{d.} Effective manifold geometry measurements.}
    \label{fig:app_rnn_cxt}
\end{figure}

\begin{figure}[ht!]
    \centering
    \includegraphics[width=\textwidth]{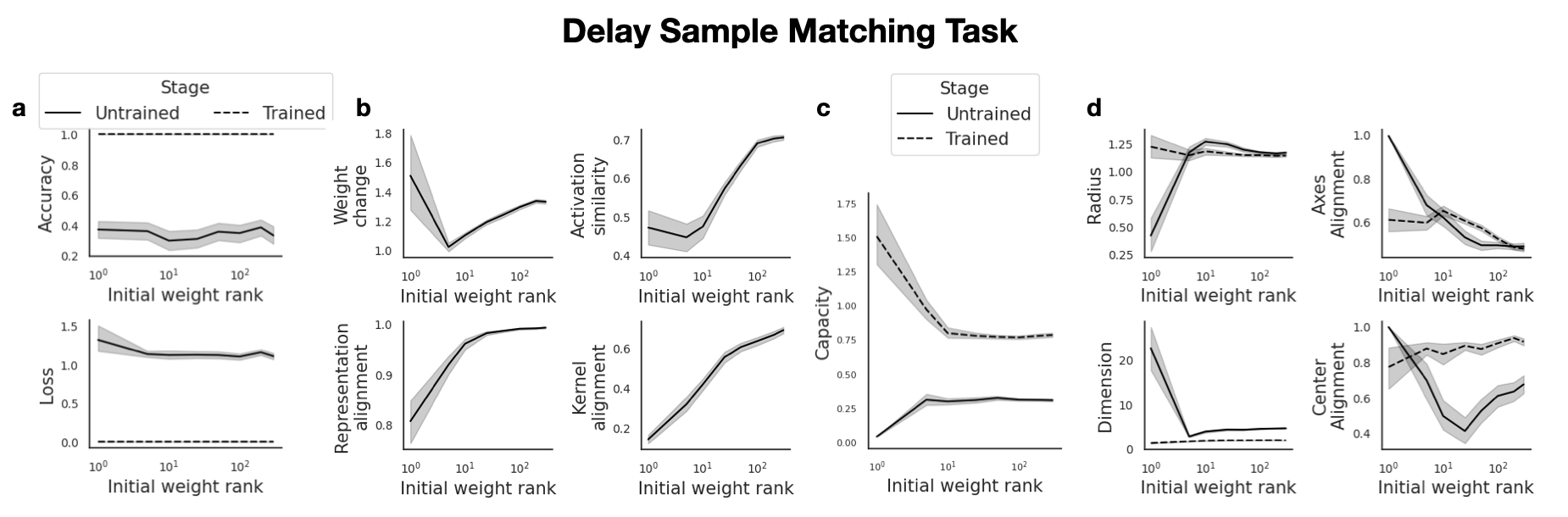}
    \caption{Structural connectivity bias in the delay mataching sample task. \textbf{a.} Model train and loss accuracy \textbf{b.} Weight change and alignment measurements \textbf{c.} Manifold capacity measurements \textbf{d.} Effective manifold geometry measurements.}
    \label{fig:app_rnn_dms}
\end{figure}

In section ~\cref{subsec:rnn}, we present the results on how the initial structural connectivity bias (initialized by varying the rank of the weight matrix) affects the feature learning regime and representational geometry of a given model in the perceptual decision making task (also called the two-alternative forced choice task) ~\citep{britten1992}. In this section, we show more detailed results (including accuracy and loss) on the perceptual decision making task in Figure ~\autoref{fig:app_rnn_2af}, along with two other cognitive tasks, which are context decision making task ~\citep{mante2013context} in Figure ~\autoref{fig:app_rnn_cxt} and delay match sample task ~\citep{miller1996neural} in Figure ~\autoref{fig:app_rnn_dms}.

\end{document}